\documentclass[runningheads]{llncs}

\usepackage{accv}

\usepackage{accvabbrv}

\usepackage{graphicx}
\usepackage{xcolor}
\usepackage{booktabs}
\usepackage{wrapfig}
\usepackage{multirow}
\usepackage{makecell}
\usepackage{graphicx}

\usepackage[table]{xcolor}
\definecolor{bestcolor}{RGB}{210, 230, 255}
\definecolor{secondcolor}{RGB}{220, 245, 220}
\definecolor{thirdcolor}{RGB}{255, 245, 210}

\usepackage[accsupp]{axessibility}  

\usepackage{hyperref}

\usepackage{orcidlink}

\begin{document}

\title{Learning Semantic Inpainting for Animatable Gaussian Head Avatars}

\titlerunning{SInGA}

\author{Pilseo Park\orcidlink{0000-0002-2621-5486} \and
Fizza Rubab\orcidlink{0009-0001-6979-7746} \and
Yiying Tong\orcidlink{0000-0002-7929-4333}}

\authorrunning{P. Park et al.}

\institute{
Michigan State University\\
\email{\{parkpils,rubabfiz,ytong\}@msu.edu}}

\maketitle

\begin{abstract}
We present SInGA, a novel method for learning \textbf{S}emantic \textbf{In}painting for animatable \textbf{G}aussian head \textbf{A}vatars from a single image. Existing avatar approaches often rely on multi-view observations and lack effective handling of unobserved regions in single-view settings, limiting their applicability in such scenarios. To address this, we propose a semantic inpainting framework defined in UV space for completing unobserved facial regions. Our key insight lies in the structured topology of the UV representation, which provides consistent spatial correspondences and enables reliable completion of identity-specific features using the inherent symmetry cues of human faces. We extract features from observed regions and use them to complete unobserved regions. The completed representation is then used to regress Gaussian attributes, effectively performing Gaussian inpainting. In addition, instead of relying on a single Gaussian at each surface or pixel location, we stack multiple Gaussians to enhance detail. The resulting avatar generalizes across identities without requiring per-identity optimization and can be animated with driving inputs. Experimental results show that our method generates high-quality head avatars with improved completeness and identity preservation, while supporting realistic animation and consistent rendering from unobserved views.
\end{abstract}

\section{Introduction}
\label{intro}

Reconstructing animatable 3D head avatars is a crucial area of research in computer vision and graphics, with broad applications in telepresence, the video game industry, virtual reality, and digital content creation. Recently, there has been growing interest in the single-image setting due to its simplicity and wide applicability. However, this area remains challenging in two key aspects: reconstructing unobserved regions from incomplete visual evidence with ambiguous geometry and appearance, and modeling realistic and consistent facial deformations across diverse expressions, making it difficult to preserve identity during animation.

Early 2D-based approaches~\cite{zhang2023sadtalker,xu2024vasa,siarohin2019first,wang2021one} employ Generative Adversarial Networks (GANs)~\cite{goodfellow2014generative} to animate source images by estimating deformation fields or conditioning on facial landmarks and latent codes. Despite their progress, the lack of explicit 3D structure limits their robustness under large pose and expression variations and leads to inconsistent results across viewpoints.

\begin{figure}[tb]
  \centering
  \includegraphics[width=\textwidth]{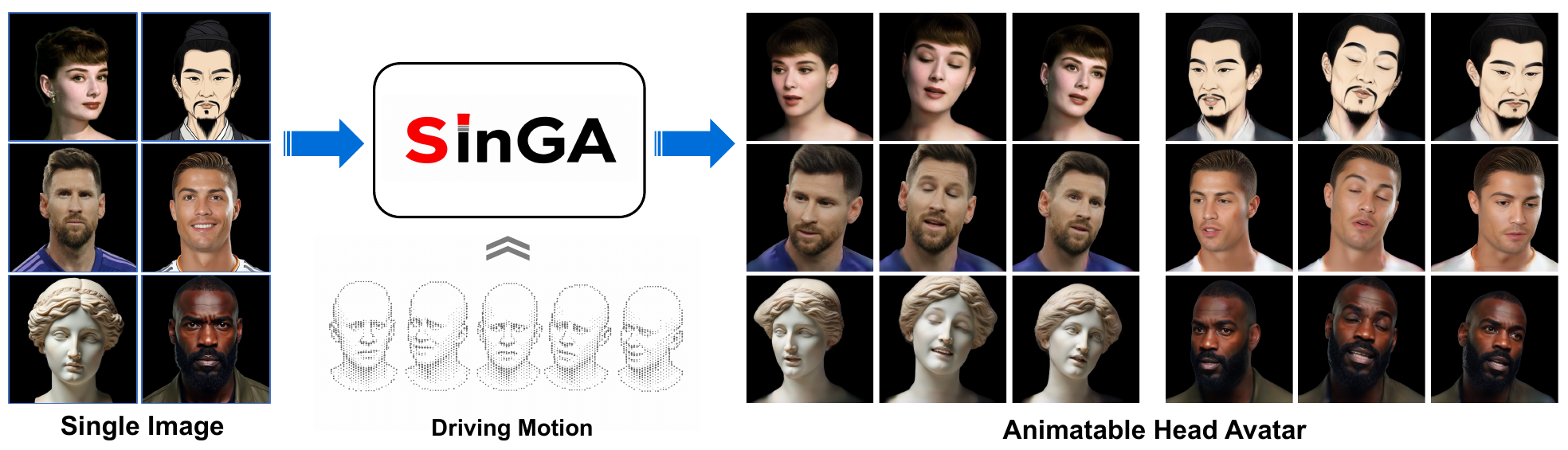}
  \caption{
\textbf{SInGA.} Our method reconstructs an animatable Gaussian head avatar from a single image in a single forward pass. The reconstructed avatar supports real-time reenactment while maintaining generalization and controllable facial animation.
  }
  \label{fig:teaser}
\end{figure}

The development of 3D-based methods such as Neural Radiance Fields~\cite{NeRF} (NeRF) and 3D Gaussian Splatting~\cite{Kerbl20233DGS} (3DGS) has improved rendering quality and multi-view consistency. NeRF-based methods~\cite{Li2023OneShotHT,ma2023otavatar,yu2023nofa,Chu2024GPAvatarGA,Deng2023Portrait4DLO,Deng2024Portrait4Dv2PM} achieve high-quality reconstruction but are computationally expensive, limiting their use in real-time applications. Recent methods~\cite{Chu2024GeneralizableAA,He2025LAMLA,Lyu2024FaceLiftLG} leverage 3DGS for real-time rendering and move toward feed-forward architectures by incorporating priors from foundation models~\cite{Oquab2023DINOv2LR,khirodkar2024sapiens}, while diffusion-based methods further address missing observations by generating additional facial views~\cite{taubner2025cap4d,taubner2025mvp4d}. However, these methods still face challenges in either explicitly completing unobserved regions or maintaining view consistency, limiting their generalization to novel viewpoints under large pose variations and their ability to preserve identity.

To address these limitations, we present SInGA, a feed-forward framework for one-shot animatable Gaussian head modeling. Given a single image, our framework reconstructs an animatable 3D Gaussian head avatar and supports real-time expression control and rendering, as shown in Fig.~\ref{fig:teaser}. A key challenge is to reconstruct a complete Gaussian head from a single image, as parts of the head remain unobserved without multi-view supervision. Specifically, we employ a FLAME UV map to establish a structured representation of the head surface, enabling spatially coherent Gaussian prediction in a canonical 2D parameterization while preserving surface correspondence. We then extract image features from the source image and map them onto this representation to obtain UV-aligned feature maps. The structured UV layout and facial symmetry are further used to complete features in unobserved regions. The completed feature map is used to regress Gaussian attributes, effectively performing Gaussian inpainting for missing regions. Additionally, we introduce a Gaussian stacking strategy that assigns multiple Gaussians to each selected UV location to improve visual fidelity. This allows the model to allocate more capacity to regions requiring fine details than a uniform per-pixel representation. The predicted Gaussians are rendered via splatting and further refined by a neural renderer. The resulting Gaussian head avatar can be directly animated using standard linear blend skinning. Experiments demonstrate that our method achieves competitive or better performance compared with previous approaches in reconstruction quality and expression accuracy while enabling real-time avatar animation.

Our main contributions are as follows:
{
\renewcommand{\labelitemi}{$\bullet$}
\begin{itemize}
\item We introduce SInGA, a feed-forward framework for one-shot animatable Gaussian head modeling that reconstructs complete Gaussian head avatars from a single image.
\item We propose a surface-aligned representation that integrates UV-aligned feature mapping, symmetry-guided completion, and Gaussian stacking for high-fidelity reconstruction.
\item Extensive experiments demonstrate that our method achieves competitive or better visual quality compared with previous approaches while enabling real-time avatar animation.
\end{itemize}
}

\section{Related Work}
\label{rw}
\subsubsection{Generalizable One-shot Head Avatar.} Creating animatable head avatars has progressed from per-identity optimization to generalizable one-shot frameworks that reconstruct avatars from a single image. Although per-identity methods~\cite{NeRFace,Park2020NerfiesDN,Kirschstein2023NeRSembleMR,Zhao2023HAvatarHH,Zhang2024LearningDT} based on NeRF~\cite{NeRF} achieve high-quality reconstruction, they remain computationally intensive in inference, restricting their applicability in interactive or real-time scenarios. The introduction of 3DGS~\cite{Kerbl20233DGS} has led to a series of methods~\cite{Qian2023GaussianAvatarsPH,Xu2024GaussianHA,Xiang2023FlashAvatarHH,Kirschstein2025Avat3rLA,Tang2024GAFGA} that improve rendering speed and fidelity. However, these approaches remain identity-specific, often requiring monocular or multi-view inputs. Motivated by these limitations, recent works have explored generalizable models capable of creating avatars from a single image. Several works~\cite{Chu2024GPAvatarGA,Deng2023Portrait4DLO,Deng2024Portrait4Dv2PM,Li2023OneShotHT,Ma2023CVTHeadOC} leverage NeRF, tri-plane attention, and vertex-feature transformers to enable one-shot head reconstruction, but are still bounded by computational cost. More recently, 3DGS-based methods have emerged~\cite{Chu2024GeneralizableAA,He2025LAMLA,Lyu2024FaceLiftLG,Zhang2025GUAVAGU}. GAGAvatar~\cite{Chu2024GeneralizableAA} uses dual-lifting to model a 3D Gaussian head representation, and LAM~\cite{He2025LAMLA} predicts Gaussian attributes by querying image features using a transformer. However, neither method explicitly handles unobserved regions. In contrast, our method explicitly assigns identity-specific features to unobserved regions, enabling Gaussian inpainting and reducing identity inconsistency. 

\subsection{UV-Space Representation}
UV-space representations map 3D surfaces onto 2D domains to encode surface attributes as regular image grids, as in geometry images~\cite{gu2002geometry} that store 3D surface positions. These representations enable consistent correspondence and efficient processing with standard 2D architectures. Prior works explore UV-space representations for various vision tasks, including 3D reconstruction~\cite{Yan2024AnOI}, face completion~\cite{3DFaceFill}, generative modeling~\cite{Alhaija2022XDGANM3}, and surface representation learning~\cite{Elizarov2024GeometryID,wang2025partuv}. Recently, these representations have also been explored for 3D avatar reconstruction~\cite{kwon2024generalizable,Yan2024GaussianDC,sun2025svg}. GHG~\cite{kwon2024generalizable} formulates 3D Gaussian parameter optimization in a 2D UV space defined by a human template and incorporates a 2D inpainting module to handle missing areas. SVG-Head~\cite{sun2025svg} and GaussianDejavu~\cite{Yan2024GaussianDC} adopt FLAME~\cite{Li2017LearningAM} to parameterize 3D Gaussians into a mesh-aware UV representation, but do not explicitly model identity-consistent completion for unseen regions. To address these limitations, our method uses the consistent surface correspondences provided by the UV parameterization of FLAME to assign identity-specific semantic features to unobserved regions.

\subsection{Foundation Model as Prior}
Foundation models~\cite{Oquab2023DINOv2LR,simeoni2025dinov3,Radford2021LearningTV} pretrained on large-scale datasets have recently demonstrated strong capabilities in learning robust and semantically meaningful feature representations. These models have been widely adopted across various vision tasks, including image classification~\cite{Zhu2025InterpretableIC}, segmentation~\cite{Jose2024DINOv2MT}, dense correspondence estimation~\cite{Tumanyan2024DINOTrackerTD}, and 3D vision tasks~\cite{Knaebel2025DINOIT,ElBanani2024ProbingT3,Lin2024BIP3DB2}, due to their strong generalization and semantic understanding. They have also been applied in 3D avatar generation~\cite{Chu2024GeneralizableAA,He2025LAMLA,Zhang2025GUAVAGU,ma2025goes}, where their rich semantic features help improve robustness under diverse conditions. In our work, we follow prior approaches and adopt DINOv2~\cite{Oquab2023DINOv2LR} as a semantic prior for identity-aware feature representation.

\section{Method}
\label{method}
\subsubsection{Overview.}
\begin{figure}[tb]
  \centering
  \includegraphics[width=1.0\textwidth]{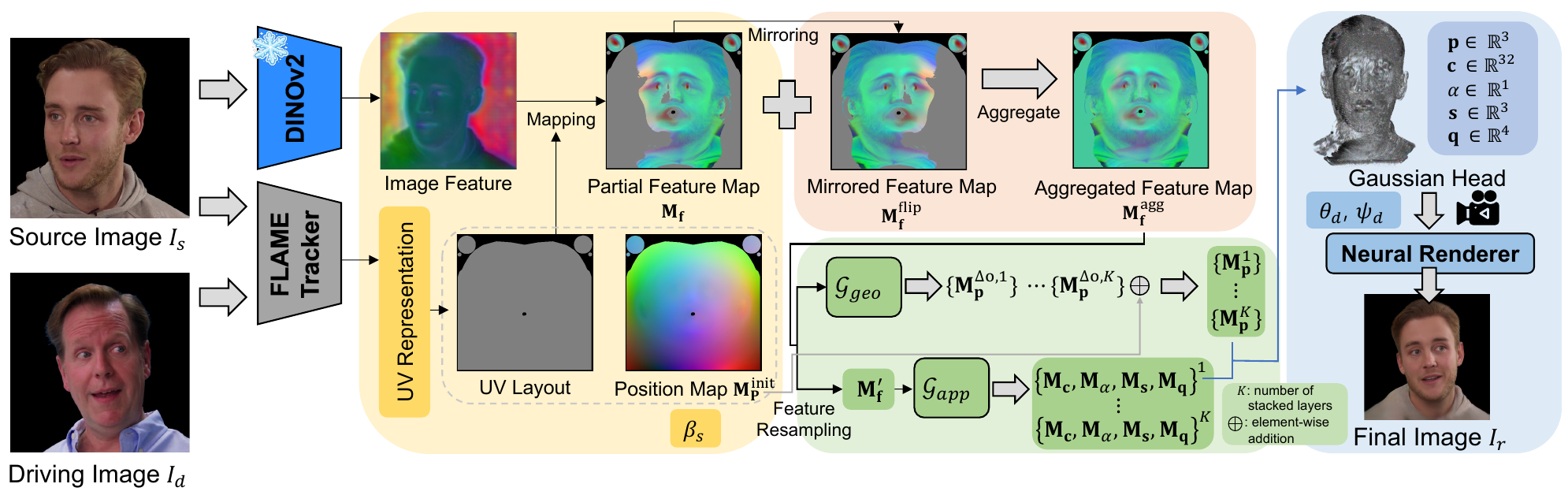}
  \caption{\textbf{Overview of SInGA}. Our framework completes unobserved regions in UV space using DINOv2 features as semantic priors to preserve identity-aware information. Given a source image and a driving signal, we represent Gaussians in the UV domain, where the fixed UV topology provides structured connectivity across facial regions. We formulate the completion of unobserved regions as semantic inpainting and fill in missing features using DINOv2 priors. Gaussian attributes are then predicted from the completed representation to construct the Gaussian head avatar. $\beta_s$, $\psi_d$, and $\theta_d$ denote the source shape, driving expression, and driving pose parameters, respectively. Note that during training, the source and driving images are sampled from the same identity.}
  \label{fig:framework}
  \vspace{-5mm}
\end{figure}

Given a source image, our goal is to generate an animatable head avatar represented by a set of Gaussians $\smash{\mathbf{G} = \{ \mathbf{p}, \mathbf{c}, \alpha, \mathbf{s}, \mathbf{q} }\}$, where $\mathbf{p} \in \mathbb{R}^3$ denotes the 3D position, $\mathbf{c} \in \mathbb{R}^{32}$ the color coding, $\alpha \in \mathbb{R}$ the opacity, $\mathbf{s} \in \mathbb{R}^3$ the scale, and $\mathbf{q} \in \mathbb{R}^4$ the rotation quaternion. An overview of our method is shown in Fig.~\ref{fig:framework}. We first map image features into a UV feature map aligned with the head surface (Sec.~\ref{featuv}). We then introduce an identity-aware facial inpainting module that exploits the structured connectivity of the UV domain (Sec.~\ref{inpaint}). Gaussian attributes are subsequently predicted in the UV domain to construct the head representation (Sec.~\ref{gauspred}). Finally, we animate the reconstructed Gaussians using driving FLAME parameters (Sec.~\ref{Reenactment}) and optimize the model by minimizing the difference between the rendered and target images (Sec.~\ref{loss}).

\subsection{Feature Mapping in UV Space}
\label{featuv}
\subsubsection{UV Space Representation.}
We represent the head surface in a canonical UV space defined over a fixed FLAME topology~\cite{Li2017LearningAM}. To better align the resolution of the UV parameterization with the underlying surface geometry, we refine the canonical mesh using two iterations of Loop subdivision~\cite{stam2003quad}. The original mesh contains $5{,}023$ vertices and $9{,}976$ faces, which are increased to $79{,}936$ vertices and $159{,}616$ faces after subdivision. This subdivision better matches the surface discretization to the $512 \times 512$ UV grid, avoiding coarse piecewise-planar approximations caused by large triangles and improving the stability of surface normal directions. We then construct a FLAME UV position map by assigning each UV coordinate to a corresponding 3D surface point via barycentric interpolation within its associated triangle. Each valid UV pixel corresponds to a fixed location on the canonical surface, providing a consistent mapping between the UV domain and the underlying geometry. The positions stored in this map initialize the Gaussians and are further refined by per-pixel offsets along the surface normals.
\vspace{-4mm}
\subsubsection{UV Feature Representation.}
We represent image features as a UV feature map aligned with the head surface. Given a source image $\mathbf{I}_s$, we extract image features using a frozen DINOv2~\cite{Oquab2023DINOv2LR} encoder. For feature sampling, we reuse the triangle index and barycentric coordinates associated with each valid UV pixel on the source FLAME mesh deformed by the source parameters, obtaining the corresponding 3D surface point. The resulting 3D points are then projected onto the source image plane using the source camera parameters, and image features are sampled at the corresponding projected locations. We determine UV visibility using z-buffering and assign sampled features only to pixels visible from the source view, yielding a partial UV feature map aligned with the head surface. The remaining unobserved regions are completed by the identity-aware facial inpainting module described in Sec.~\ref{inpaint}. In total, $N = 234{,}888$ valid UV pixels define the set of Gaussians.

\subsection{Identity-Consistent Inpainting}
\label{inpaint}
\subsubsection{Symmetry-Guided Completion.}
The UV feature map constructed from the source image is partial, as features are only available for regions visible in the source view. To complete missing regions, we exploit bilateral symmetry of the head geometry together with the structured UV layout. Let $\mathbf{M}_{\mathrm{f}} \in \mathbb{R}^{H \times W \times C}$ denote the partial UV feature map. For each valid UV pixel $p$, we mirror its corresponding canonical 3D surface point across the symmetry plane $x=0$. Using a KD tree built from the canonical 3D positions of valid UV pixels, we find the nearest point to the mirrored position and use its UV coordinate as $\phi_{\mathrm{geo}}(p)$. This correspondence is precomputed on the canonical surface. This mapping defines a geometry-aware symmetry correspondence in the UV domain. Using this correspondence, we construct a mirrored feature map by sampling features from the symmetric UV locations:

\begin{equation}
\mathbf{M}_{\mathrm{f}}^{\mathrm{flip}}(p) =
\mathcal{B}\left(
\mathbf{M}_{\mathrm{f}}, \phi_{\mathrm{geo}}(p)
\right),
\end{equation}
where $\mathcal{B}$ denotes bilinear sampling. Thus, $\mathbf{M}_{\mathrm{f}}^{\mathrm{flip}}$ provides features sampled from the opposite side of the face based on the canonical 3D geometry, offering complementary information for missing regions in $\mathbf{M}_{\mathrm{f}}$.

\subsubsection{Confidence-Aware Aggregation.}
\begin{wrapfigure}{r}{0.5\linewidth}
\centering
\vspace{-2.0em}
\scalebox{0.95}{%
\includegraphics[width=\linewidth]{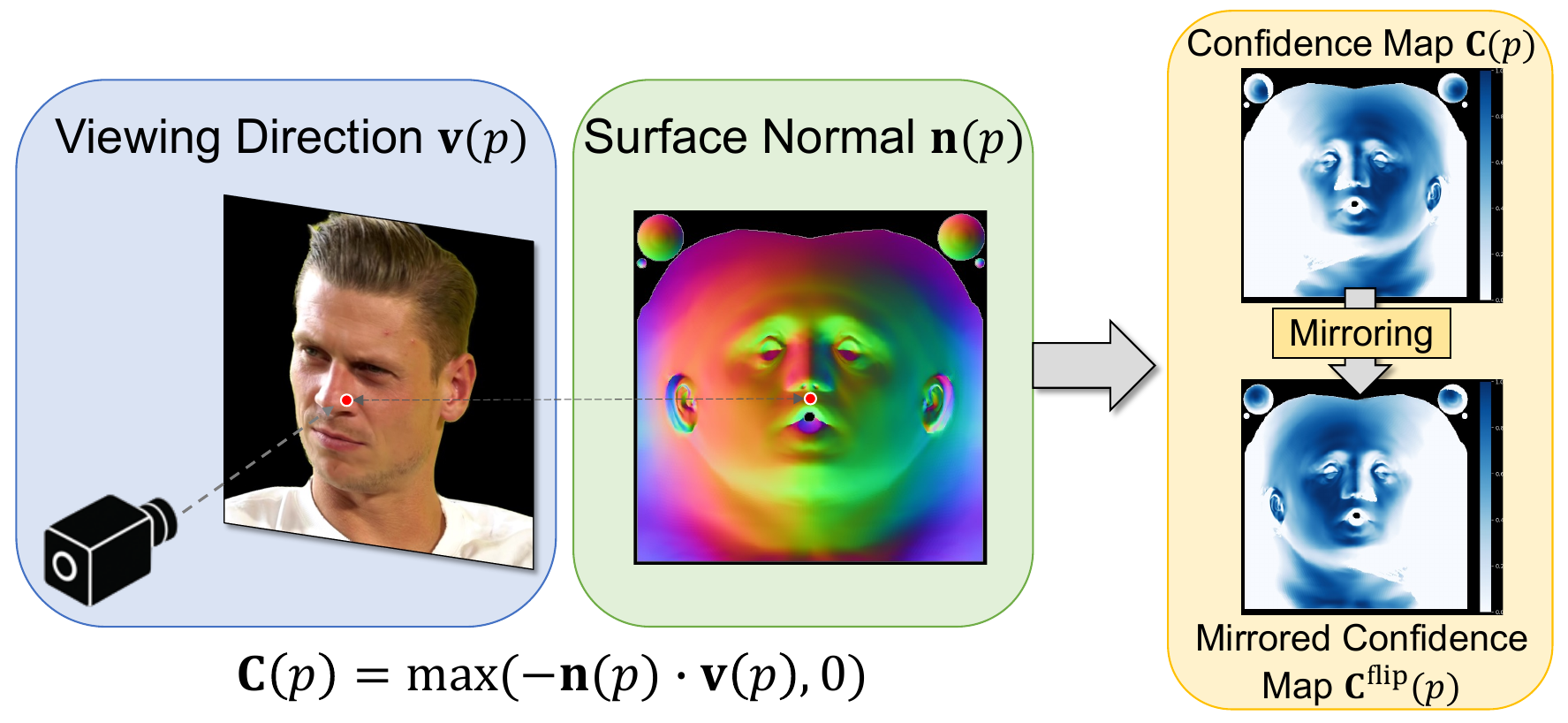}
}
\caption{\textbf{Confidence map.} We compute a confidence map from the source view to identify well-observed and uncertain UV regions. High confidence is assigned to surface regions facing the source camera, while low confidence appears in unobserved regions and near silhouettes.}
\label{fig:confidence}
\vspace{-1.0em}
\end{wrapfigure}
Although symmetry-guided completion fills in features for the unobserved side of the head, simply inserting the mirrored features into missing regions can introduce artifacts, especially near silhouettes and regions observed under grazing angles. To address this, we estimate a visibility confidence for each UV pixel as illustrated in Fig.~\ref{fig:confidence}, and use it to softly aggregate the original and mirrored feature maps. Specifically, we compute the surface normal $\mathbf{n}(p)$ by applying finite differences to the positions stored in the UV position map, and estimate the confidence map $\mathbf{C}(p)$ from its dot product with the viewing direction $\mathbf{v}(p)$:
\begin{equation}
\mathbf{C}(p) = \max \left( -\mathbf{n}(p) \cdot \mathbf{v}(p), 0 \right).
\end{equation}
Surface points facing the source camera have high confidence, while pixels near the silhouette or under grazing angles receive low confidence. We compute the same confidence map for the mirrored feature map, denoted as $\mathbf{C}^{\mathrm{flip}}(p)$. The blended UV feature map is then obtained by confidence-weighted aggregation:
\begin{equation}
\label{eq:3}
\mathbf{M}_{\mathrm{f}}^{\mathrm{blend}}(p)
=
\frac{
\mathbf{C}(p)\mathbf{M}_{\mathrm{f}}(p)
+
\mathbf{C}^{\mathrm{flip}}(p)\mathbf{M}_{\mathrm{f}}^{\mathrm{flip}}(p)
}{
\mathbf{C}(p)+\mathbf{C}^{\mathrm{flip}}(p)
}.
\end{equation}
This allows the model to preserve reliable observations from the source view while incorporating symmetric features in weakly observed regions. However, directly using the blended feature map may compromise reliable source features. We therefore softly preserve the original UV feature using source view confidence, allowing high confidence source view DINOv2 features to dominate the aggregation. Additionally, some UV pixels may remain unfilled even after symmetry-based completion, since their symmetric counterparts can also be unobserved in the source view. We use learnable parameters to fill the remaining regions and define the aggregated UV feature map as:
\begin{equation}
\mathbf{M}_{\mathrm{f}}^{\mathrm{agg}}(p)
=
\begin{cases}
\begin{aligned}
&\mathbf{C}(p)\mathbf{M}_{\mathrm{f}}(p) \\
&\quad+
\mathbf{C}^{\mathrm{flip}}(p)
\mathbf{M}_{\mathrm{f}}^{\mathrm{blend}}(p),
\end{aligned}
& p \in \Omega_{\mathrm{filled}}, \\[6pt]
\mathbf{P}_{\mathrm{f}}(p),
& p \in \Omega_{\mathrm{unfilled}} .
\end{cases}
\end{equation}
where $\Omega_{\mathrm{filled}}$ denotes valid UV pixels with features obtainable from the source view or their symmetric counterparts, $\Omega_{\mathrm{unfilled}}$ denotes UV pixels remaining unfilled after symmetry-guided completion, and $\mathbf{P}_{\mathrm{f}}$ denotes learnable UV feature parameters. When both confidence values are zero, we skip Eq.~\ref{eq:3} and assign $\mathbf{P}_{\mathrm{f}}(p)$ to $\mathbf{M}_{\mathrm{f}}^{\mathrm{agg}}(p)$. The resulting feature map $\mathbf{M}_{\mathrm{f}}^{\mathrm{agg}}$ is used for Gaussian prediction.

\subsection{Region-Aware Gaussian Stacking}
\label{stack}
\begin{wrapfigure}{r}{0.5\linewidth}
\centering
\vspace{-2.0em}
\scalebox{0.95}{%
\includegraphics[width=\linewidth]{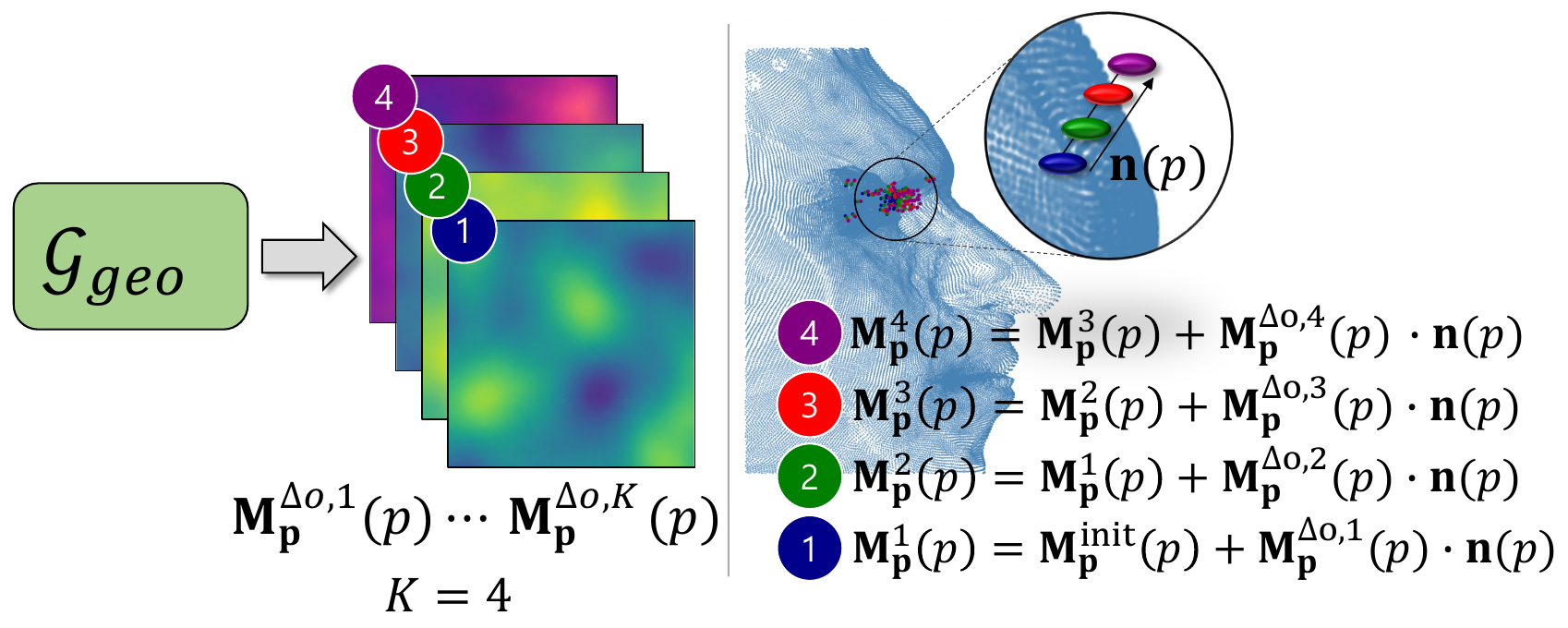}
}
\caption{\textbf{Gaussian stacking.} For selected regions, we stack $K$ Gaussians along the surface normal by cumulatively adding the predicted position offsets, increasing local geometric detail. The figure illustrates the case of $K=4$.}
\label{fig:stacking}
\vspace{-1.0em}
\end{wrapfigure}
While assigning a single Gaussian to each UV pixel provides an efficient surface-aligned representation, it can be insufficient for geometrically complex regions, where fine structures or multiple nearby surfaces must be represented. To increase local representational capacity without densely increasing the number of Gaussians over the entire head, we introduce a region-aware Gaussian stacking strategy. Let $\Omega_{\mathrm{base}}$ denote the initial set of base Gaussian locations, where one Gaussian is assigned to each valid UV pixel. We select a stacking region $\Omega_{\mathrm{stack}}\subset\Omega_{\mathrm{base}}$, and allocate $K-1$ additional Gaussians for each pixel in this region. To keep the Gaussian budget unchanged, we remove the same number of base Gaussians from relatively coarse regions. We denote the removed base Gaussian locations by $\Omega_{\mathrm{coarse}}$, where:
\begin{equation}
|\Omega_{\mathrm{coarse}}|
=
(K-1)|\Omega_{\mathrm{stack}}|.
\end{equation}
Given the number of Gaussian layers $K$, the final Gaussian index set is defined as:
\begin{equation}
\mathcal{I}_{\mathbf{G}}
=
\left\{(p,1)\mid p\in\Omega_{\mathrm{base}}\setminus\Omega_{\mathrm{coarse}}\right\}
\cup
\left\{(p,k)\mid p\in\Omega_{\mathrm{stack}},\; k=2,\ldots,K\right\}.
\end{equation}
To preserve spatial coverage in the subsampled coarse region, we adjust the scale of the retained Gaussians in that region as:
\begin{equation}
\mathbf{s}' = \frac{\mathbf{s}}{\sqrt{r}},
\end{equation}
where $r$ is the ratio of the number of retained Gaussians to the original number of Gaussians in the subsampled region. This strategy increases capacity in regions that require finer details while maintaining the overall Gaussian budget and preserving coverage in coarser regions.

\subsection{Gaussian Attribute Prediction}
\label{gauspred}
We extend the pixel-wise Gaussian representation to a layer-wise formulation in the UV map. Given the Gaussian index set $\mathcal{I}_{\mathbf{G}}$, each Gaussian is indexed by a UV coordinate $p$ and a layer index $k$. This allows each UV pixel to represent either a single Gaussian or multiple stacked Gaussians depending on its region. We define the Gaussian attributes as:
\begin{equation}
\mathbf{G}(p,k)
=
\left\{
\mathbf{M}_{\mathbf{p}}^{k}(p),
\mathbf{M}_{\mathbf{c}}^{k}(p),
\mathbf{M}_{\alpha}^{k}(p),
\mathbf{M}_{\mathbf{s}}^{k}(p),
\mathbf{M}_{\mathbf{q}}^{k}(p)
\right\},
\quad (p,k)\in\mathcal{I}_{\mathbf{G}} .
\end{equation}
Here, $\mathbf{M}_{\mathbf{p}}^{k}$, $\mathbf{M}_{\mathbf{c}}^{k}$, $\mathbf{M}_{\alpha}^{k}$, $\mathbf{M}_{\mathbf{s}}^{k}$, and $\mathbf{M}_{\mathbf{q}}^{k}$ denote the layer-wise maps for position, color coding, opacity, scaling, and rotation. We first compute the Gaussian positions by predicting a scalar displacement along the surface normal for each layer. Specifically, the geometry network $\mathcal{G}_{\mathrm{geo}}$ takes the aggregated UV feature map $\mathbf{M}_{\mathrm{f}}^{\mathrm{agg}}$ as input and predicts a layer-wise offset map:
\begin{equation}
\mathbf{M}_{\mathbf{p}}^{\Delta \mathbf{o},k}(p)
=
\mathcal{G}_{\mathrm{geo}}\!\left(\mathbf{M}_{\mathrm{f}}^{\mathrm{agg}}\right)^{k}(p),
\end{equation}
where $\mathbf{M}_{\mathbf{p}}^{\Delta \mathbf{o},k}(p) \in \mathbb{R}$ denotes the offset magnitude for the $k$-th Gaussian layer at UV coordinate $p$. Given the initial UV position map $\mathbf{M}_{\mathbf{p}}^{\mathrm{init}}(p)$ and the corresponding surface normal $\mathbf{n}(p)$, the position of each Gaussian is computed as:
\begin{equation}
\mathbf{M}_{\mathbf{p}}^{k}(p)
=
\begin{cases}
\mathbf{M}_{\mathbf{p}}^{\mathrm{init}}(p)
+
\mathbf{M}_{\mathbf{p}}^{\Delta \mathbf{o},1}(p)\cdot \mathbf{n}(p),
& (k=1), \\[1mm]
\mathbf{M}_{\mathbf{p}}^{k-1}(p)
+
\mathbf{M}_{\mathbf{p}}^{\Delta \mathbf{o},k}(p)\cdot \mathbf{n}(p)
& (k=2,\dots,K).
\end{cases}
\end{equation}
For stacked Gaussians, the additional layers are placed outward from the base layer along the surface normal. Using the updated 3D positions of the base Gaussian layer ($k=1$), we resample image features from the source view. Specifically, we project the updated positions onto the source image plane, sample the corresponding image features, and apply the same confidence-aware aggregation described in  Sec.~\ref{inpaint}. This predicts the updated UV feature map $\mathbf{M}_{\mathrm{f}}^{\prime}$ for appearance prediction. The appearance network $\mathcal{G}_{\mathrm{app}}$ then predicts the remaining Gaussian attributes for each layer:
\begin{equation}
\left\{
\mathbf{M}_{\mathbf{c}}^{k}(p),
\mathbf{M}_{\alpha}^{k}(p),
\mathbf{M}_{\mathbf{s}}^{k}(p),
\mathbf{M}_{\mathbf{q}}^{k}(p)
\right\}
=
\mathcal{G}_{\mathrm{app}}\!\left(\mathbf{M}_{\mathrm{f}}^{\prime}\right)^{k}(p).
\end{equation}
The resulting Gaussian attribute maps are used to construct the Gaussian head representation.

\subsection{Reenactment and Refinement}
\label{Reenactment}
After reconstructing the Gaussian representation, our framework enables efficient reenactment by driving the Gaussians with the motion of a target sequence. Specifically, we use FLAME parameters to transfer the driving motion while preserving the source identity. Given the source shape parameter $\beta_s$, and the driving pose and expression parameters $\theta_d$ and $\psi_d$, we construct the driving geometry with the source identity and target motion. During reenactment, only the Gaussian position map $\mathbf{M}_{\mathbf{p}}^{k}(p)$ needs to be updated. Additionally, we follow prior works~\cite{Chu2024GPAvatarGA,Chu2024GeneralizableAA} and adopt a refinement stage to improve the final rendering quality. Rather than rendering only RGB colors, we render $\mathbf{M}_{\mathbf{c}}^{k}(p) \in \mathbb{R}^{32}$, where the first three channels correspond to a coarse RGB image $I_c$ and the remaining channels provide additional features. The rendered feature map is then processed by a UNet-based~\cite{wang2021towards} neural renderer to output the final refined image $I_r$.

\subsection{Training Strategy and Optimization}
\label{loss}
During training, we randomly sample two frames from the same video, using one as the source image $I_s$ and the other as the driving (target) image $I_d$. The network is trained to generate an output image that matches the appearance and motion of the target. We supervise both the coarse image $I_c$ and the refined image $I_r$ with an $\mathrm{L}_1$ reconstruction loss and a perceptual loss:
\begin{equation}
\mathcal{L}_{\mathrm{l1}} = \| I_{\mathrm{c}} - I_d \|_1 + \| I_{\mathrm{r}} - I_d \|_1 ,
\end{equation}
\begin{equation}
\mathcal{L}_{\mathrm{perc}} = \operatorname{LPIPS}(I_{\mathrm{c}}, I_d) + \operatorname{LPIPS}(I_{\mathrm{r}}, I_d).
\end{equation}
Additionally, we regularize the predicted offset $\Delta \mathbf{o}$ to stabilize the Gaussian geometry with mean, magnitude, and total variation penalties:
\begin{equation}
\mathcal{L}_{\mathrm{mean}} = \left\| \operatorname{avg}(\Delta \mathbf{o}) \right\|_2^2 ,
\end{equation}
\begin{equation}
\mathcal{L}_{\mathrm{mag}} = \left\| \Delta \mathbf{o} \right\|_2^2 ,
\end{equation}
\begin{equation}
\mathcal{L}_{\mathrm{tv}} = \left\| \nabla_x \Delta \mathbf{o} \right\|_1 + \left\| \nabla_y \Delta \mathbf{o} \right\|_1 ,
\end{equation}
where $\nabla_x$ and $\nabla_y$ denote finite differences along the horizontal and vertical directions, respectively. The overall training objective is defined as:
\begin{equation}
\begin{aligned}
\mathcal{L}_{\mathrm{total}} =
&\lambda_{\mathrm{l1}}\mathcal{L}_{\mathrm{l1}}
+
\lambda_{\mathrm{perc}}\mathcal{L}_{\mathrm{perc}}
+
\lambda_{\mathrm{mean}}\mathcal{L}_{\mathrm{mean}} \\
&+
\lambda_{\mathrm{mag}}\mathcal{L}_{\mathrm{mag}}
+
\lambda_{\mathrm{tv}}\mathcal{L}_{\mathrm{tv}} ,
\end{aligned}
\end{equation}
where $\lambda_{\mathrm{l1}} = 1.0$, $\lambda_{\mathrm{perc}}=0.01$, and $\lambda_{\mathrm{mean}} = \lambda_{\mathrm{mag}} = \lambda_{\mathrm{tv}} = 0.1$.

\begin{figure}[t]
\centering
\setlength{\tabcolsep}{0pt}
\renewcommand{\arraystretch}{0}
\resizebox{\textwidth}{!}{
\begin{tabular}{@{}cccccccc@{}}
\includegraphics[width=0.125\textwidth]{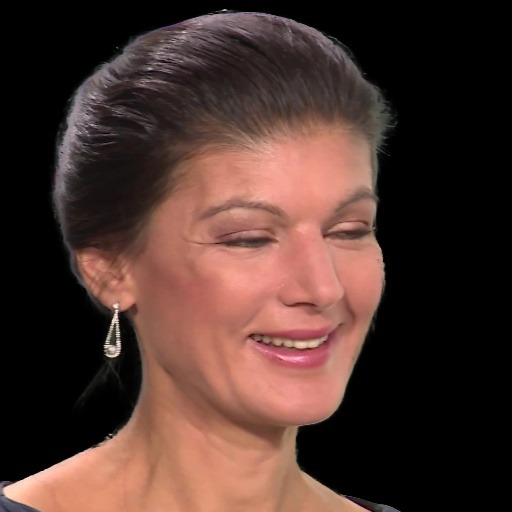} &
\includegraphics[width=0.125\textwidth]{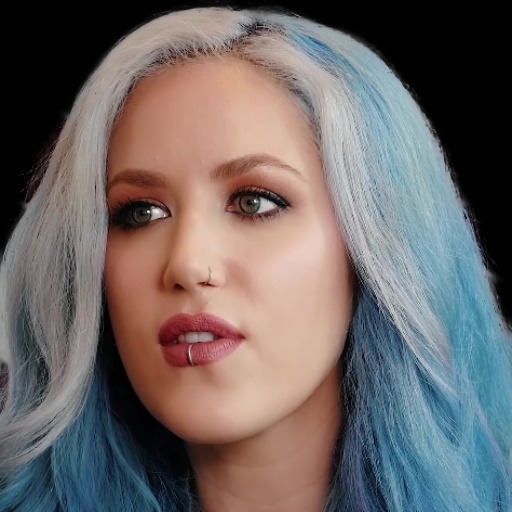} &
\includegraphics[width=0.125\textwidth]{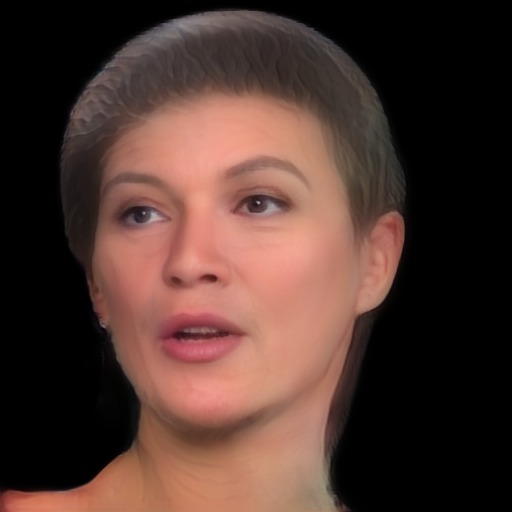} &
\includegraphics[width=0.125\textwidth]{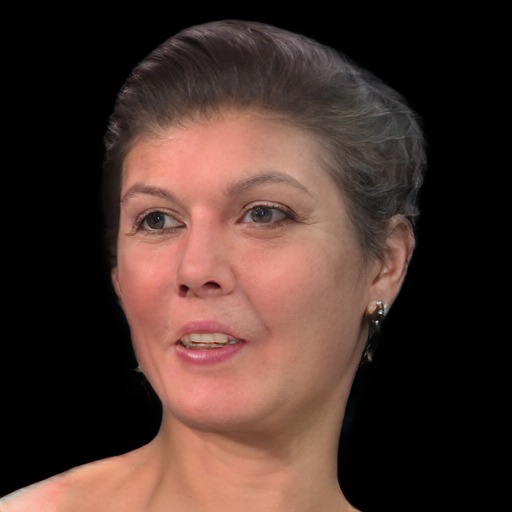} &
\includegraphics[width=0.125\textwidth]{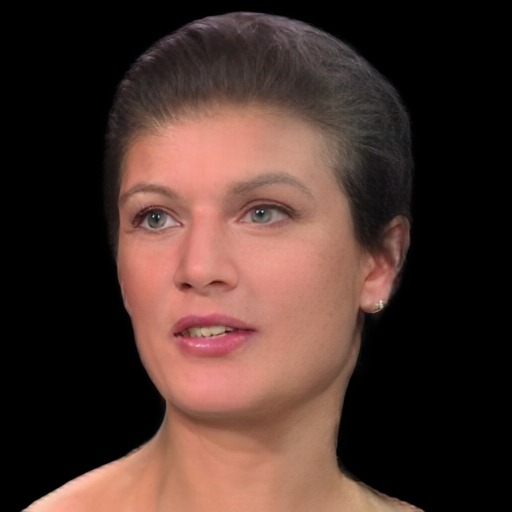} &
\includegraphics[width=0.125\textwidth]{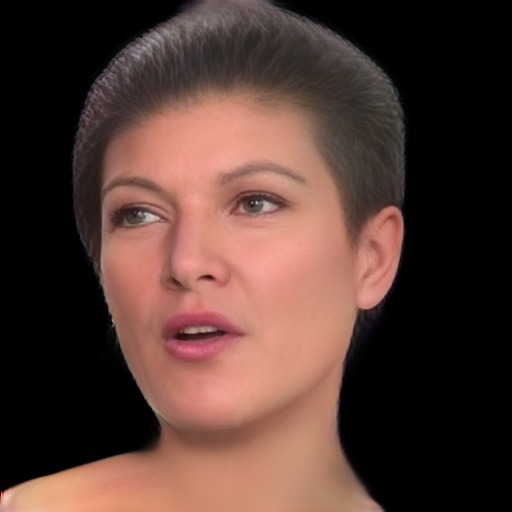} &
\includegraphics[width=0.125\textwidth]{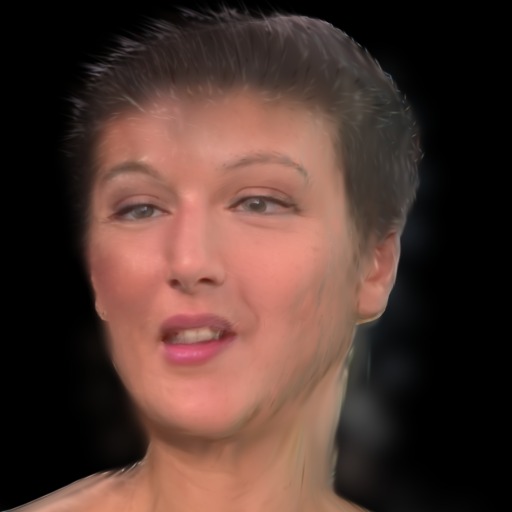} &
\includegraphics[width=0.125\textwidth]{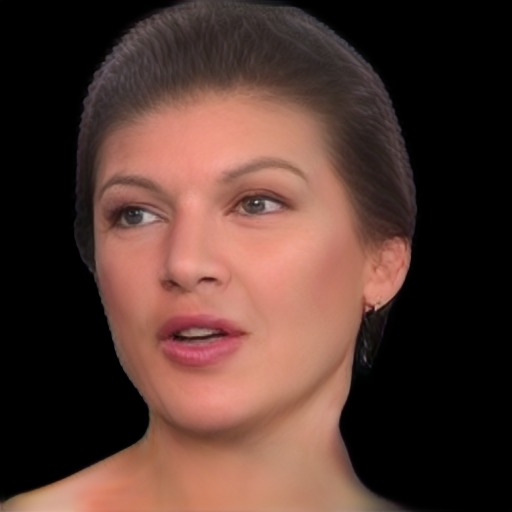}  \\
\includegraphics[width=0.125\textwidth]{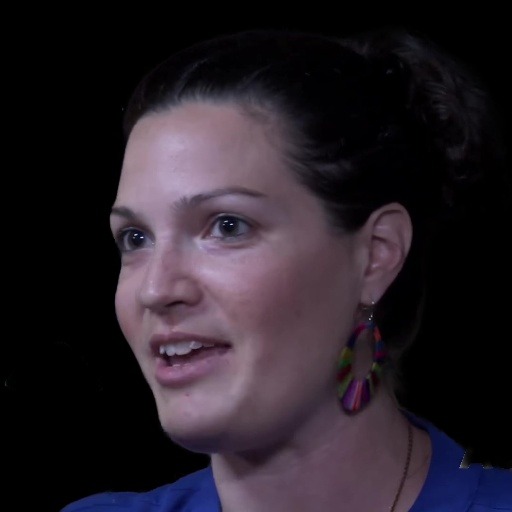} &
\includegraphics[width=0.125\textwidth]{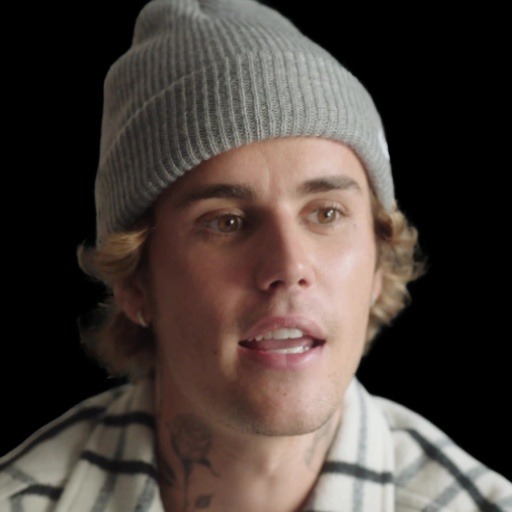} &
\includegraphics[width=0.125\textwidth]{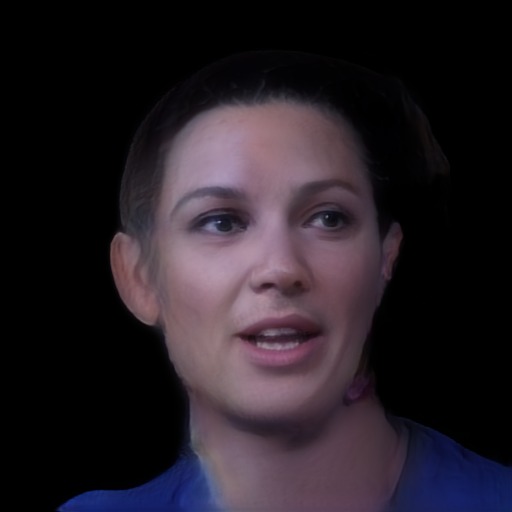} &
\includegraphics[width=0.125\textwidth]{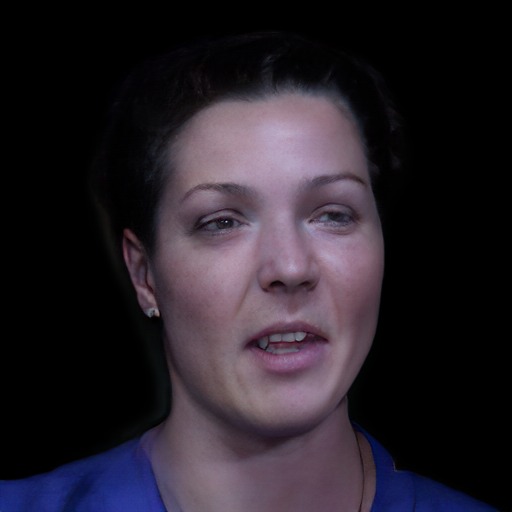} &
\includegraphics[width=0.125\textwidth]{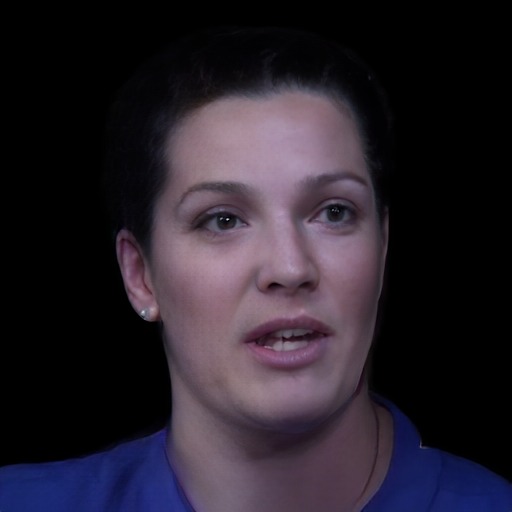} &
\includegraphics[width=0.125\textwidth]{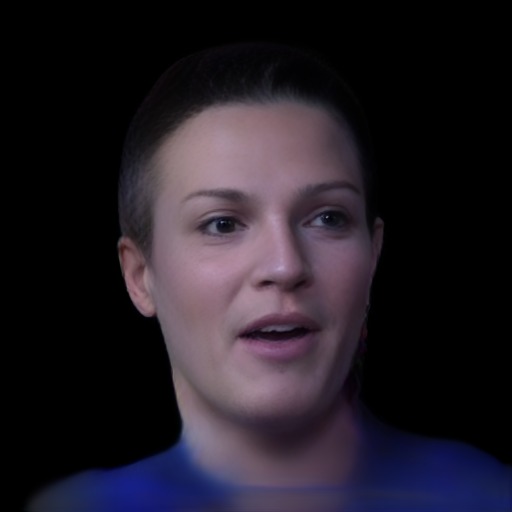} &
\includegraphics[width=0.125\textwidth]{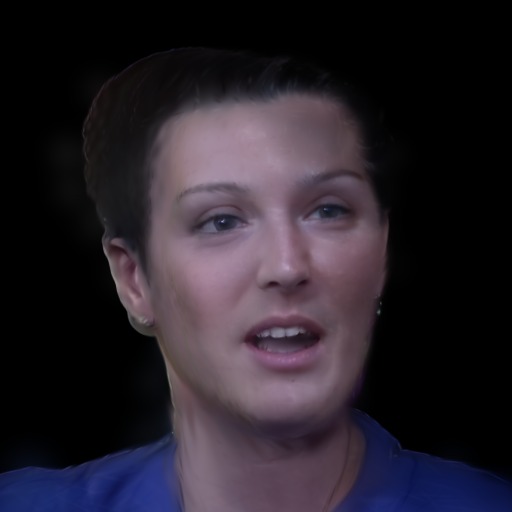} &
\includegraphics[width=0.125\textwidth]{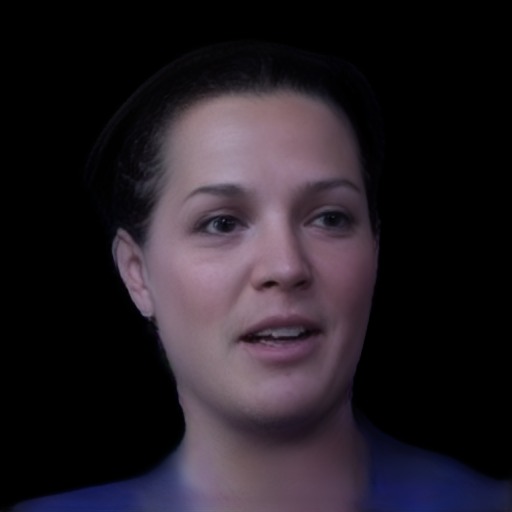}  \\
\includegraphics[width=0.125\textwidth]{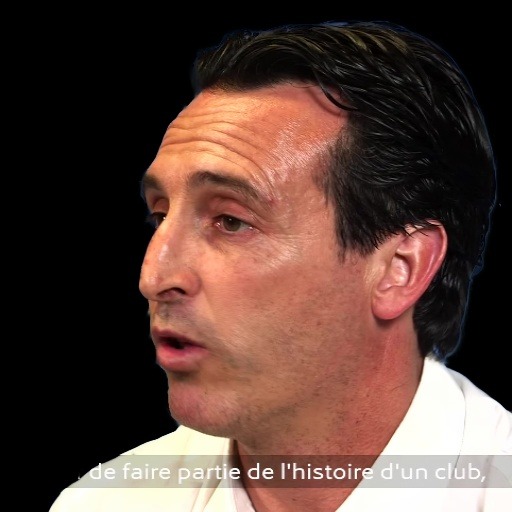} &
\includegraphics[width=0.125\textwidth]{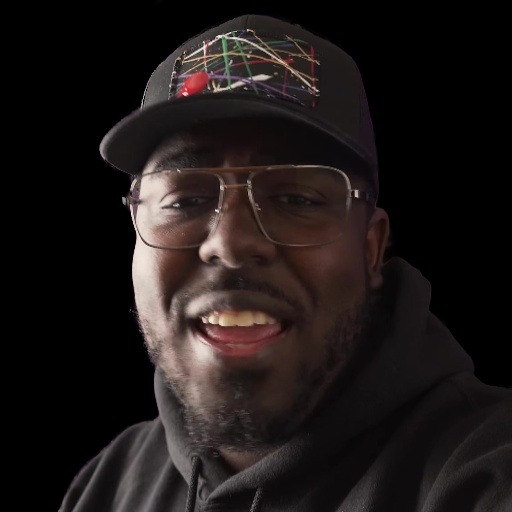} &
\includegraphics[width=0.125\textwidth]{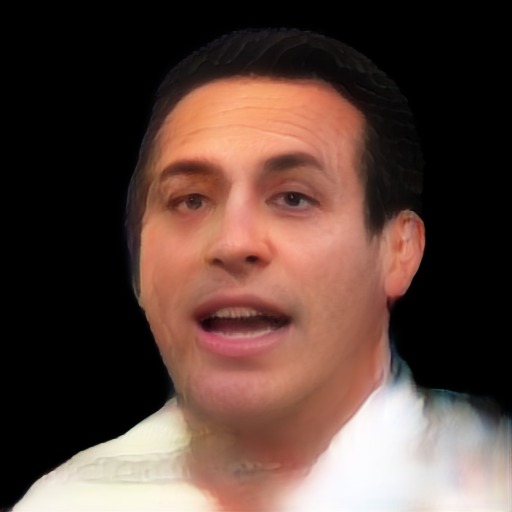} &
\includegraphics[width=0.125\textwidth]{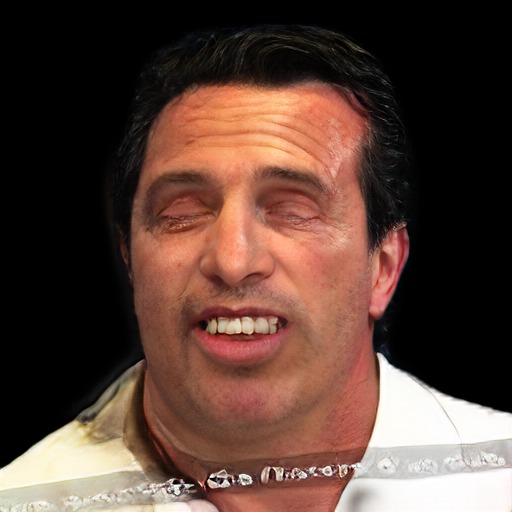} &
\includegraphics[width=0.125\textwidth]{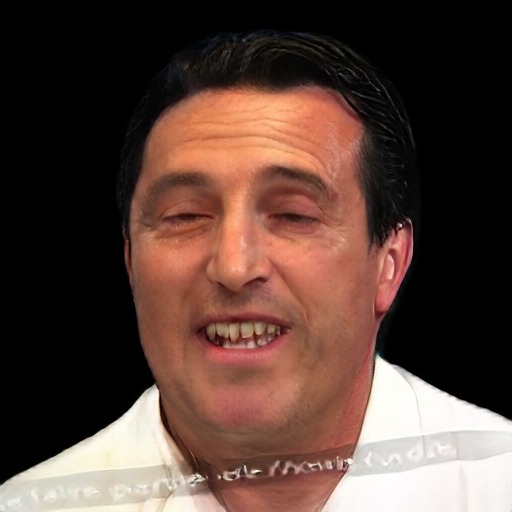} &
\includegraphics[width=0.125\textwidth]{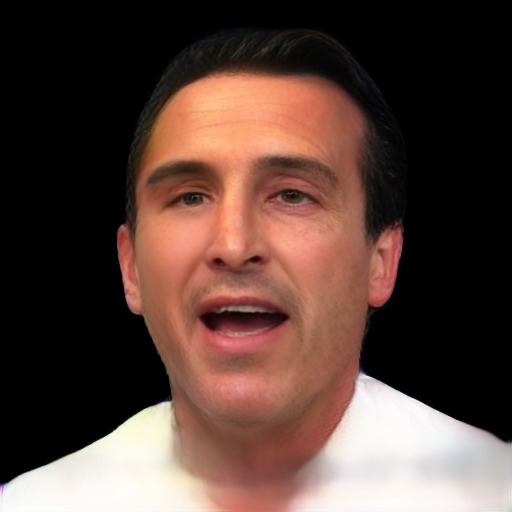} &
\includegraphics[width=0.125\textwidth]{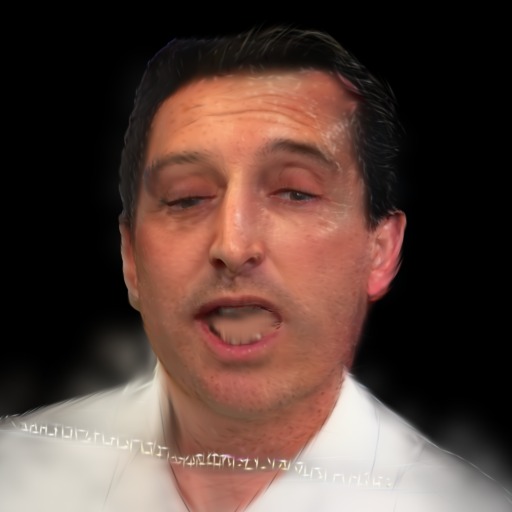} &
\includegraphics[width=0.125\textwidth]{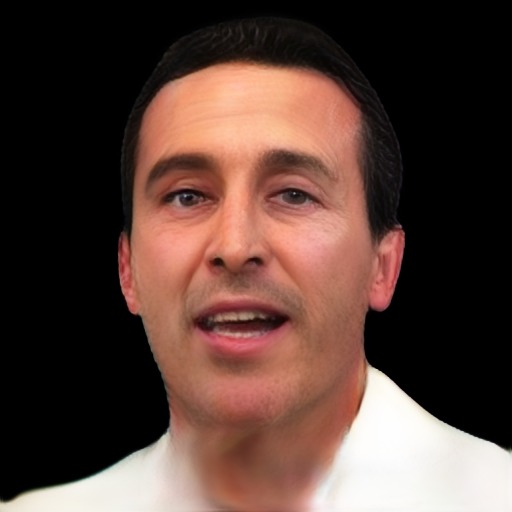}  \\
\includegraphics[width=0.125\textwidth]{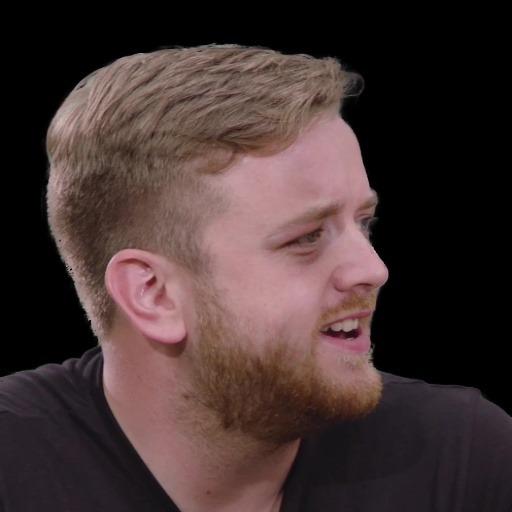} &
\includegraphics[width=0.125\textwidth]{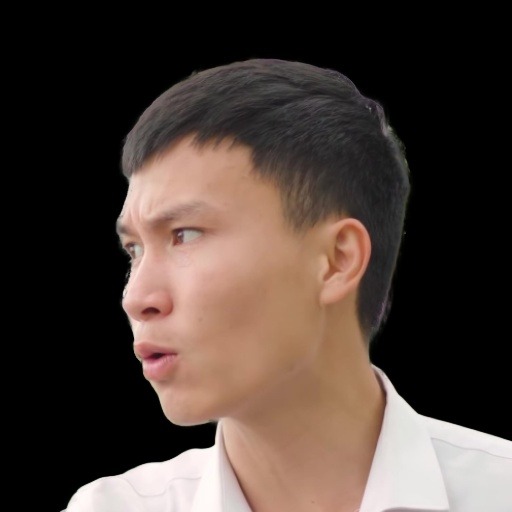} &
\includegraphics[width=0.125\textwidth]{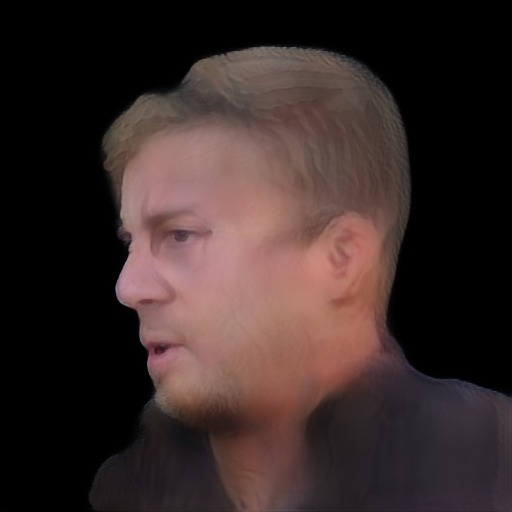} &
\includegraphics[width=0.125\textwidth]{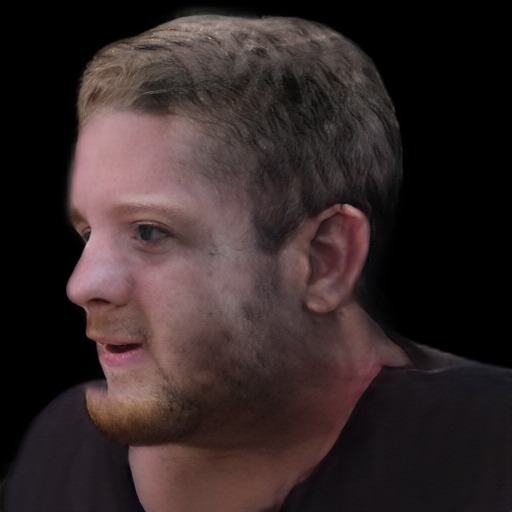} &
\includegraphics[width=0.125\textwidth]{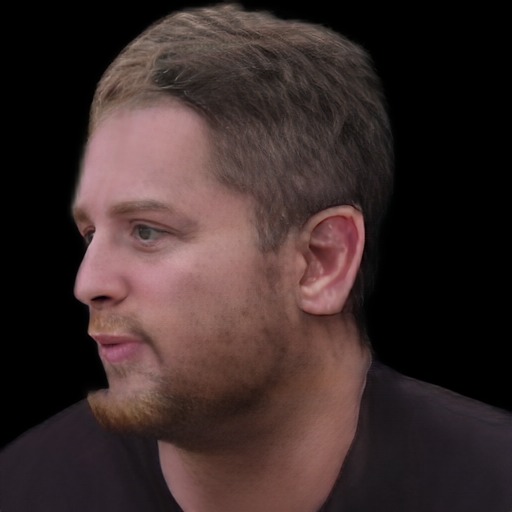} &
\includegraphics[width=0.125\textwidth]{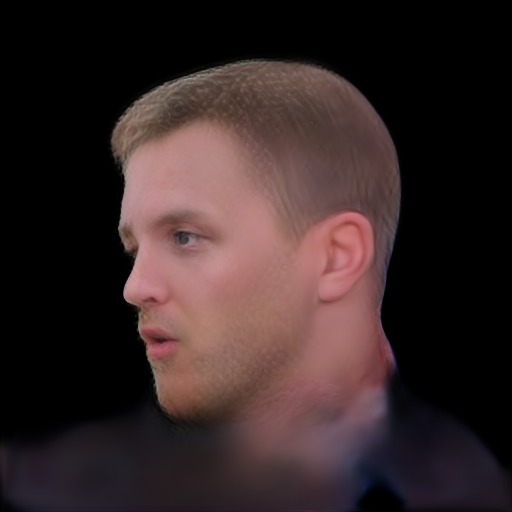} &
\includegraphics[width=0.125\textwidth]{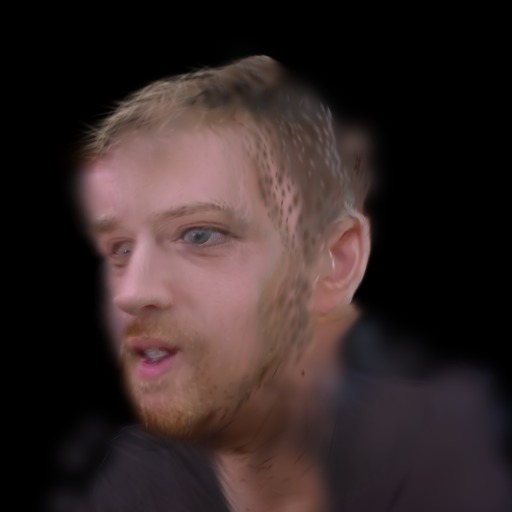} &
\includegraphics[width=0.125\textwidth]{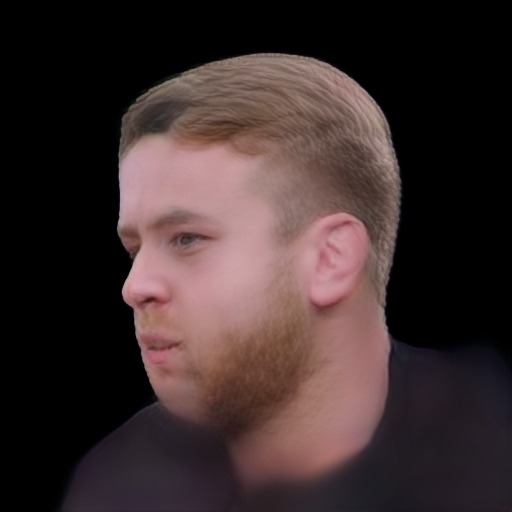}  \\
\includegraphics[width=0.125\textwidth]{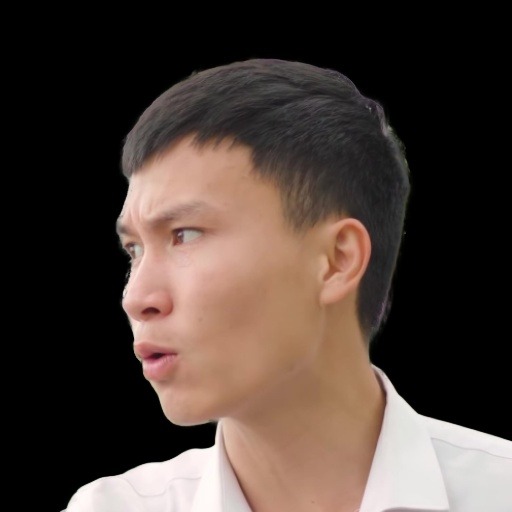} &
\includegraphics[width=0.125\textwidth]{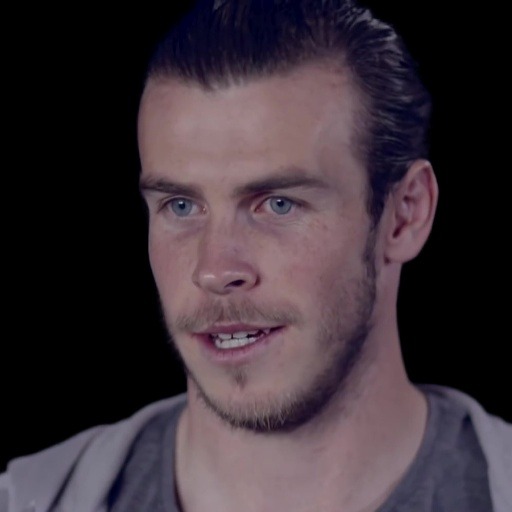} &
\includegraphics[width=0.125\textwidth]{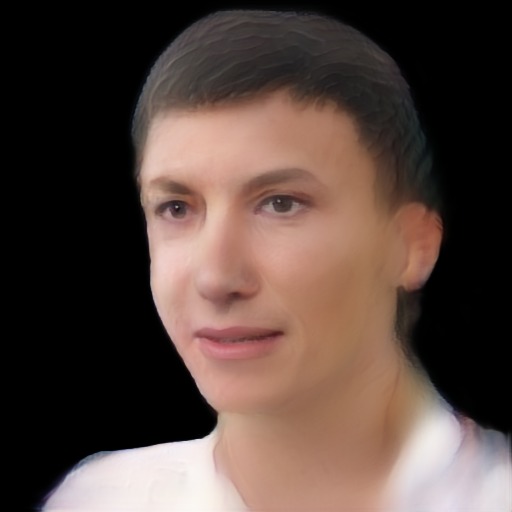} &
\includegraphics[width=0.125\textwidth]{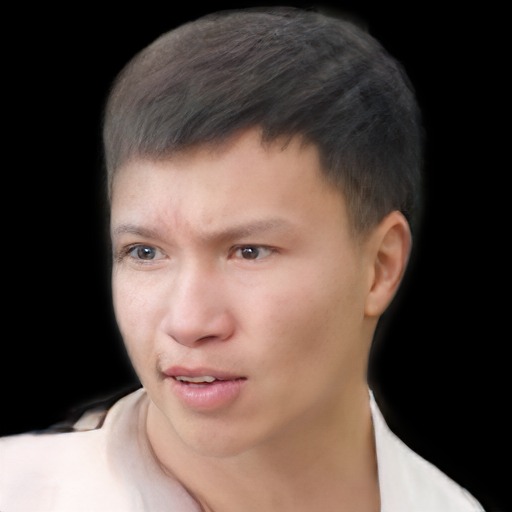} &
\includegraphics[width=0.125\textwidth]{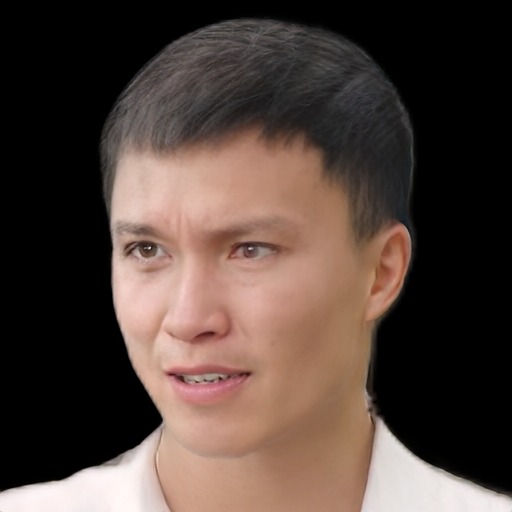} &
\includegraphics[width=0.125\textwidth]{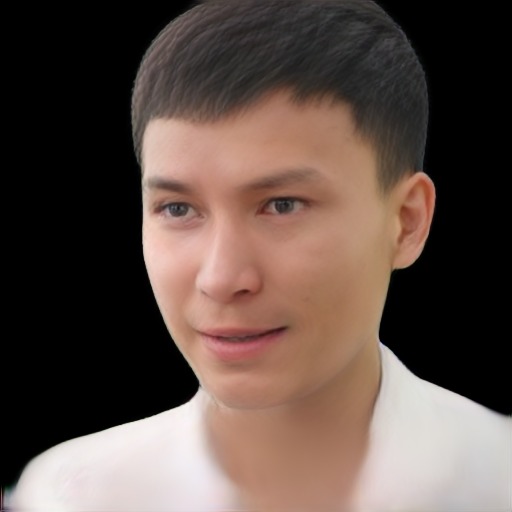} &
\includegraphics[width=0.125\textwidth]{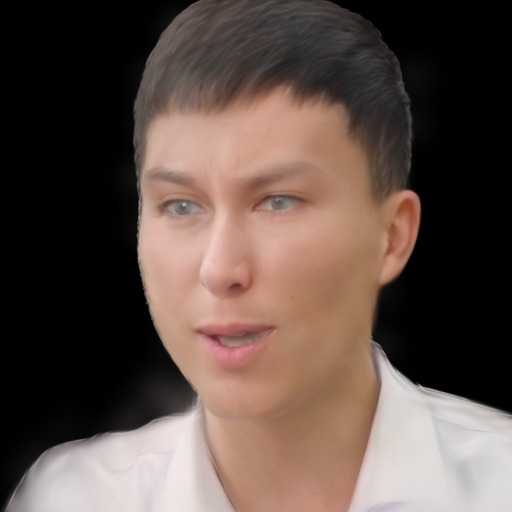} &
\includegraphics[width=0.125\textwidth]{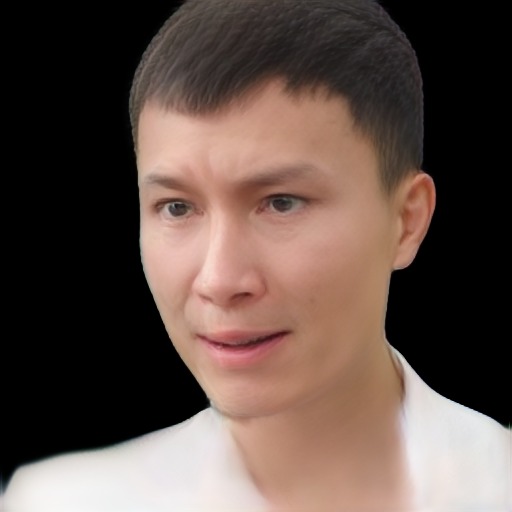}  \\[5pt]
Source & Driving & GPAvatar & P4D & P4D-v2 & GAG & LAM & Ours \\
\end{tabular}
}
\caption{\textbf{Qualitative comparisons.} We compare our method with existing methods in the cross reenactment setting.}
\label{fig:qual-cross}
\vspace{-5mm}
\end{figure}

\section{Experiments}
\label{exp}
\subsection{Experimental Settings}
\subsubsection{Implementation Details.} Our framework is implemented in PyTorch\cite{paszke2019pytorch}. We adopt FLAME~\cite{Li2017LearningAM} as the driving 3DMM, and use its UV parameterization, with the UV map resolution set to $512 \times 512$. We set $K=4$ Gaussian layers for stacking and manually define $\Omega_{\mathrm{stack}}$ as the eye region using FLAME semantic masks, since its closely spaced eyelid and eyeball surfaces make it suitable for stacking. We obtain $\Omega_{\mathrm{coarse}}$ by subsampling the base Gaussian locations whose features are completed by $\mathbf{P}_\mathrm{f}$. We optimize the model using the Adam optimizer~\cite{Kingma2014AdamAM} with a learning rate of $1.0\!\times\!10^{-4}$, while keeping the DINOv2~\cite{Oquab2023DINOv2LR} feature extraction backbone frozen. The model is trained for $200K$ iterations with a batch size of 8. All experiments are conducted using NVIDIA RTX A6000 GPU hardware.

\vspace{-3.5mm}
\subsubsection{Datasets.}
We train our model on the VFHQ dataset~\cite{xie2022vfhq}, which contains 15,204 interview videos with diverse identities, expressions, and head poses. Following GAGAvatar~\cite{Chu2024GeneralizableAA}, we extract frames, detect the face region, crop it with an expanded bounding box, and resize the result to $512 \times 512$. We also remove the background following GPAvatar~\cite{Chu2024GPAvatarGA} and uniformly sample frames from each video, resulting in $500K$ frames in total. For evaluation, we use sampled frames from the official VFHQ test set and the HDTF dataset~\cite{zhang2021flow} following the test split used in prior work~\cite{ma2023otavatar,li2023generalizable}.
\vspace{-3.5mm}
\subsubsection{Evaluation Metrics.}
We evaluate our model under two settings: self and cross reenactment. For self reenactment, where ground truth images are available, we measure reconstruction quality using PSNR, SSIM~\cite{wang2004image}, and LPIPS~\cite{zhang2018unreasonable}. Identity similarity is computed as the cosine similarity (CSIM) between ArcFace~\cite{deng2019arcface} features. To assess expression and pose accuracy, we report the average expression distance (AED) and average pose distance (APD) based on a 3DMM estimator~\cite{deng2019accurate}, along with the average keypoint distance (AKD) using a facial landmark detector~\cite{bulat2017far}. For cross reenactment, where ground truth is unavailable, we report CSIM, AED, and APD.

\begin{table}[t]
\centering
\small
\setlength{\tabcolsep}{4pt}
\renewcommand{\arraystretch}{1.15}
\caption{\textbf{Quantitative comparison on VFHQ.}
Comparison with state-of-the-art methods on the VFHQ dataset. Colors denote the \colorbox{bestcolor}{best} and \colorbox{secondcolor}{second-best} results.}
\label{tab:vfhq}
\resizebox{0.95\linewidth}{!}{%
\begin{tabular}{lccccccc|ccc}
\Xhline{1.0pt}
\addlinespace[2.0pt]
& \multicolumn{7}{c|}{Self Reenactment} 
& \multicolumn{3}{c}{Cross Reenactment} \\
\cline{2-11}
\addlinespace[2.0pt]
Method
& PSNR$\uparrow$ 
& SSIM$\uparrow$ 
& LPIPS$\downarrow$ 
& CSIM$\uparrow$ 
& AKD$\downarrow$ 
& AED$\downarrow$ 
& APD$\downarrow$ 
& CSIM$\uparrow$ 
& AED$\downarrow$ 
& APD$\downarrow$ \\
\addlinespace[2.0pt]
\hline
\addlinespace[2.0pt]
GPAvatar~\cite{Chu2024GPAvatarGA}           & 20.85 & 0.767 & 0.149 & 0.781 & 4.44 & 0.130 & 0.318 & 0.527 & 0.218 & 0.351 \\
Portrait4D~\cite{Deng2023Portrait4DLO}      & 17.60 & 0.696 & 0.173 & 0.801 & 5.15 & 0.148 & \colorbox{bestcolor}{0.251} & 0.619 & 0.263 & \colorbox{bestcolor}{0.250} \\
Portrait4D-v2~\cite{Deng2024Portrait4Dv2PM} & 18.45 & 0.721 & 0.150 & 0.848 & 4.95 & 0.123 & \colorbox{secondcolor}{0.294} & 0.672 & 0.235 & \colorbox{secondcolor}{0.303} \\
GAGAvatar~\cite{Chu2024GeneralizableAA}     & \colorbox{secondcolor}{21.68} & \colorbox{secondcolor}{0.779} & \colorbox{bestcolor}{0.118} & \colorbox{bestcolor}{0.875} & \colorbox{secondcolor}{4.27} & \colorbox{secondcolor}{0.105} & 0.358 & \colorbox{secondcolor}{0.674} & \colorbox{bestcolor}{0.202} & 0.415 \\
LAM~\cite{He2025LAMLA}                      & 18.55 & 0.741 & 0.145 & 0.744 & 4.95 & 0.160 & 0.310 & 0.594 & 0.257 & 0.336 \\
\addlinespace[2.0pt]
\hline
\addlinespace[2.0pt]
Ours                                         & \colorbox{bestcolor}{21.76} & \colorbox{bestcolor}{0.784} & \colorbox{secondcolor}{0.119} & \colorbox{secondcolor}{0.871} & \colorbox{bestcolor}{3.92} & \colorbox{bestcolor}{0.101} & 0.399 & \colorbox{bestcolor}{0.676} & \colorbox{secondcolor}{0.206} & 0.445 \\
\addlinespace[2.0pt]
\Xhline{1.0pt}
\end{tabular}
} %
\end{table}

\begin{table}[t]
\centering
\small
\setlength{\tabcolsep}{4pt}
\renewcommand{\arraystretch}{1.15}
\caption{\textbf{Quantitative comparison on HDTF.}
Comparison with state-of-the-art methods on the HDTF dataset. Colors denote the \colorbox{bestcolor}{best} and \colorbox{secondcolor}{second-best} results.}
\label{tab:hdtf}
\resizebox{0.95\linewidth}{!}{%
\begin{tabular}{lccccccc|ccc}
\Xhline{1.0pt}
\addlinespace[2.0pt]
& \multicolumn{7}{c|}{Self Reenactment} 
& \multicolumn{3}{c}{Cross Reenactment} \\
\cline{2-11}
\addlinespace[2.0pt]
Method
& PSNR$\uparrow$ 
& SSIM$\uparrow$ 
& LPIPS$\downarrow$ 
& CSIM$\uparrow$
& AKD$\downarrow$ 
& AED$\downarrow$ 
& APD$\downarrow$ 
& CSIM$\uparrow$ 
& AED$\downarrow$ 
& APD$\downarrow$ \\
\addlinespace[2.0pt]
\hline
\addlinespace[2.0pt]
GPAvatar~\cite{Chu2024GPAvatarGA}           & 24.19 & 0.857 & 0.076 & 0.918 & \colorbox{secondcolor}{3.42} & 0.130 & \colorbox{bestcolor}{0.180} & 0.839 & 0.249 & \colorbox{secondcolor}{0.227} \\
Portrait4D~\cite{Deng2023Portrait4DLO}      & 19.26 & 0.776 & 0.125 & 0.879 & 4.27 & 0.174 & \colorbox{secondcolor}{0.183} & 0.805 & 0.298 & \colorbox{bestcolor}{0.198} \\
Portrait4D-v2~\cite{Deng2024Portrait4Dv2PM} & 19.16 & 0.791 & 0.103 & 0.912 & 3.95 & 0.132 & 0.220 & 0.848 & 0.260 & 0.239 \\
GAGAvatar~\cite{Chu2024GeneralizableAA}     & \colorbox{secondcolor}{24.98} & \colorbox{secondcolor}{0.865} & \colorbox{bestcolor}{0.067} & \colorbox{secondcolor}{0.933} & 3.53 & \colorbox{secondcolor}{0.120} & 0.244 & \colorbox{secondcolor}{0.871} & \colorbox{secondcolor}{0.214} & 0.289 \\
LAM~\cite{He2025LAMLA}                      & 21.36 & 0.808 & 0.096 & 0.831 & 4.27 & 0.184 & 0.203 & 0.766 & 0.279 & 0.229 \\
\addlinespace[2.0pt]
\hline
\addlinespace[2.0pt]
Ours                                         & \colorbox{bestcolor}{25.48} & \colorbox{bestcolor}{0.872} & \colorbox{secondcolor}{0.068} & \colorbox{bestcolor}{0.936} & \colorbox{bestcolor}{3.12} & \colorbox{bestcolor}{0.105} & 0.270 & \colorbox{bestcolor}{0.875} & \colorbox{bestcolor}{0.202} & 0.306 \\
\addlinespace[2.0pt]
\Xhline{1.0pt}
\end{tabular}
} %
\vspace{-4mm}
\end{table}

\subsection{Main Results}
\subsubsection{Baselines.}
We compare our method with existing state-of-the-art approaches, including GPAvatar~\cite{Chu2024GPAvatarGA}, Portrait4D~\cite{Deng2023Portrait4DLO}, Portrait4D-v2~\cite{Deng2024Portrait4Dv2PM}, GAGAvatar~\cite{Chu2024GeneralizableAA}, and LAM~\cite{He2025LAMLA}. All these methods support feed-forward avatar reconstruction from a single source image. We use their official implementations to obtain the results.
\vspace{-3mm}
\subsubsection{Qualitative Results.}
Fig.~\ref{fig:qual-cross} shows qualitative comparisons for cross reenactment. GPAvatar shows limited preservation of fine texture details and noticeable identity shifts. Portrait4D and Portrait4D-v2 preserve facial appearance well, but show inaccurate expressions and chin misalignment under large pose variations (fourth row). GAGAvatar achieves generally plausible results, but lacks explicit treatment of unobserved regions and tends to complete missing areas with generic head appearance rather than identity-specific details. LAM shows degraded rendering quality when the target view differs substantially from the source view. In contrast, our method better preserves identity under large pose changes by reconstructing identity-consistent details in previously unobserved regions, including hair color, hairstyle, and facial hair. These results demonstrate that our method improves identity preservation while maintaining accurate reenactment. Additional qualitative results are provided in the supplementary material.
\vspace{-4mm}
\subsubsection{Quantitative Results.}
We also report quantitative results on the VFHQ and HDTF datasets in Tab.~\ref{tab:vfhq} and Tab.~\ref{tab:hdtf}, respectively. Our method achieves better reconstruction performance on metrics such as PSNR and SSIM, while obtaining competitive LPIPS results compared to GAGAvatar. Moreover, the CSIM, AKD, and AED results indicate competitive or better performance in identity preservation and expression accuracy compared with previous approaches. We also observe that our method shows worse APD. Symmetry-based completion improves texture and identity preservation in unobserved regions. However, predicting geometry offsets from completed features without geometric supervision may introduce subtle geometric inconsistencies. Nevertheless, the qualitative results demonstrate that our method faithfully preserves the target pose in most cases.

\vspace{-3mm}

\subsection{Evaluation on Unobserved Regions}
\vspace{-2mm}
\setlength{\intextsep}{1pt}
\begin{wraptable}{r}{0.5\textwidth}
\vspace{-4mm}
\centering
\small
\setlength{\tabcolsep}{4pt}
\renewcommand{\arraystretch}{1.15}
\caption{\textbf{Evaluation on unobserved regions.}
Reconstruction metrics for source-unobserved and target-visible regions.}
\label{tab:unobserved}
\resizebox{\linewidth}{!}{%
\begin{tabular}{lccc}
\Xhline{1.0pt}
Method & PSNR$\uparrow$ & SSIM$\uparrow$ & LPIPS$\downarrow$ \\
\hline
GAGAvatar          & 18.28 & 0.526 & \colorbox{bestcolor}{0.175} \\
Ours (w/o symmetry) & 18.78 & 0.543 & 0.182 \\
Ours (full)         & \colorbox{bestcolor}{19.06} & \colorbox{bestcolor}{0.546} & 0.179 \\
\Xhline{1.0pt}
\end{tabular}
}
\end{wraptable}
We evaluate reconstruction quality in regions unobserved in the source view but visible in the target view. We select VFHQ self-reenactment pairs with source-target yaw differences greater than $20^\circ$ and compute PSNR, SSIM, and LPIPS over these regions. As shown in Tab.~\ref{tab:unobserved}, our full model improves all three metrics over the variant without symmetry and achieves higher PSNR and SSIM than GAGAvatar. The PSNR gain over the variant without symmetry and the PSNR/SSIM margins over GAGAvatar are larger than in full-image evaluation, supporting the effectiveness of the proposed method in unobserved regions.


\subsection{Efficiency}
\vspace{-2mm}
We evaluate model efficiency in terms of rendering speed (Tab.~\ref{tab:fps}) and reconstruction time (Tab.~\ref{tab:recons}). Reconstruction time is measured in our A6000 environment and denotes the duration required to prepare an avatar from a given input image. All reported runtimes measure model inference only, excluding source preprocessing and FLAME parameter estimation. We compute the rendering speed under two GPU configurations. For the A100 setting, we adopt the FPS values from prior work~\cite{Chu2024GeneralizableAA,He2025LAMLA}. We then compare our method and the baselines that support real-time rendering in our A6000 environment. Although LAM achieves the highest rendering speed, it requires a few seconds for avatar reconstruction due to its cross-attention module. Our method is not the best in either metric, but it maintains a short reconstruction time while supporting real-time rendering.

\begin{table}[t]
\centering
\small
\setlength{\tabcolsep}{4pt}
\renewcommand{\arraystretch}{1.15}
\caption{\textbf{Runtime comparison.}
Runtime comparison measured in FPS. The results are averaged over 100 frames, excluding the time for estimating driving parameters that can be computed in advance.}
\label{tab:fps}
\vspace{-2mm}
\resizebox{0.95\linewidth}{!}{%
\begin{tabular}{lccccc|ccc}
\Xhline{1.0pt}
\addlinespace[2.0pt]
& \multicolumn{5}{c|}{A100 GPU} 
& \multicolumn{3}{c}{A6000 GPU} \\
\cline{2-9}
\addlinespace[2.0pt]
Method & GPAvatar & Portrait4D  & Portrait4D-v2 & GAGAvatar  & LAM  & GAGAvatar  & LAM  & Ours \\
\addlinespace[2.0pt]
\hline
\addlinespace[2.0pt]
Rendering speed (FPS) 
& 16.86  & 9.49  & 9.62  & 67.12  & 280.96  & 56.76  & 259.58  & 53.73 \\
\addlinespace[2.0pt]
\Xhline{1.0pt}
\end{tabular}
} %
\end{table}

\begin{table}[t]
\centering
\small
\setlength{\tabcolsep}{4pt}
\renewcommand{\arraystretch}{1.15}
\caption{\textbf{Reconstruction time.}
Reconstruction time comparison measured in seconds.}
\label{tab:recons}
\vspace{-4mm}
\resizebox{0.5\linewidth}{!}{%
\begin{tabular}{lccc}
\Xhline{1.0pt}
\addlinespace[2.0pt]
& \multicolumn{3}{c}{A6000 GPU} \\
\cline{2-4}
\addlinespace[2.0pt]
Method & GAGAvatar  & LAM  & Ours \\
\addlinespace[2.0pt]
\hline
\addlinespace[2.0pt]
Reconstruction time (sec) 
& 0.04  & 2.82  & 0.07 \\
\addlinespace[2.0pt]
\Xhline{1.0pt}
\end{tabular}
} %
\end{table}

\begin{table}[!t]
\centering
\small
\setlength{\tabcolsep}{4pt}
\renewcommand{\arraystretch}{1.15}
\caption{\textbf{Ablation study.}
Ablations on the VFHQ dataset.}
\label{tab:ablation}
\resizebox{0.95\linewidth}{!}{%
\begin{tabular}{lccccccc|ccc}
\Xhline{1.0pt}
\addlinespace[2.0pt]
& \multicolumn{7}{c|}{Self Reenactment} 
& \multicolumn{3}{c}{Cross Reenactment} \\
\cline{2-11}
\addlinespace[2.0pt]
Method
& PSNR$\uparrow$  & SSIM$\uparrow$  & LPIPS$\downarrow$  & CSIM$\uparrow$ & AKD$\downarrow$  & AED$\downarrow$  & APD$\downarrow$ 
& CSIM$\uparrow$  & AED$\downarrow$  & APD$\downarrow$ \\
\addlinespace[2.0pt]
\hline
\addlinespace[2.0pt]
w/o position offset & 21.54 & 0.780 & 0.123 & 0.869 & \colorbox{bestcolor}{3.90} & 0.101 & 0.412 & 0.671 & \colorbox{bestcolor}{0.204} & 0.434 \\
w/o symmetry        & 21.63 & 0.781 & 0.121 & 0.868 & 3.92 & \colorbox{bestcolor}{0.100} & \colorbox{bestcolor}{0.385} & 0.664 & 0.208 & \colorbox{bestcolor}{0.425} \\
\addlinespace[2.0pt]
\hline
\addlinespace[2.0pt]
Ours                & \colorbox{bestcolor}{21.76} & \colorbox{bestcolor}{0.784} & \colorbox{bestcolor}{0.119} & \colorbox{bestcolor}{0.871} & 3.92 & 0.101 & 0.399 & \colorbox{bestcolor}{0.676} & 0.206 & 0.445 \\
\addlinespace[2.0pt]
\Xhline{1.0pt}
\end{tabular}
} %
\end{table}

\begin{table}[!t]
\centering
\small
\setlength{\tabcolsep}{4pt}
\renewcommand{\arraystretch}{1.15}
\caption{\textbf{Ablation study on Gaussian stacking.}
Ablation results for Gaussian stacking in the self reenactment setting.}
\label{tab:stacking}
\resizebox{0.4\linewidth}{!}{%
\begin{tabular}{lccc}
\Xhline{1.0pt}
\addlinespace[2.0pt]
Method
& PSNR$\uparrow$ & SSIM$\uparrow$ & LPIPS$\downarrow$ \\
\addlinespace[2.0pt]
\hline
\addlinespace[2.0pt]
w/o stacking
& 24.77 & 0.774 & 0.082 \\
\addlinespace[2.0pt]
\hline
\addlinespace[2.0pt]
Ours
& \colorbox{bestcolor}{24.99} & \colorbox{bestcolor}{0.782} & \colorbox{bestcolor}{0.081} \\
\addlinespace[2.0pt]
\Xhline{1.0pt}
\end{tabular}
} %
\vspace{-4mm}
\end{table}

\subsection{Ablation Studies}
We conduct ablation studies to evaluate the effectiveness of each proposed component. All variants are trained using the same settings described above and evaluated on the VFHQ test set~\cite{xie2022vfhq}.
\vspace{-4mm}
\subsubsection{Position Offset.}
We conduct an ablation study by removing the offset used to compute the final Gaussian positions. Without the offset, each Gaussian remains at its initial position on the FLAME surface, and samples its UV feature from that initial location. The results in Tab.~\ref{tab:ablation} indicate that the predicted offset enables more accurate UV feature sampling, leading to improved detail reconstruction.
\vspace{-4mm}
\subsubsection{Symmetry Completion.}
\begin{figure}[t]
\centering
\begin{minipage}[t]{0.48\textwidth}
\centering
\setlength{\tabcolsep}{0pt}
\renewcommand{\arraystretch}{0}
\resizebox{\linewidth}{!}{
\begin{tabular}{@{}cccc@{}}
\includegraphics[width=0.25\linewidth]{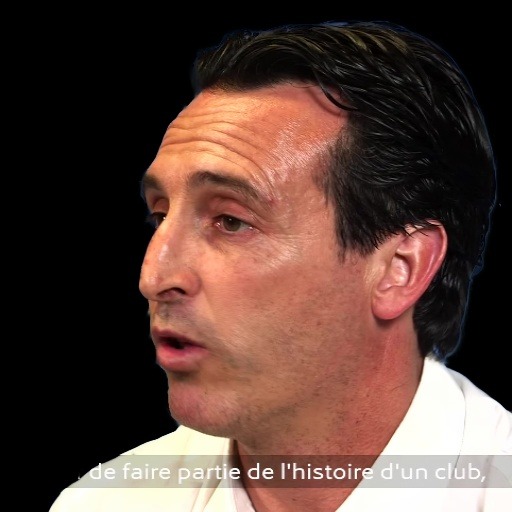} &
\includegraphics[width=0.25\linewidth]{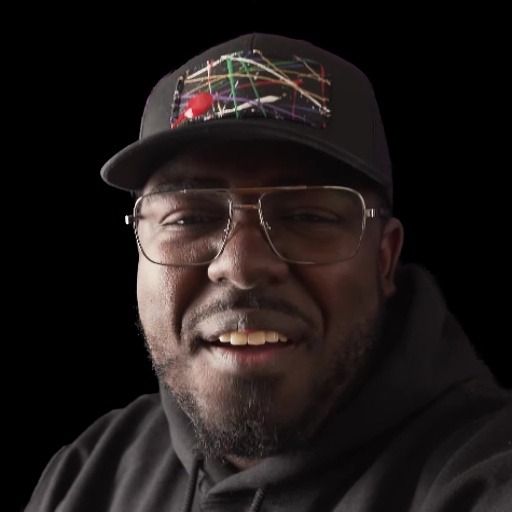} &
\includegraphics[width=0.25\linewidth]{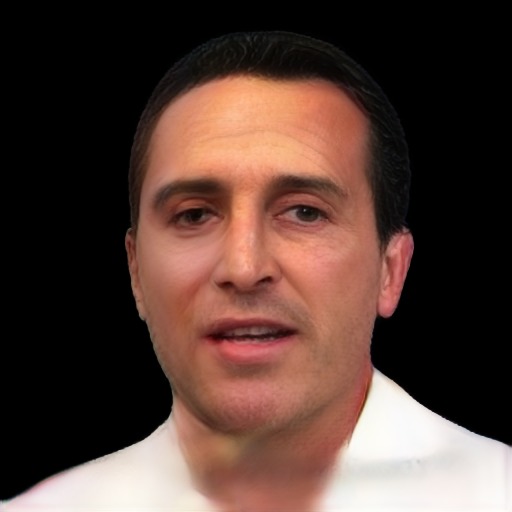} &
\includegraphics[width=0.25\linewidth]{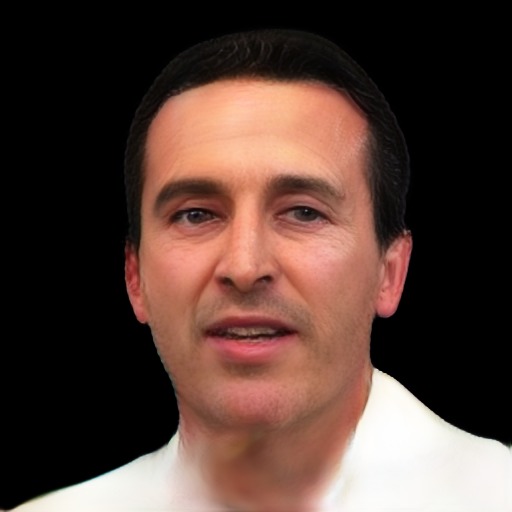} \\
\includegraphics[width=0.25\linewidth]{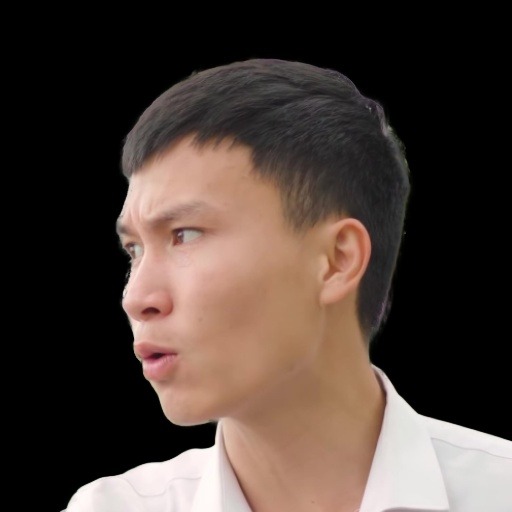} &
\includegraphics[width=0.25\linewidth]{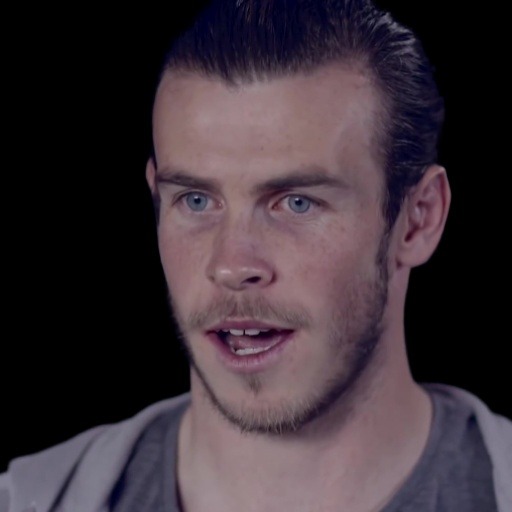} &
\includegraphics[width=0.25\linewidth]{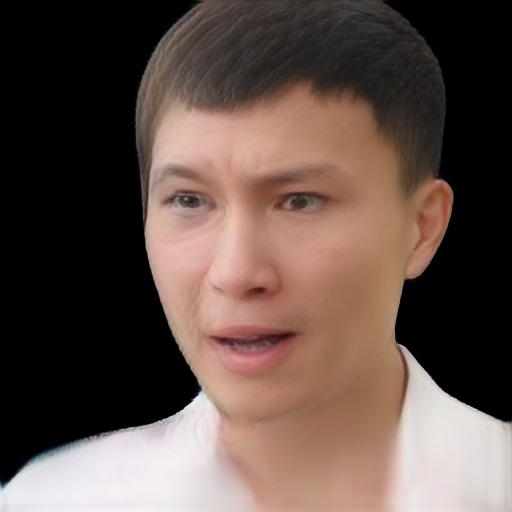} &
\includegraphics[width=0.25\linewidth]{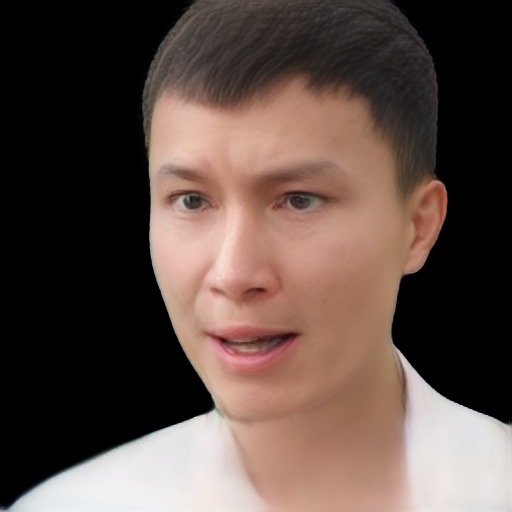} \\[3pt]
Source & Driving & w/o sym. & Ours \\
\end{tabular}
}
\caption{\textbf{Qualitative comparison of symmetry completion.} Comparison without (third column) and with symmetry completion (fourth column).}
\label{fig:ab_sym}
\end{minipage}
\hfill
\begin{minipage}[t]{0.48\textwidth}
\centering
\setlength{\tabcolsep}{0pt}
\renewcommand{\arraystretch}{0}
\resizebox{\linewidth}{!}{
\begin{tabular}{@{}cccc@{}}
\includegraphics[width=0.25\linewidth]{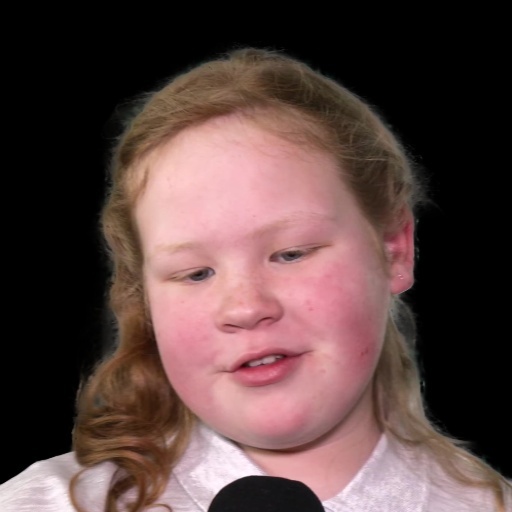} &
\includegraphics[width=0.25\linewidth]{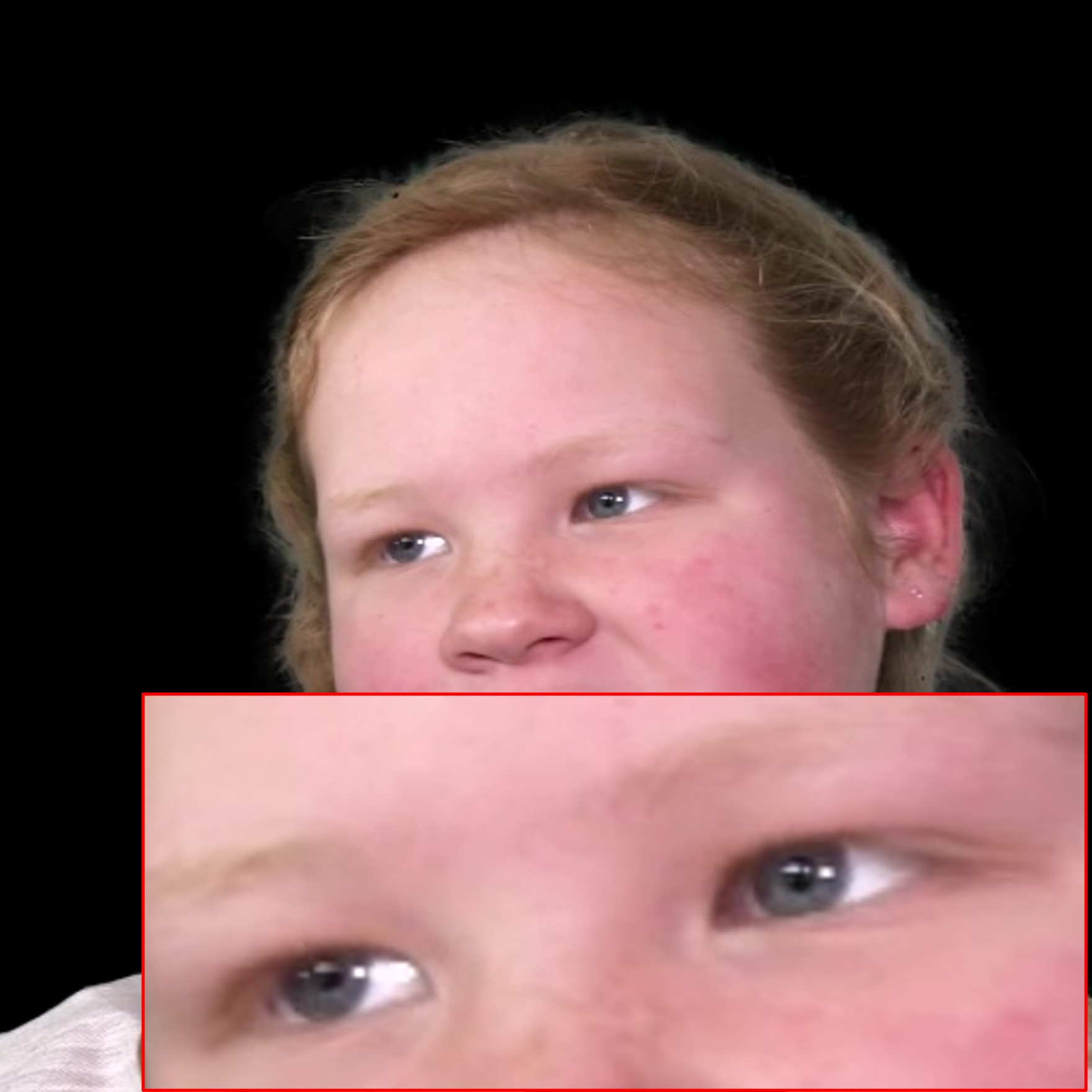} &
\includegraphics[width=0.25\linewidth]{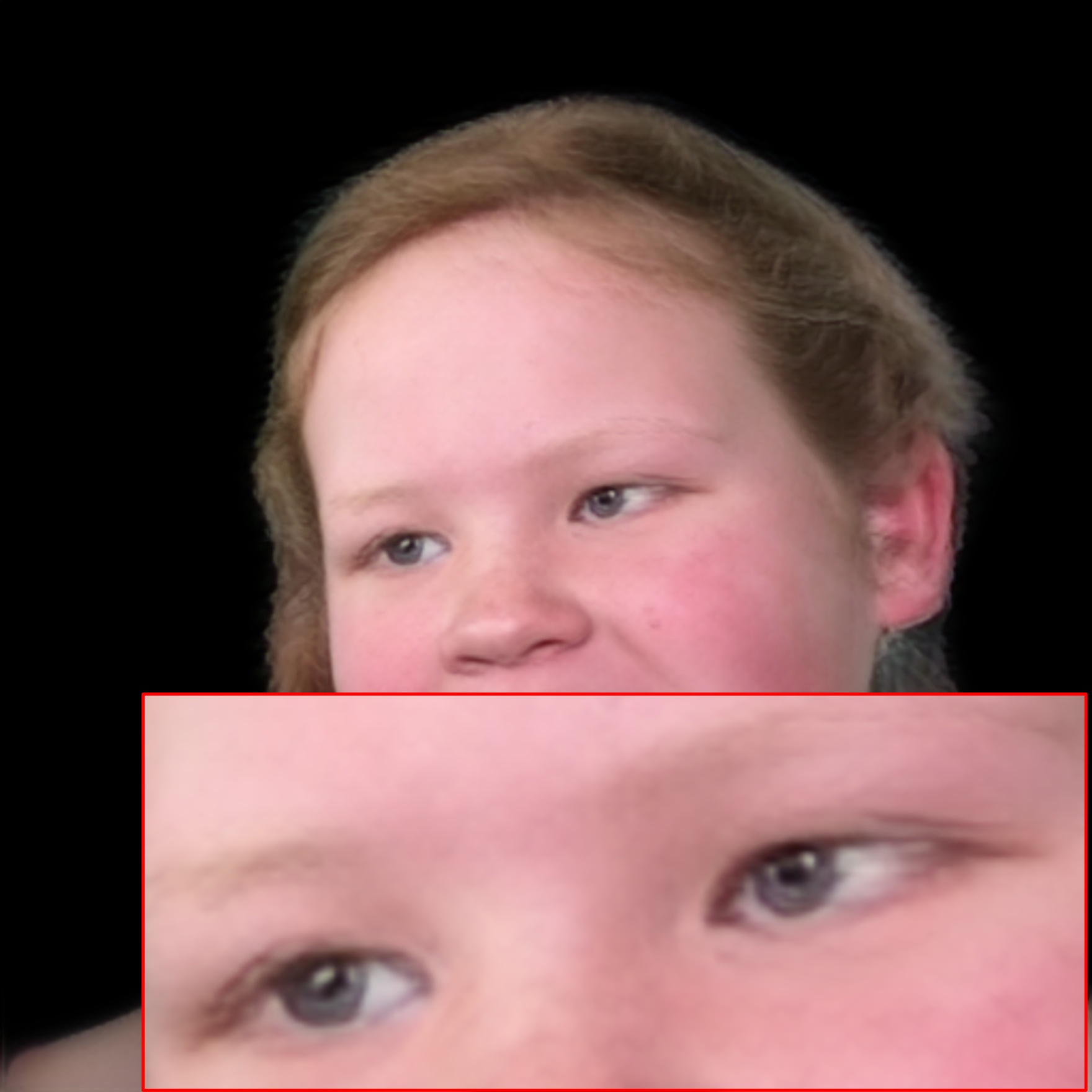} &
\includegraphics[width=0.25\linewidth]{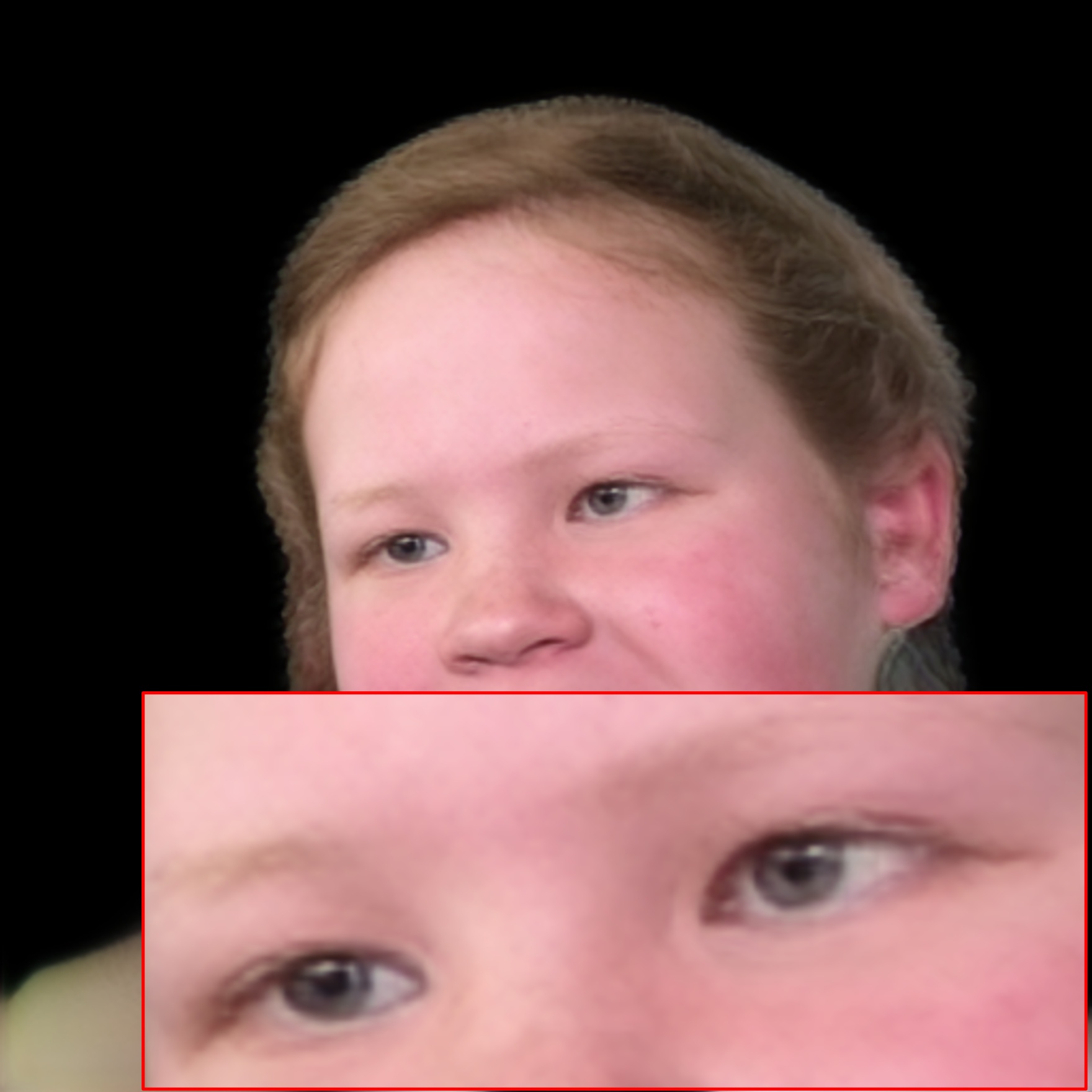} \\
\includegraphics[width=0.25\linewidth]{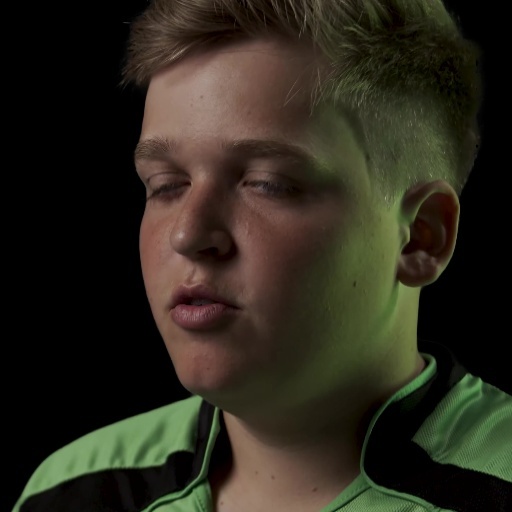} &
\includegraphics[width=0.25\linewidth]{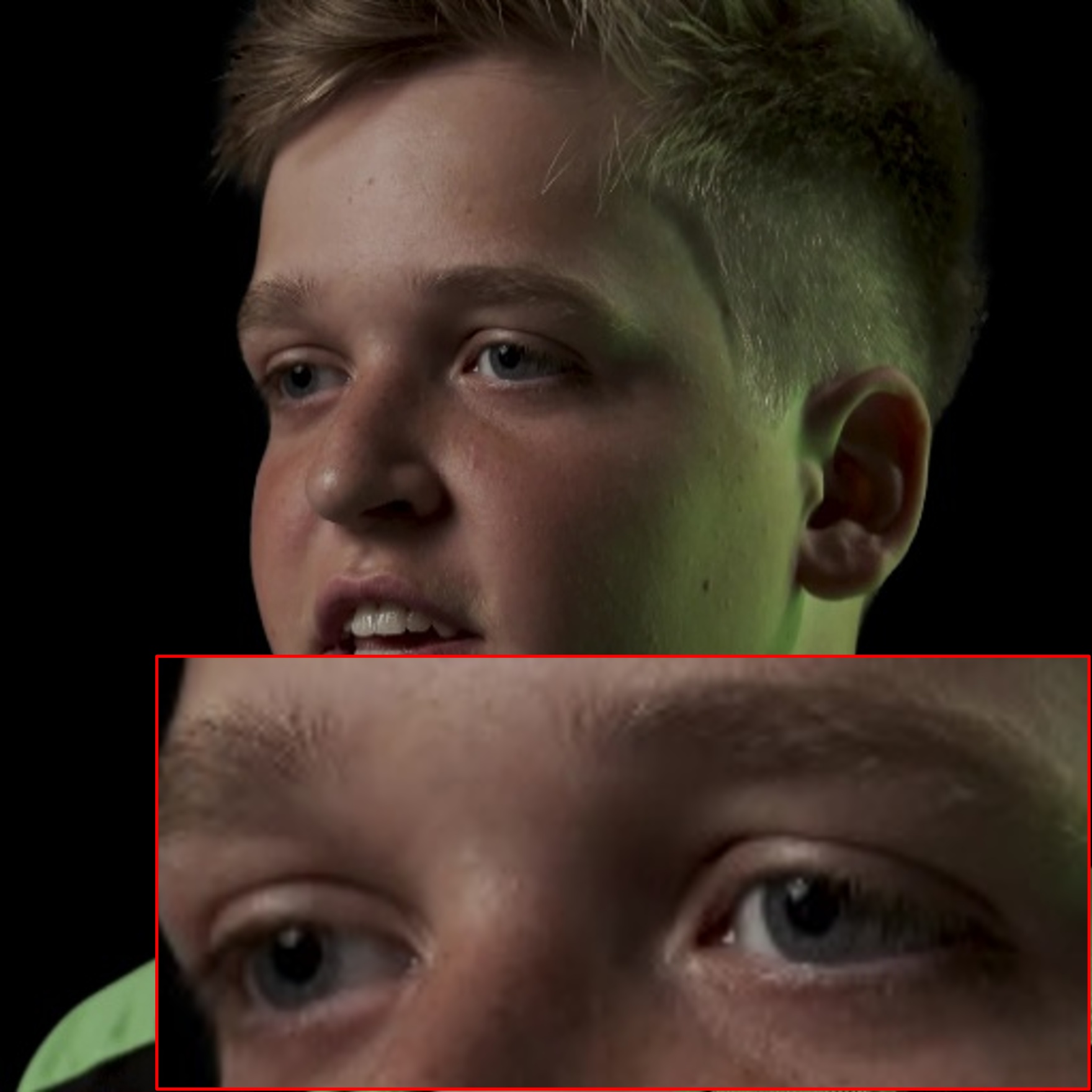} &
\includegraphics[width=0.25\linewidth]{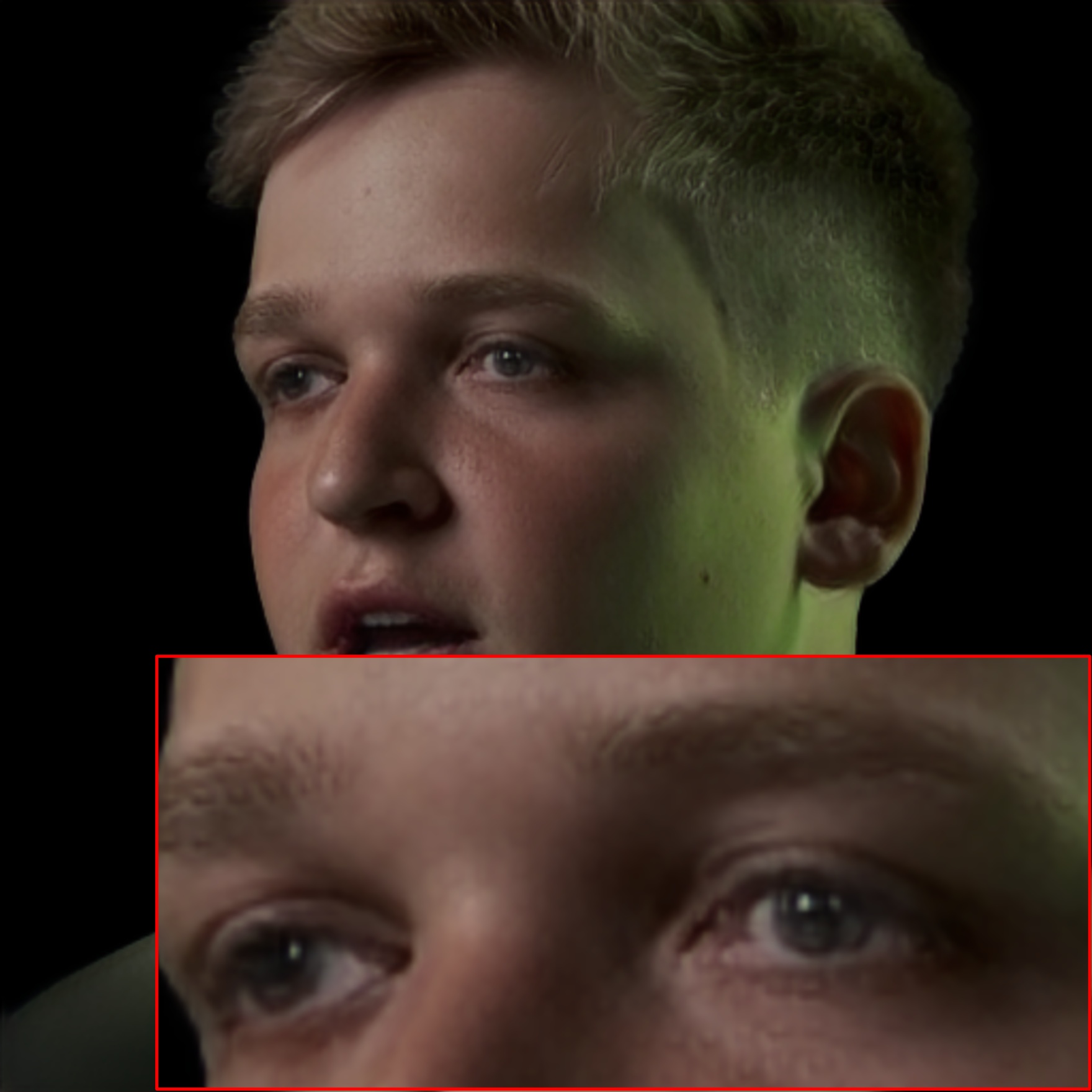} &
\includegraphics[width=0.25\linewidth]{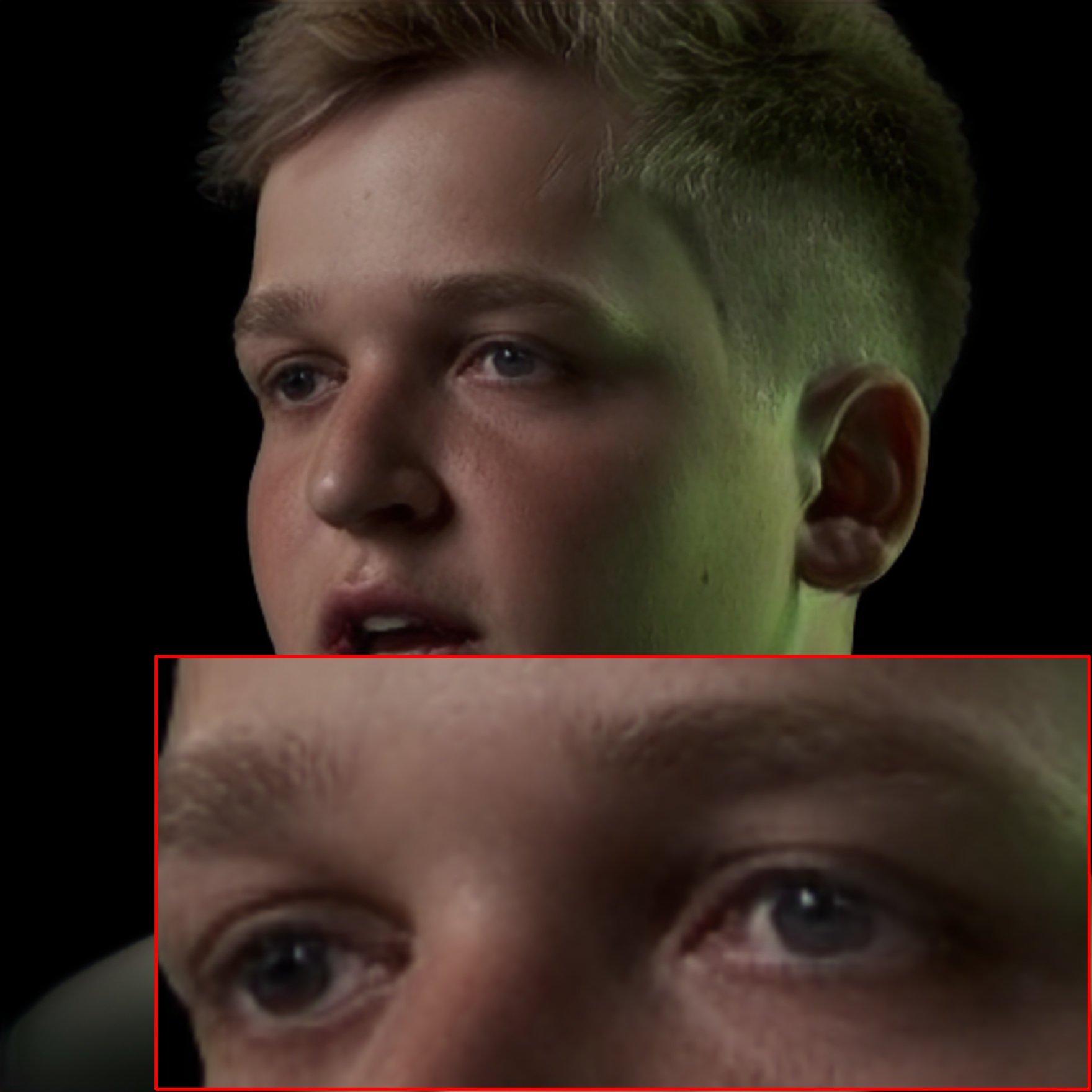} \\[3pt]
Source & Driving & w/o stack & Ours \\
\end{tabular}
}
\caption{\textbf{Qualitative comparison of Gaussian stacking in the eye region.} Comparison without (third column) and with Gaussian stacking (fourth column).}
\label{fig:ab_stack}
\end{minipage}
\end{figure}

Fig.~\ref{fig:ab_sym} qualitatively evaluates symmetry completion by replacing symmetric features with learnable features for unobserved regions. Since these features are not directly derived from the input image, they fail to preserve identity-specific details in missing areas. In contrast, our symmetry completion transfers information from the corresponding visible regions, resulting in more reliable and identity-consistent reconstructions. The quantitative results in Tab.~\ref{tab:ablation} further validate the effectiveness of this design.

\subsubsection{Gaussian Stacking.}
We further evaluate the effect of Gaussian stacking. In our experiments, stacking is applied to the eye region, so the metrics are computed over the corresponding area. We report these results in the self reenactment setting, where ground-truth images are available. The quantitative results are presented in Tab.~\ref{tab:stacking}. As shown in Fig.~\ref{fig:ab_stack}, the model without stacking exhibits color bleeding between the iris and sclera and blurred eye corners (first row), whereas Gaussian stacking produces a cleaner and more clearly defined eyelid boundary (second row). These results demonstrate that Gaussian stacking helps preserve fine structural details.

\vspace{-1mm}

\section{Conclusion and Limitations}
\label{conc}
In this paper, we propose SInGA, a feed-forward framework that reconstructs an animatable Gaussian head avatar from a single image. Our method constructs Gaussian primitives using a UV-aligned surface representation, enabling structured placement in the UV domain. By anchoring the Gaussians to its fixed topology, our framework facilitates semantically guided inpainting of unobserved regions and region-adaptive enhancement. To recover regions not visible in the input image, we introduce symmetry-guided completion, which transfers information from corresponding visible regions to obtain more reliable and identity-consistent features. We further introduce Gaussian stacking to increase the representation capacity of selected regions and capture fine details. Experiments demonstrate that our method improves reconstruction quality while preserving identity consistency. We hope our work encourages further research on identity-preserving head avatar reconstruction from limited visual observations.

Our symmetry-guided completion is limited by hidden asymmetric details, which remain ambiguous from a single image, and is sensitive to lighting asymmetry. Our method also exhibits higher APD than some baselines and can show jaw contour drift in challenging cases, leaving room for improvement. Future work will explore decoupling lighting and reflectance to improve inpainting and enable relighting, improving pose accuracy, and adaptively selecting stacking regions beyond the predefined eye region.

%
%
\bibliographystyle{splncs04}
\bibliography{main}
\end{document}


\title{\texorpdfstring{
Learning Semantic Inpainting for Animatable Gaussian Head Avatars\\[0.5em]
{\large Supplementary Material}
}{
Learning Semantic Inpainting for Animatable Gaussian Head Avatars: Supplementary Material
}}

\titlerunning{SInGA}

\author{}

\authorrunning{}

\institute{}

\maketitle

\setcounter{section}{5}
\setcounter{table}{7}
\setcounter{figure}{7}

\section{Preliminaries}
\label{pre}
\textbf{FLAME}~\cite{Li2017LearningAM} is a parametric 3D head model to represent animatable facial geometry. Given identity shape parameters $\boldsymbol{\beta}$, expression parameters $\boldsymbol{\psi}$, and pose parameters $\boldsymbol{\theta}$, FLAME first deforms a template mesh $\overline{\mathbf{V}}$ by applying additive blendshapes:
\begin{equation}
    \mathbf{T}_{P} =
    \overline{\mathbf{V}}
    + B_{S}(\boldsymbol{\beta})
    + B_{E}(\boldsymbol{\psi})
    + B_{P}(\boldsymbol{\theta}),
\end{equation}
where $B_{S}$, $B_{E}$, and $B_{P}$ denote the shape, expression, and pose corrective blendshapes, respectively. 
The posed mesh vertices are then obtained through linear blend skinning:
\begin{equation}
    \mathbf{V} =
    W\left(
    \mathbf{T}_{P},
    \mathbf{J}(\boldsymbol{\beta}),
    \boldsymbol{\theta},
    \mathcal{W}
    \right),
\end{equation}
where $\mathbf{J}(\boldsymbol{\beta})$ represents the articulated joint locations and $\mathcal{W}$ denotes the skinning weights. 
In our method, FLAME provides a structured head prior for modeling animatable human head avatars.

\vspace{5mm}

\noindent\textbf{3D Gaussian Splatting} (3DGS)~\cite{Kerbl20233DGS} models a scene as a set of anisotropic 3D Gaussian primitives.
Each Gaussian is parameterized by a center $\boldsymbol{\mu} \in \mathbb{R}^{3}$, a scaling vector $\mathbf{s} \in \mathbb{R}^{3}$, a rotation quaternion $\mathbf{q} \in \mathbb{R}^{4}$, an opacity $\alpha \in \mathbb{R}$, and appearance coefficients $\mathbf{c}$, which are commonly represented by spherical harmonics in 3DGS 
(we use $\mathbf{c} \in \mathbb{R}^{32}$ as color features).
A Gaussian centered at $\boldsymbol{\mu}$ is defined as:
\begin{equation}
    G(\mathbf{x}) =
    \exp\left(
    -\frac{1}{2}
    (\mathbf{x}-\boldsymbol{\mu})^{\top}
    \boldsymbol{\Sigma}^{-1}
    (\mathbf{x}-\boldsymbol{\mu})
    \right),
\end{equation}
where the covariance matrix $\boldsymbol{\Sigma} \in \mathbb{R}^{3 \times 3}$ is constructed from $\mathbf{s}$ and $\mathbf{q}$. 
With differentiable tile-based rasterization and alpha compositing, 3DGS enables efficient and high-fidelity rendering of explicit 3D primitives.

\clearpage

\section{Network Architecture}
\label{net}
We describe the architectures of the geometry network $\mathcal{G}_{\mathrm{geo}}$ and the appearance network $\mathcal{G}_{\mathrm{app}}$, which predict offset magnitudes and Gaussian attributes, respectively. $\mathcal{G}_{\mathrm{geo}}$ takes the aggregated feature map $\mathbf{M}_{\mathrm{f}}^{\mathrm{agg}}$ as input, while $\mathcal{G}_{\mathrm{app}}$ operates on the resampled feature map $\mathbf{M}_{\mathrm{f}}'$. Both feature maps have a size of $256 \times 512 \times 512$. For both networks, a $27$-dimensional harmonic encoding of the viewing direction is concatenated with the input feature map along the channel dimension. We report the architecture for a single Gaussian layer in Tab.~\ref{tab:net_geo} and Tab.~\ref{tab:net_app}. For stacked regions with $K$ layers, the final prediction layers output $K \times 1$ geometry offsets and $K \times 40$ Gaussian attributes. Each $40$-dimensional attribute vector consists of $32$ color channels, $1$ opacity value, $3$ scale parameters, and $4$ rotation parameters. Sigmoid activation is applied to the first three color channels, opacity, and scale, while the rotation parameters are L$2$ normalized.

\begin{table}[h]
\centering
\small
\setlength{\tabcolsep}{4pt}
\renewcommand{\arraystretch}{1.15}
\caption{\textbf{Architecture of the geometry network.}}
\label{tab:net_geo}
\resizebox{0.75\linewidth}{!}{%
\begin{tabular}{lcccc}
\Xhline{1.0pt}
\addlinespace[2.0pt]
Layer & Filter Size & Input Channels & Output Channels & Activation \\
\Xhline{0.6pt}
\addlinespace[1.0pt]
$\text{Conv}_{1}$ & $3 \times 3$ & $283$ & $128$ & ReLU \\
$\text{Conv}_{2}$ & $3 \times 3$ & $128$ & $128$ & ReLU \\
$\text{Conv}_{3}$ & $3 \times 3$ & $128$ & $128$ & ReLU \\
$\text{Conv}_{4}$ & $1 \times 1$ & $128$ & $1$   & Tanh \\
\addlinespace[1.0pt]
\Xhline{1.0pt}
\end{tabular}%
}
\end{table}

\begin{table}[h]
\centering
\small
\setlength{\tabcolsep}{4pt}
\renewcommand{\arraystretch}{1.15}
\caption{\textbf{Architecture of the appearance network.}}
\label{tab:net_app}
\resizebox{0.75\linewidth}{!}{%
\begin{tabular}{lcccc}
\Xhline{1.0pt}
\addlinespace[2.0pt]
Layer & Filter Size & Input Channels & Output Channels & Activation \\
\Xhline{0.6pt}
\addlinespace[1.0pt]
$\text{Conv}_{1}$ & $3 \times 3$ & $283$ & $128$ & ReLU \\
$\text{Conv}_{2}$ & $3 \times 3$ & $128$ & $128$ & ReLU \\
$\text{Conv}_{3}$ & $3 \times 3$ & $128$ & $128$ & ReLU \\
$\text{Conv}_{4}$ & $1 \times 1$ & $128$ & $40$  & - \\
\addlinespace[1.0pt]
\Xhline{1.0pt}
\end{tabular}%
}
\end{table}
\vspace{-3mm}

\section{Additional Qualitative Results}
\label{more_qual}
\begin{figure}[!p]
\centering
\setlength{\tabcolsep}{0pt}
\renewcommand{\arraystretch}{0}
\resizebox{\textwidth}{!}{
\begin{tabular}{@{}cccccccc@{}}
\includegraphics[width=0.125\textwidth]{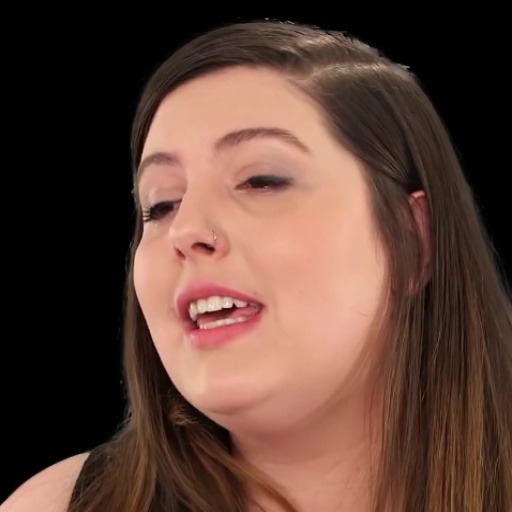} &
\includegraphics[width=0.125\textwidth]{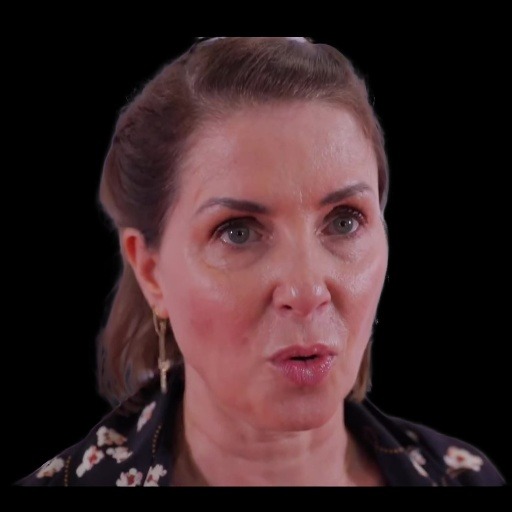} &
\includegraphics[width=0.125\textwidth]{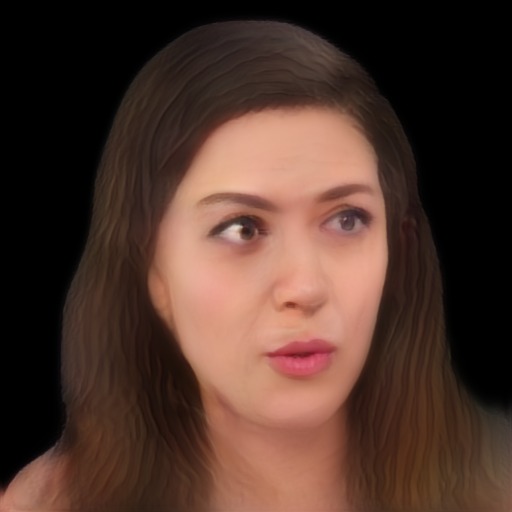} &
\includegraphics[width=0.125\textwidth]{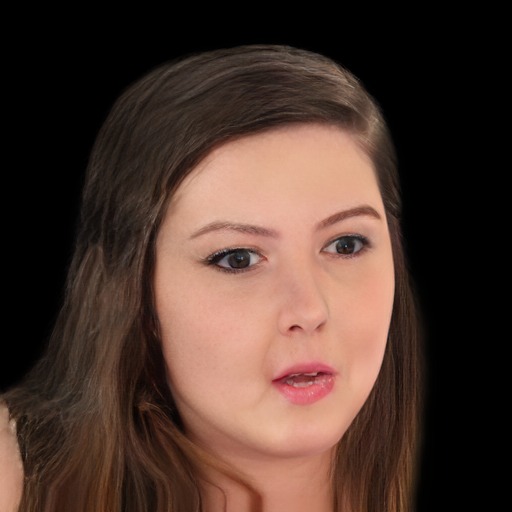} &
\includegraphics[width=0.125\textwidth]{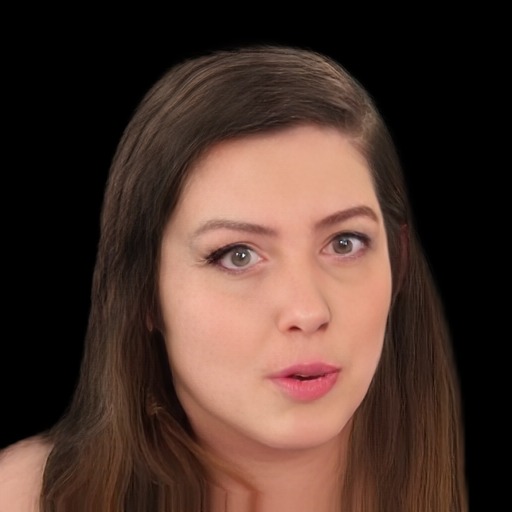} &
\includegraphics[width=0.125\textwidth]{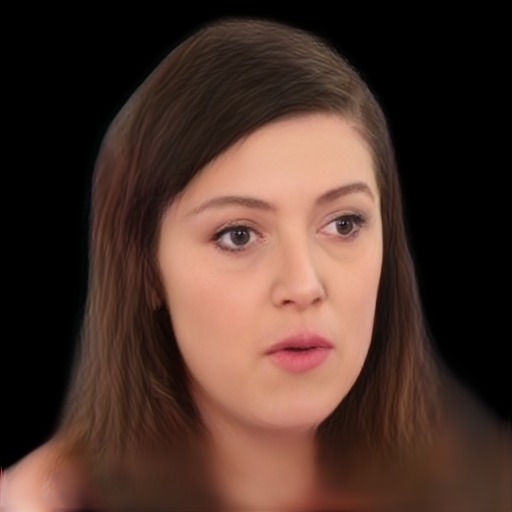} &
\includegraphics[width=0.125\textwidth]{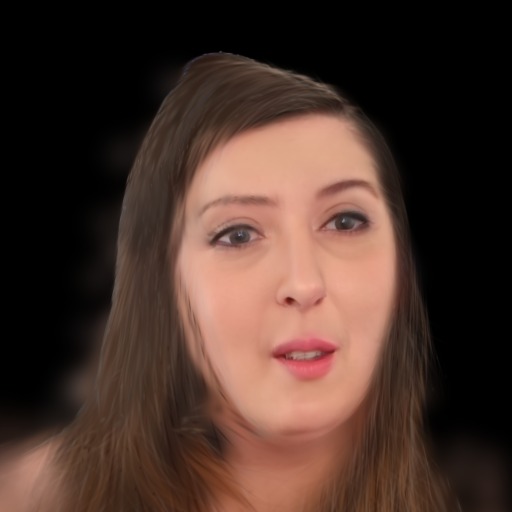} &
\includegraphics[width=0.125\textwidth]{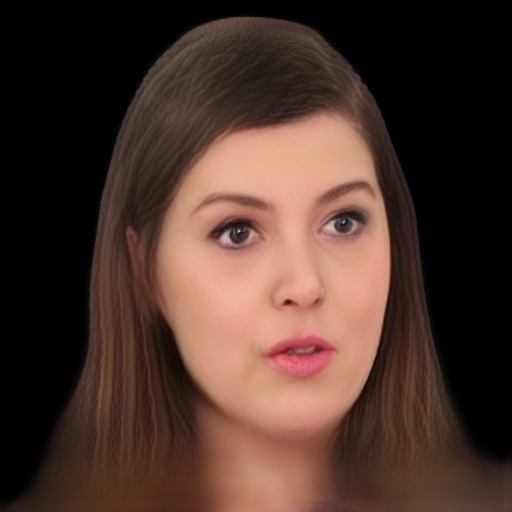}  \\
\includegraphics[width=0.125\textwidth]{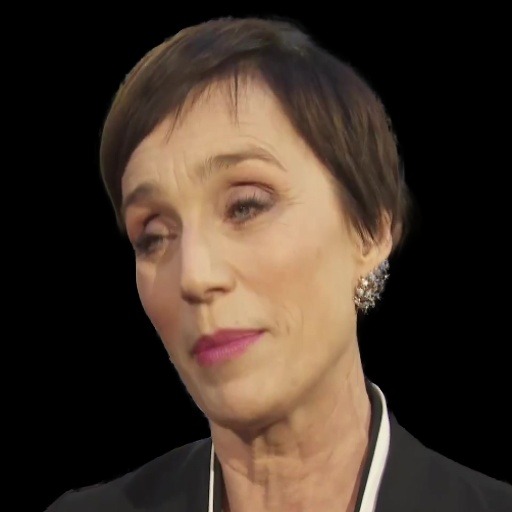} &
\includegraphics[width=0.125\textwidth]{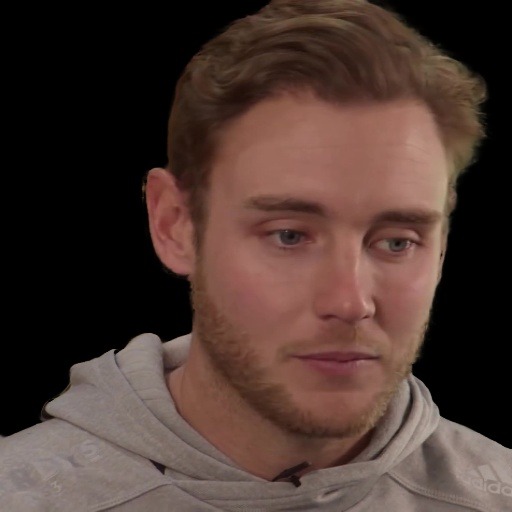} &
\includegraphics[width=0.125\textwidth]{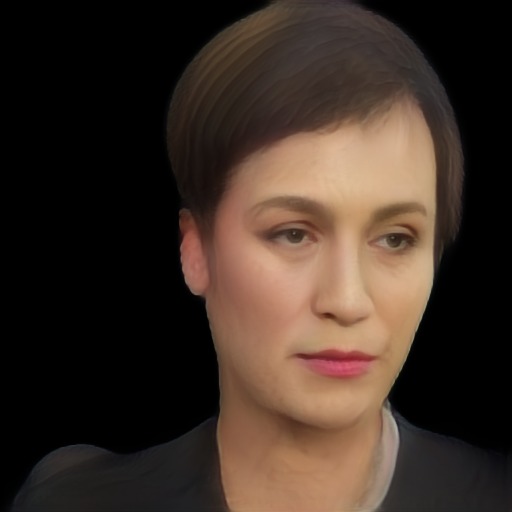} &
\includegraphics[width=0.125\textwidth]{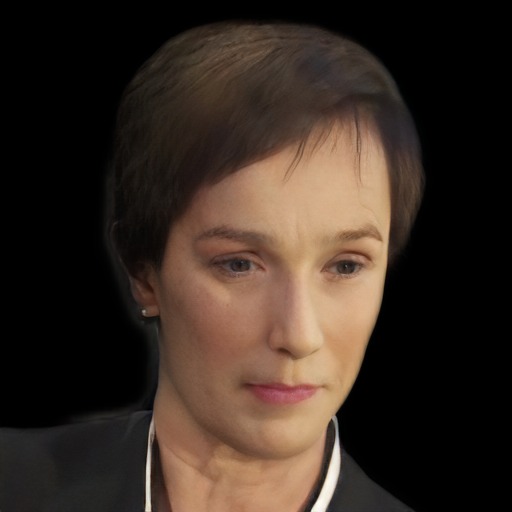} &
\includegraphics[width=0.125\textwidth]{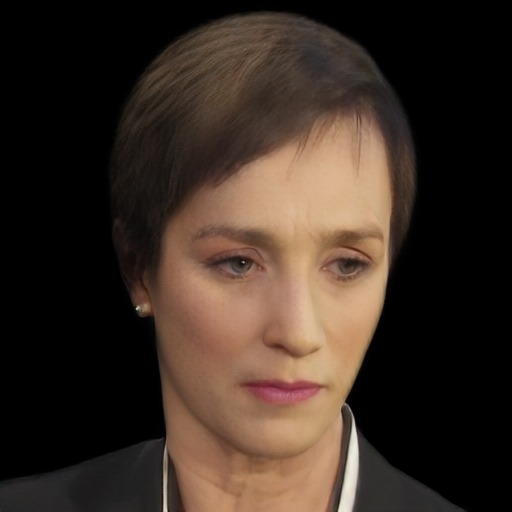} &
\includegraphics[width=0.125\textwidth]{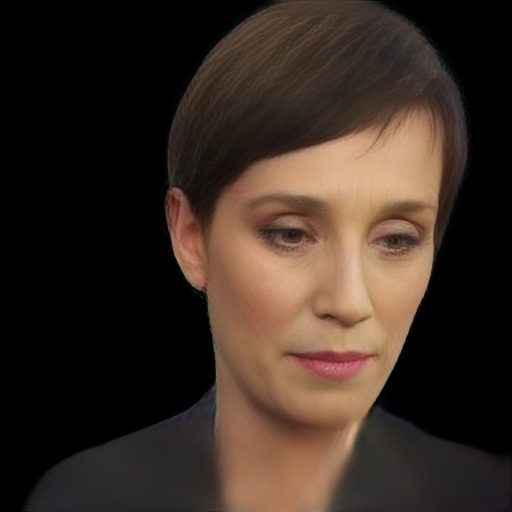} &
\includegraphics[width=0.125\textwidth]{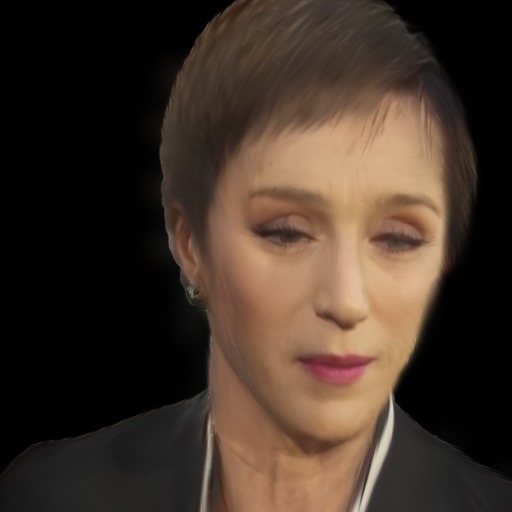} &
\includegraphics[width=0.125\textwidth]{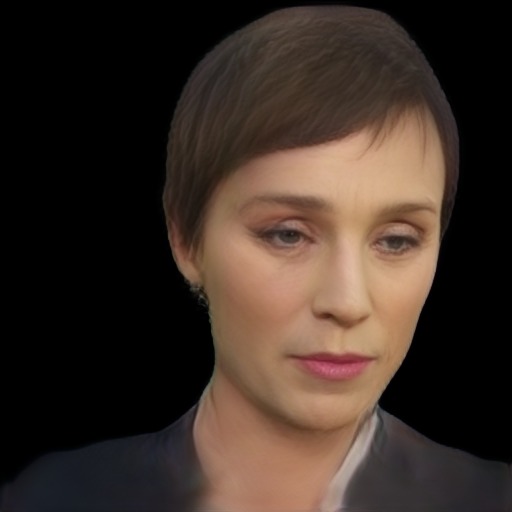}  \\
\includegraphics[width=0.125\textwidth]{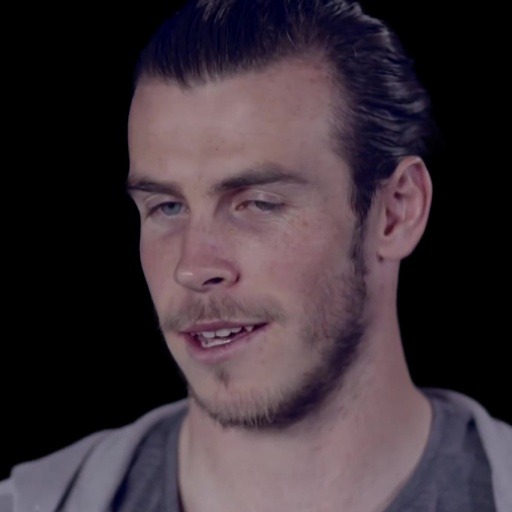} &
\includegraphics[width=0.125\textwidth]{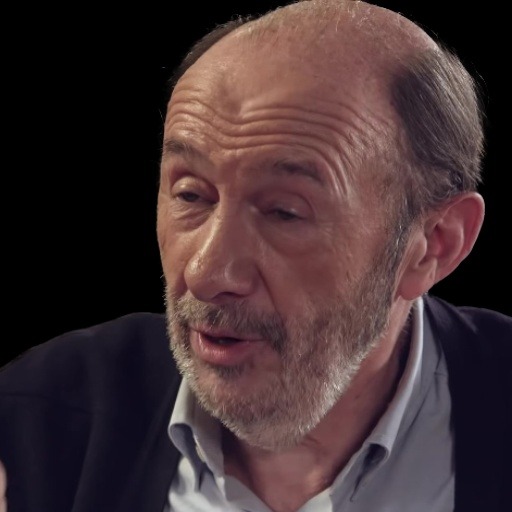} &
\includegraphics[width=0.125\textwidth]{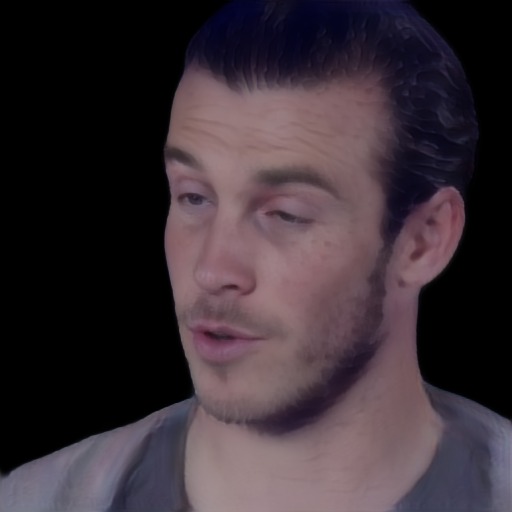} &
\includegraphics[width=0.125\textwidth]{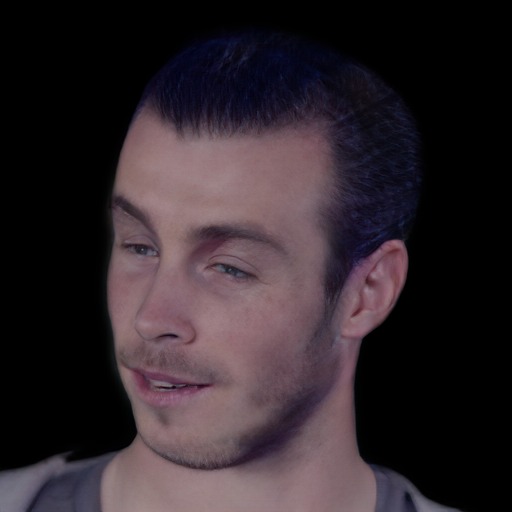} &
\includegraphics[width=0.125\textwidth]{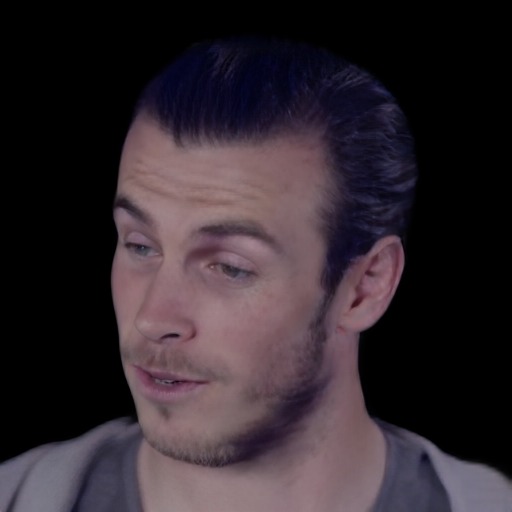} &
\includegraphics[width=0.125\textwidth]{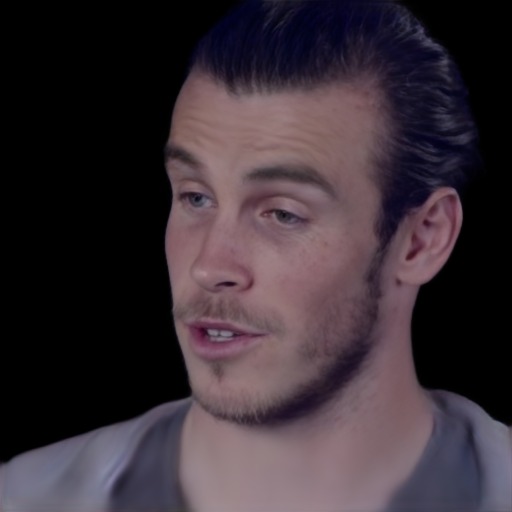} &
\includegraphics[width=0.125\textwidth]{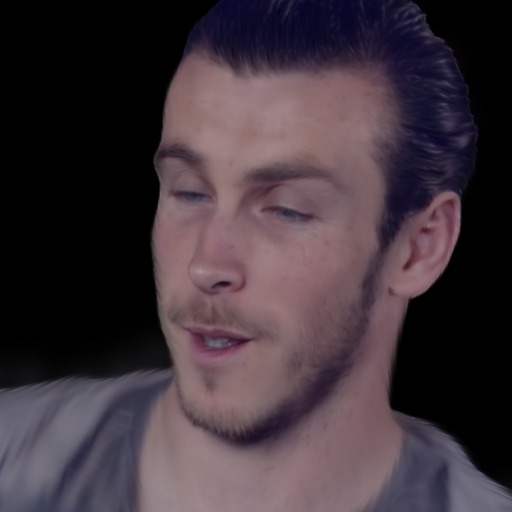} &
\includegraphics[width=0.125\textwidth]{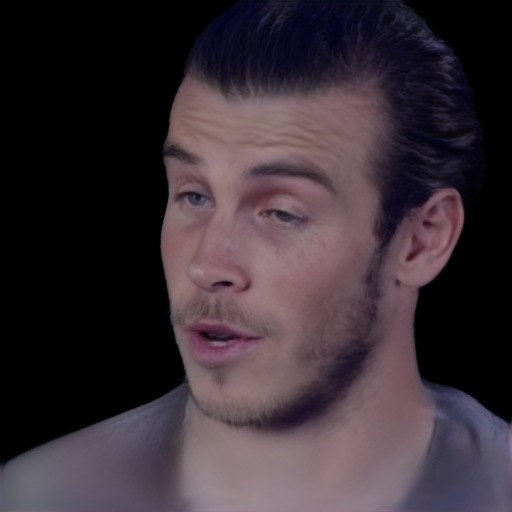}  \\
\includegraphics[width=0.125\textwidth]{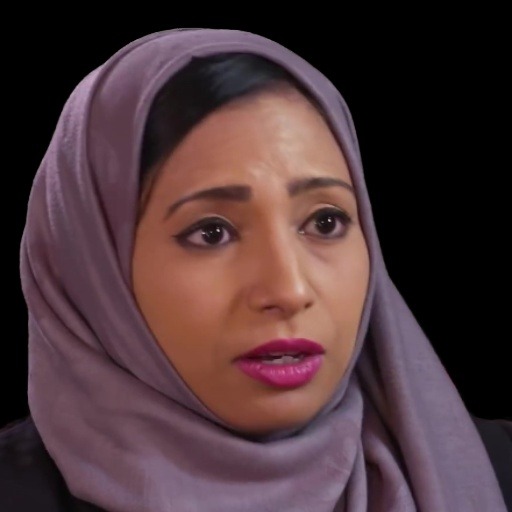} &
\includegraphics[width=0.125\textwidth]{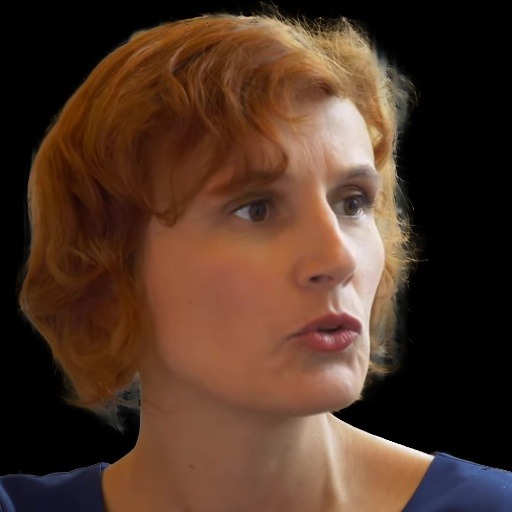} &
\includegraphics[width=0.125\textwidth]{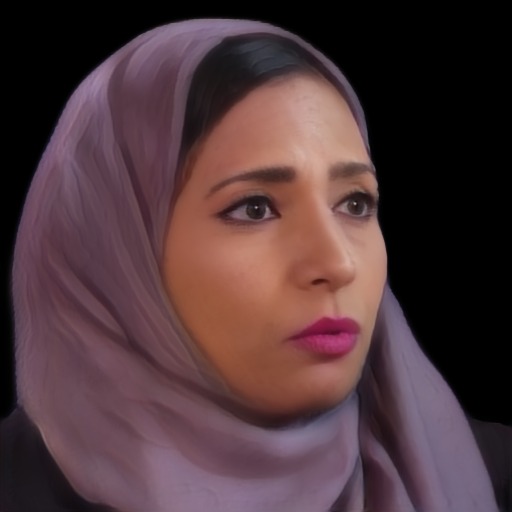} &
\includegraphics[width=0.125\textwidth]{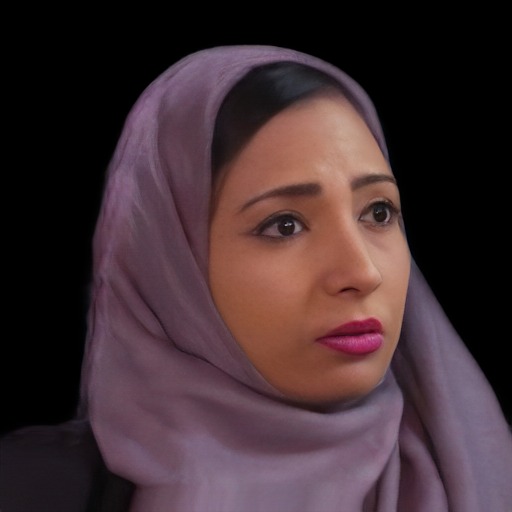} &
\includegraphics[width=0.125\textwidth]{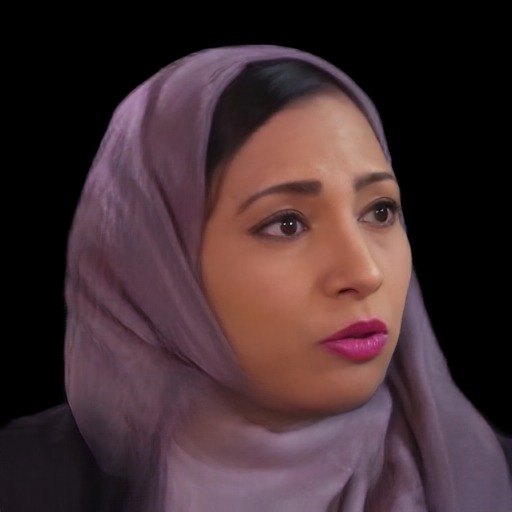} &
\includegraphics[width=0.125\textwidth]{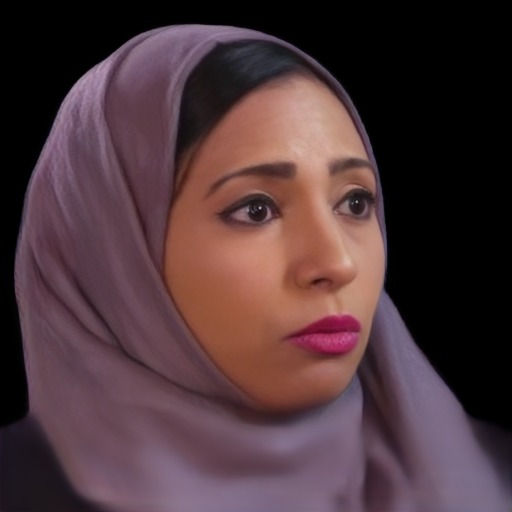} &
\includegraphics[width=0.125\textwidth]{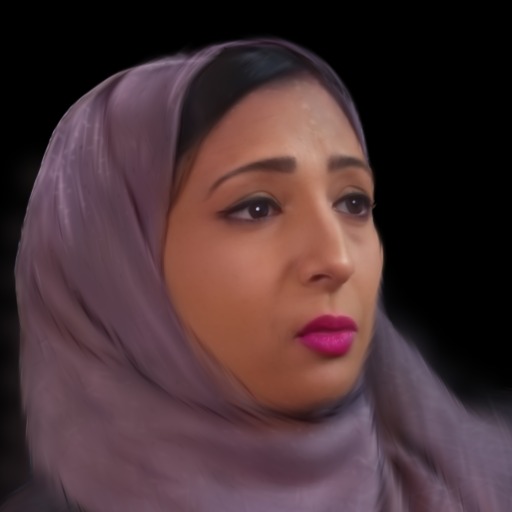} &
\includegraphics[width=0.125\textwidth]{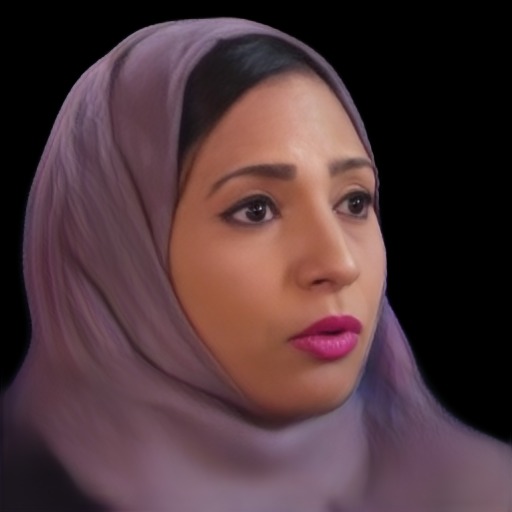}  \\
\includegraphics[width=0.125\textwidth]{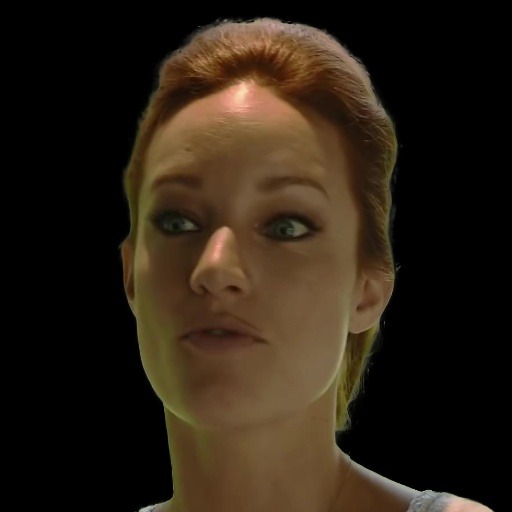} &
\includegraphics[width=0.125\textwidth]{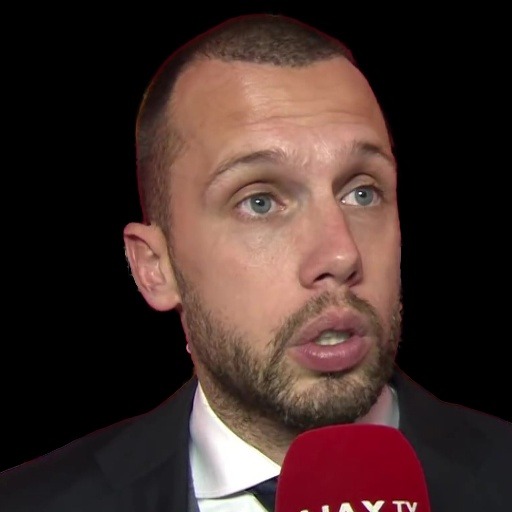} &
\includegraphics[width=0.125\textwidth]{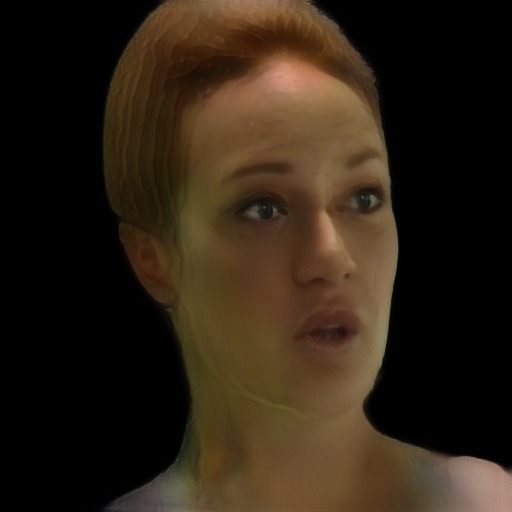} &
\includegraphics[width=0.125\textwidth]{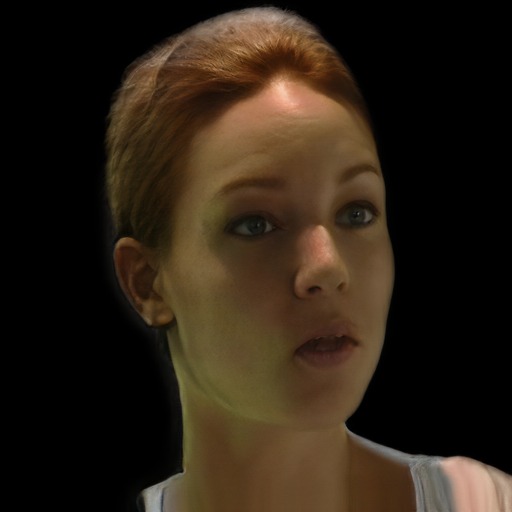} &
\includegraphics[width=0.125\textwidth]{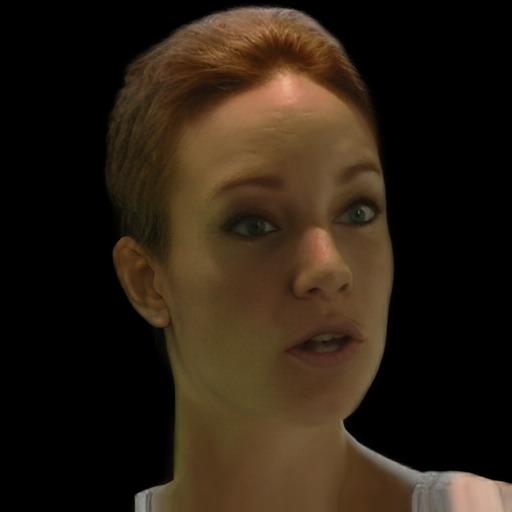} &
\includegraphics[width=0.125\textwidth]{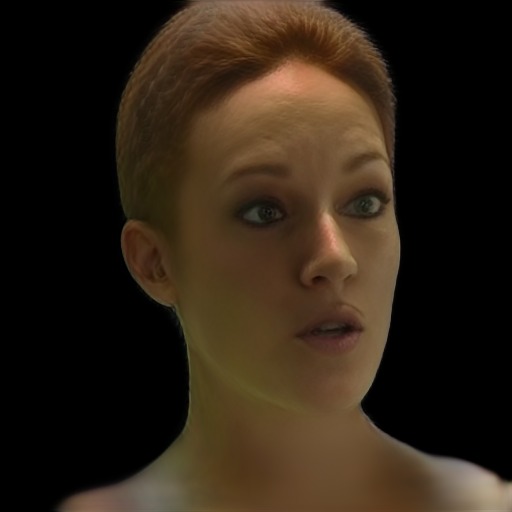} &
\includegraphics[width=0.125\textwidth]{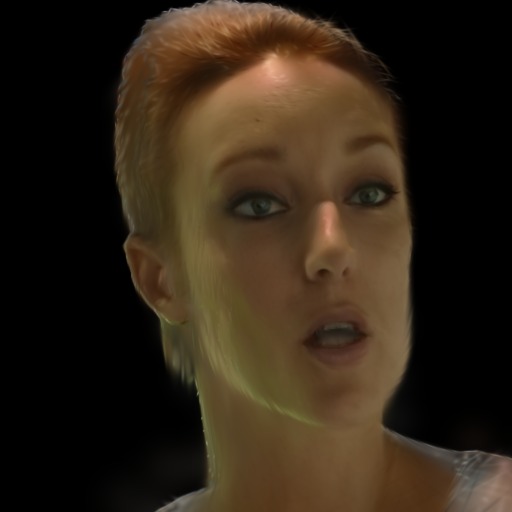} &
\includegraphics[width=0.125\textwidth]{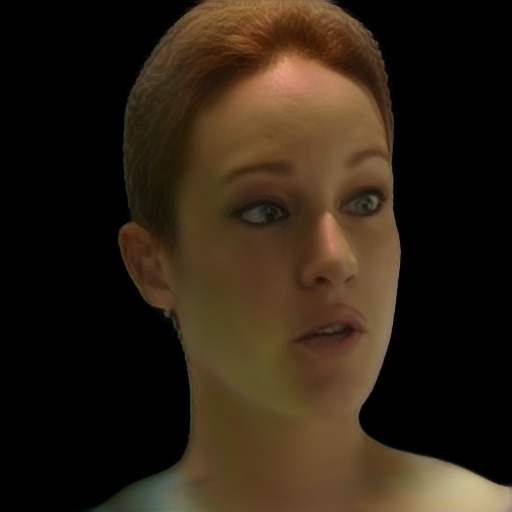}  \\
\includegraphics[width=0.125\textwidth]{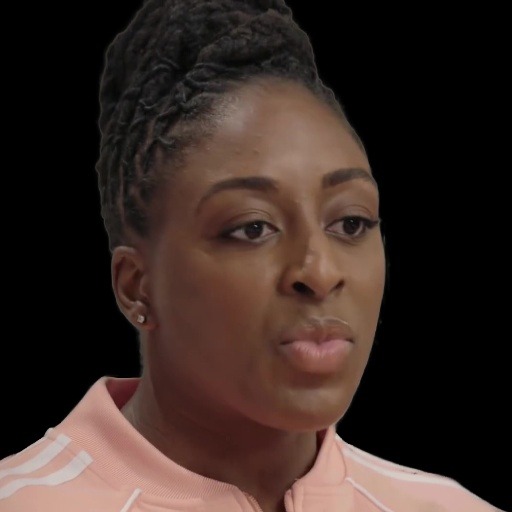} &
\includegraphics[width=0.125\textwidth]{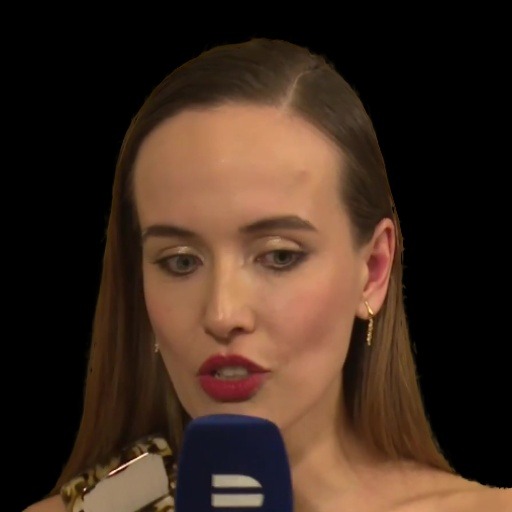} &
\includegraphics[width=0.125\textwidth]{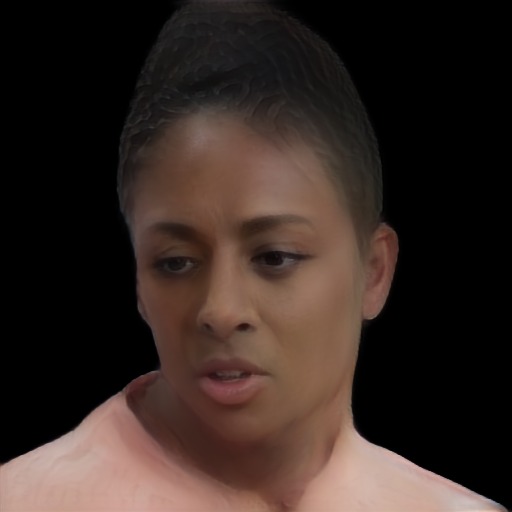} &
\includegraphics[width=0.125\textwidth]{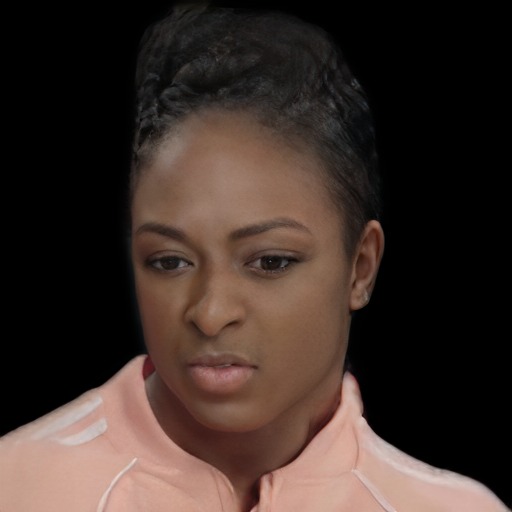} &
\includegraphics[width=0.125\textwidth]{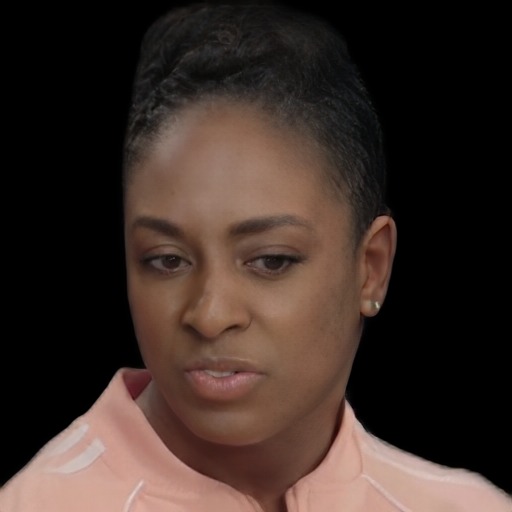} &
\includegraphics[width=0.125\textwidth]{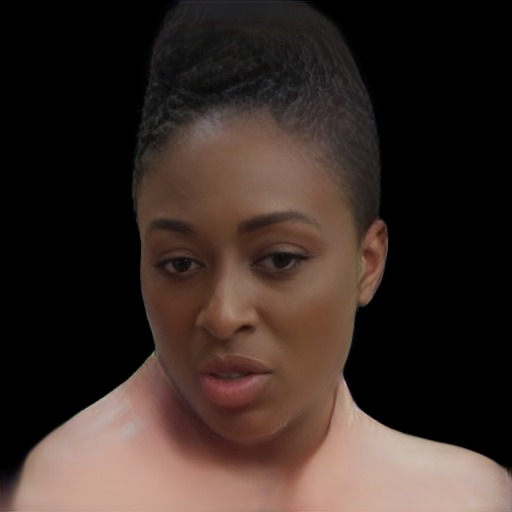} &
\includegraphics[width=0.125\textwidth]{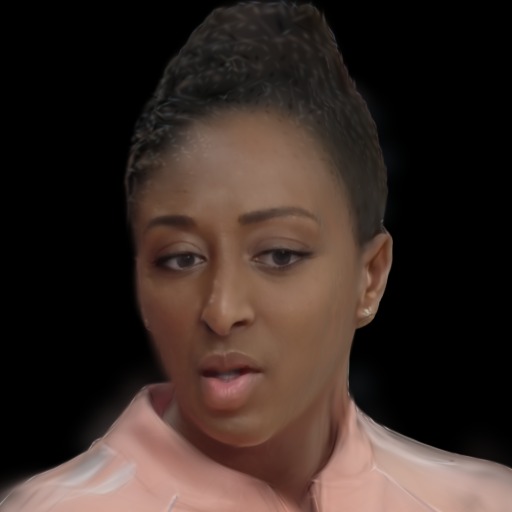} &
\includegraphics[width=0.125\textwidth]{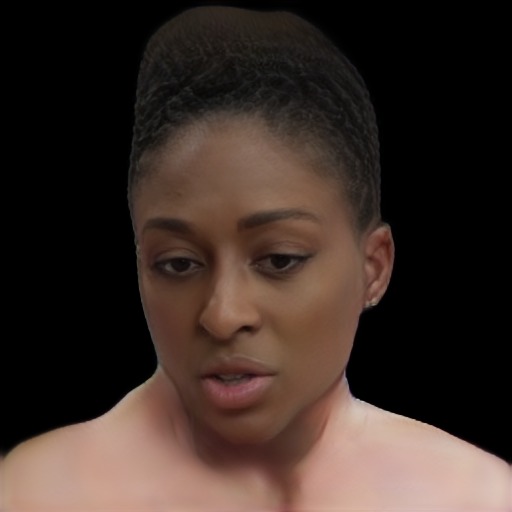}  \\
\includegraphics[width=0.125\textwidth]{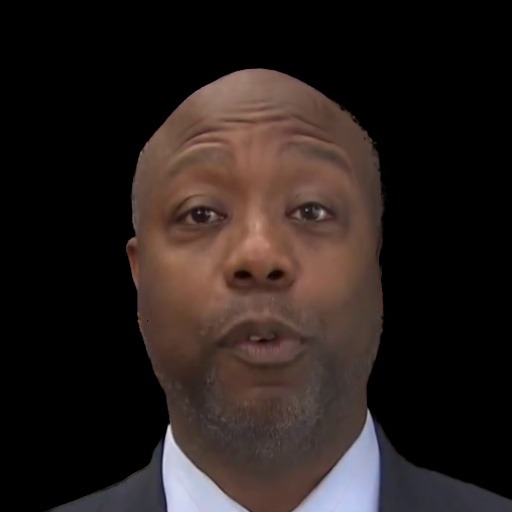} &
\includegraphics[width=0.125\textwidth]{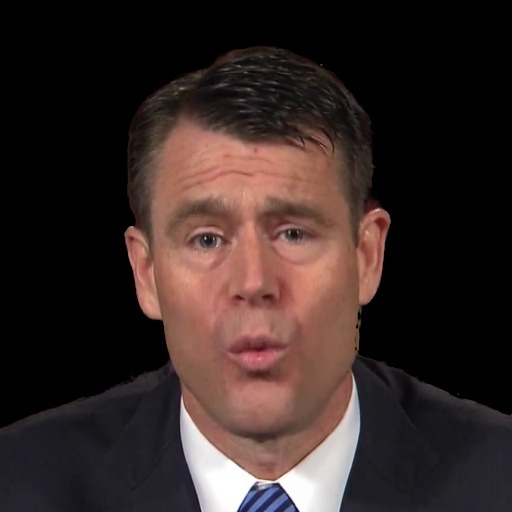} &
\includegraphics[width=0.125\textwidth]{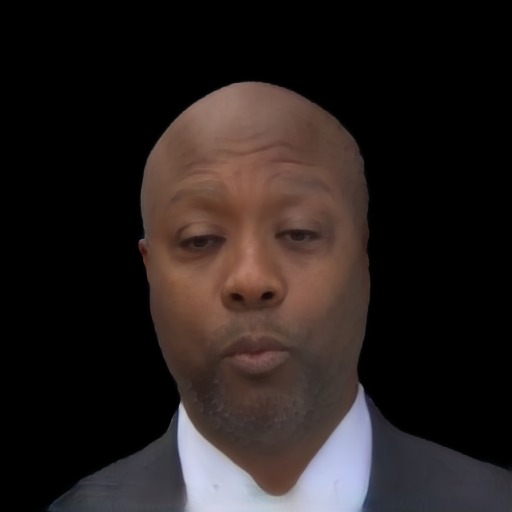} &
\includegraphics[width=0.125\textwidth]{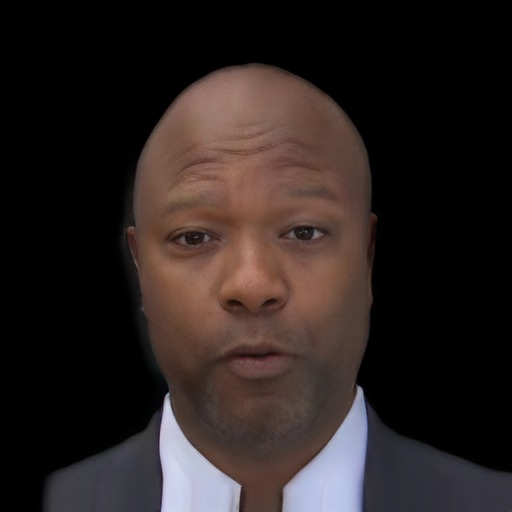} &
\includegraphics[width=0.125\textwidth]{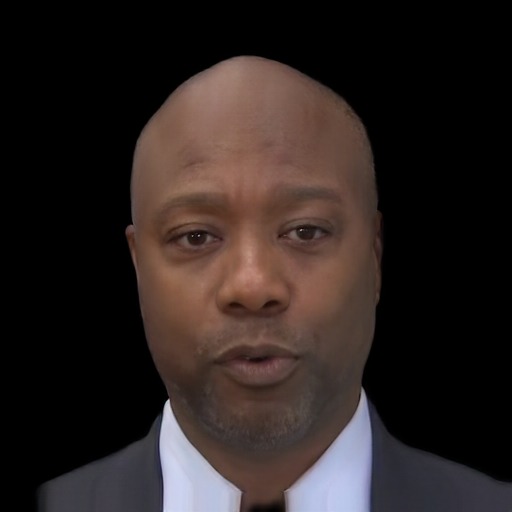} &
\includegraphics[width=0.125\textwidth]{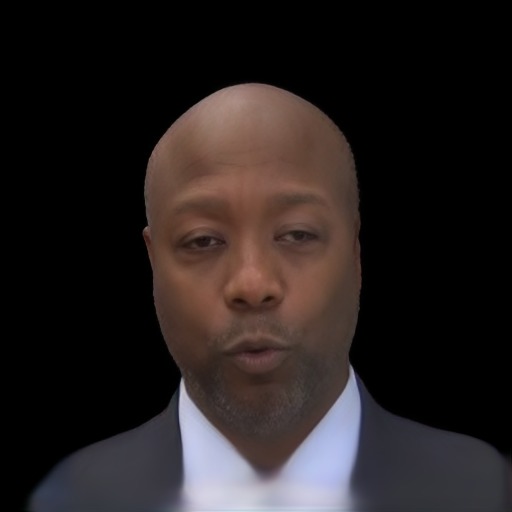} &
\includegraphics[width=0.125\textwidth]{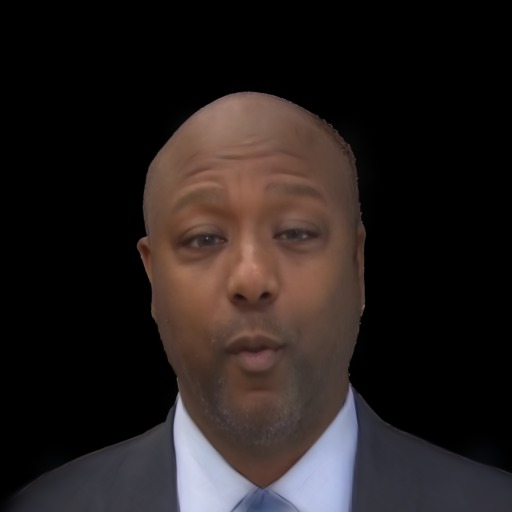} &
\includegraphics[width=0.125\textwidth]{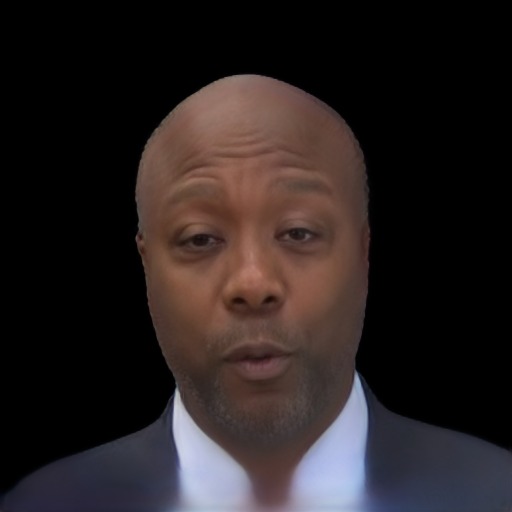}  \\
\includegraphics[width=0.125\textwidth]{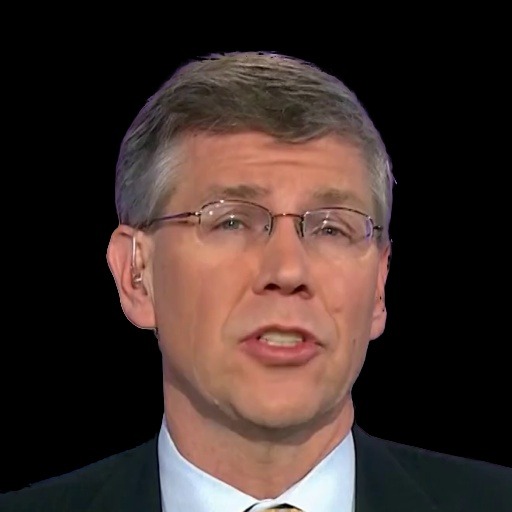} &
\includegraphics[width=0.125\textwidth]{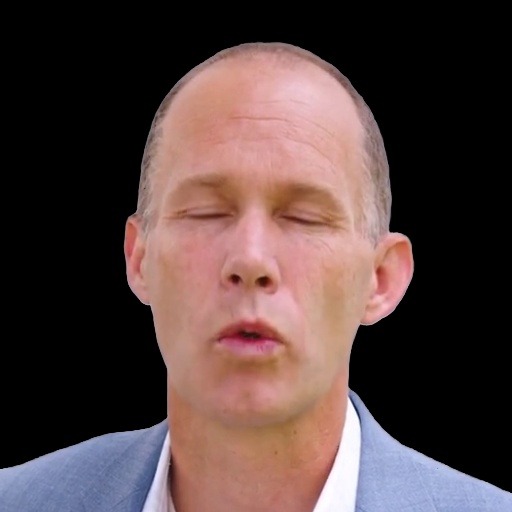} &
\includegraphics[width=0.125\textwidth]{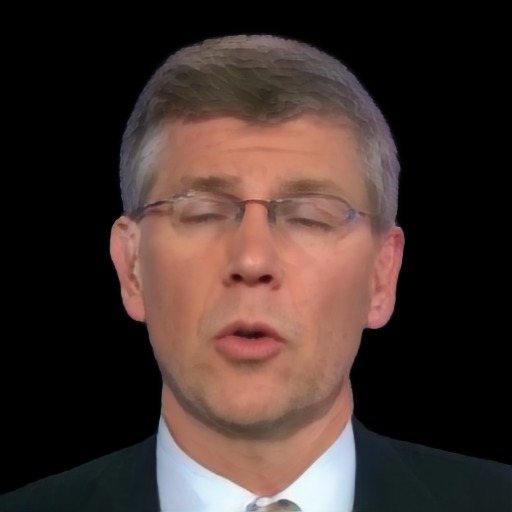} &
\includegraphics[width=0.125\textwidth]{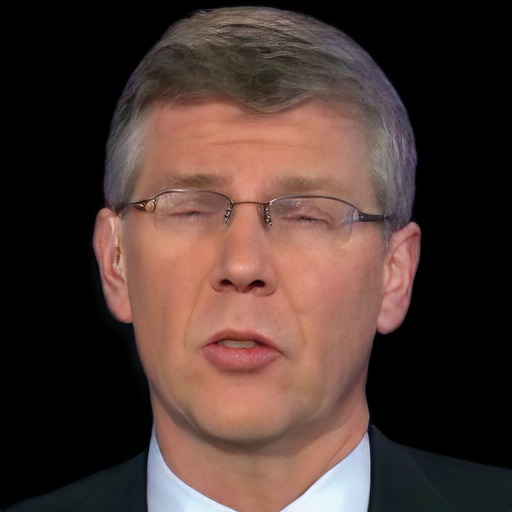} &
\includegraphics[width=0.125\textwidth]{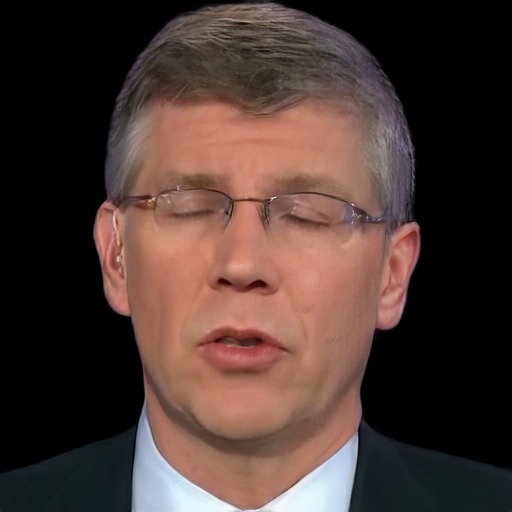} &
\includegraphics[width=0.125\textwidth]{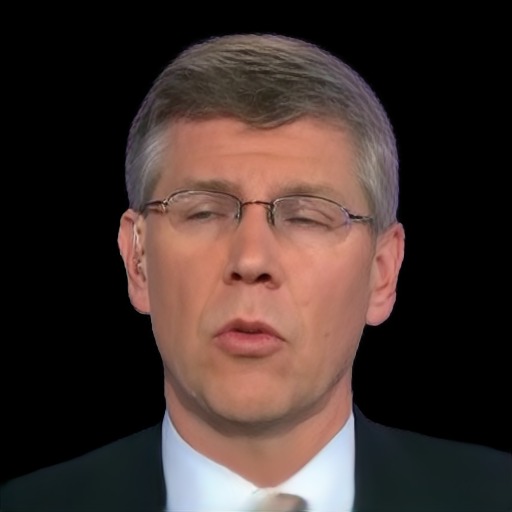} &
\includegraphics[width=0.125\textwidth]{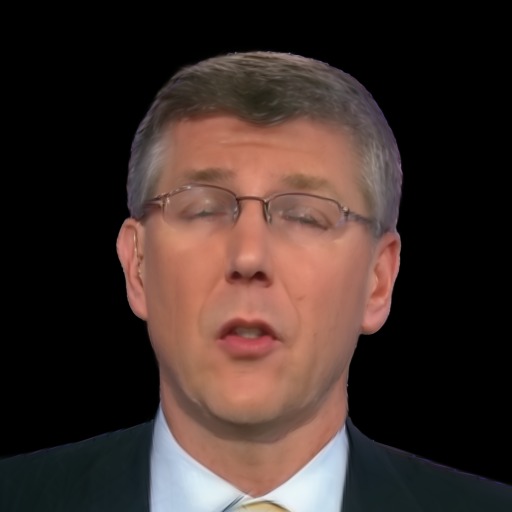} &
\includegraphics[width=0.125\textwidth]{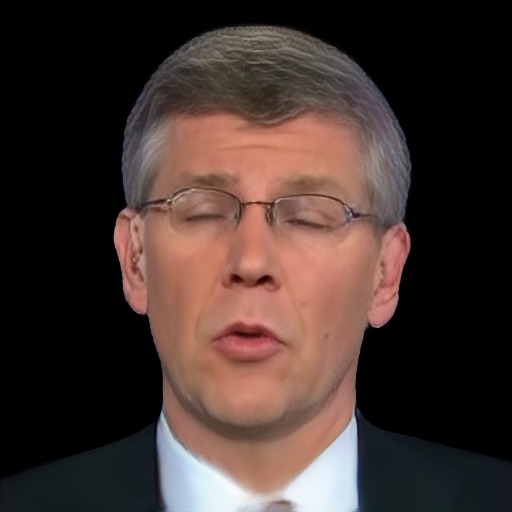}  \\
\includegraphics[width=0.125\textwidth]{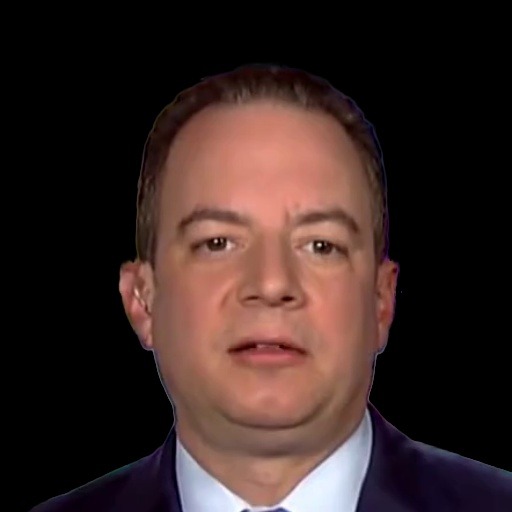} &
\includegraphics[width=0.125\textwidth]{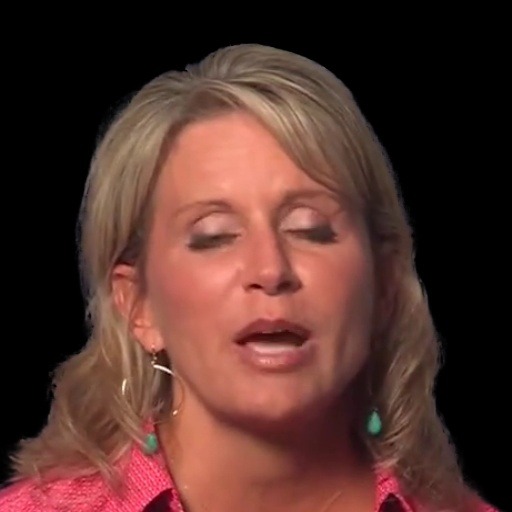} &
\includegraphics[width=0.125\textwidth]{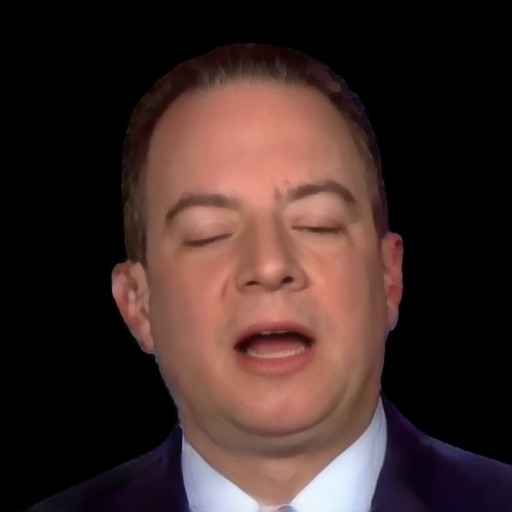} &
\includegraphics[width=0.125\textwidth]{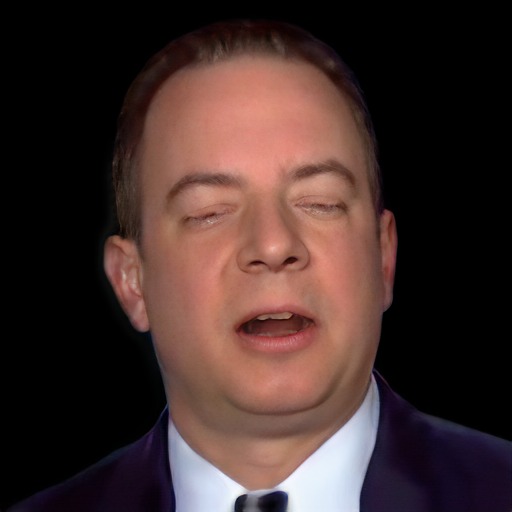} &
\includegraphics[width=0.125\textwidth]{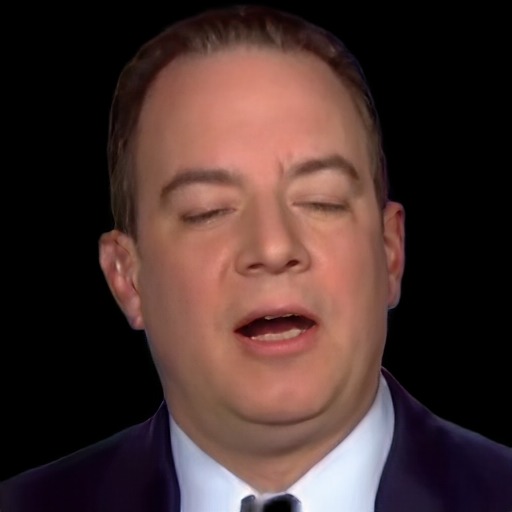} &
\includegraphics[width=0.125\textwidth]{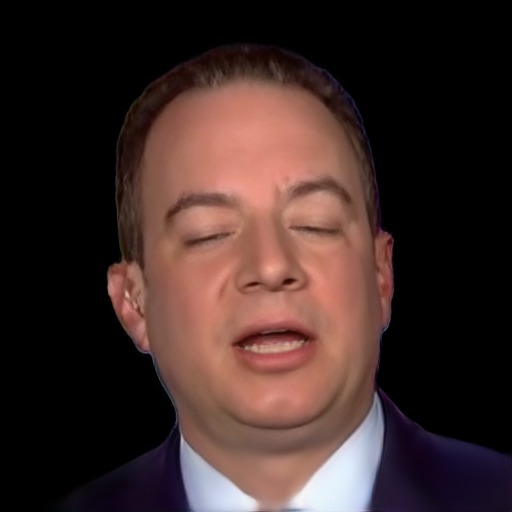} &
\includegraphics[width=0.125\textwidth]{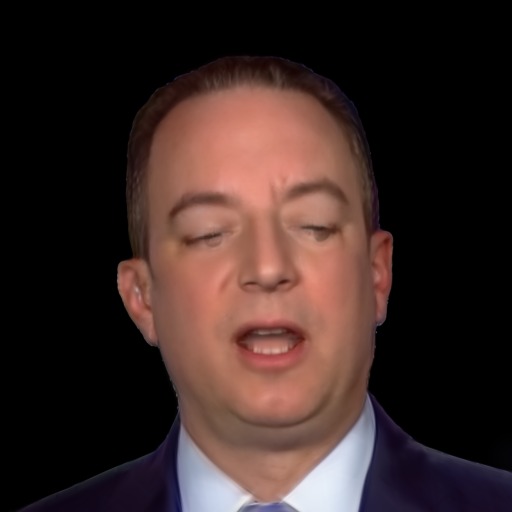} &
\includegraphics[width=0.125\textwidth]{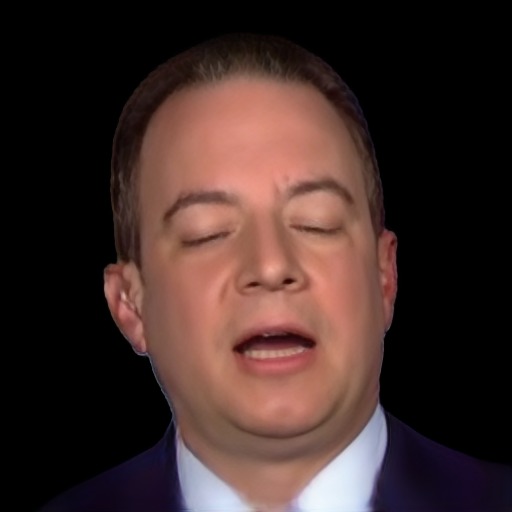}  \\
\includegraphics[width=0.125\textwidth]{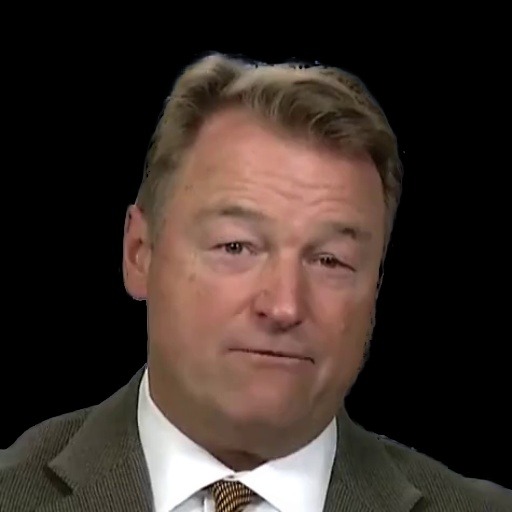} &
\includegraphics[width=0.125\textwidth]{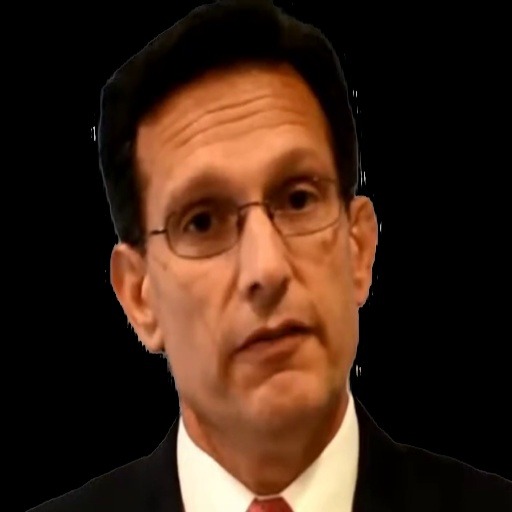} &
\includegraphics[width=0.125\textwidth]{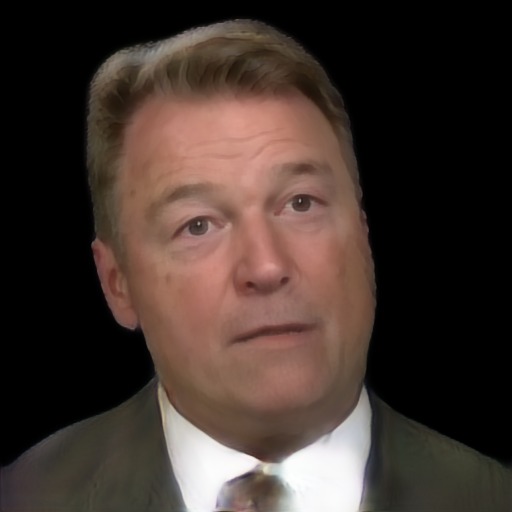} &
\includegraphics[width=0.125\textwidth]{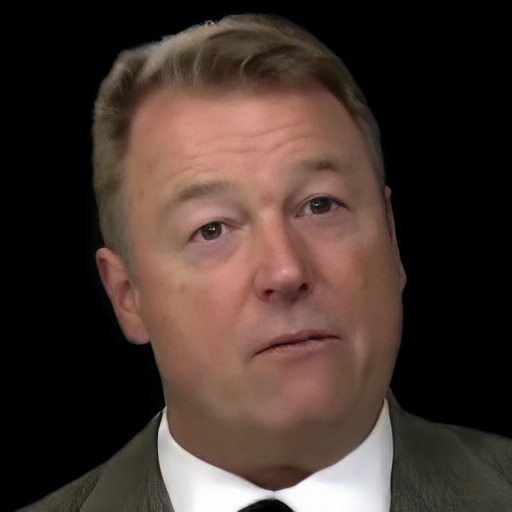} &
\includegraphics[width=0.125\textwidth]{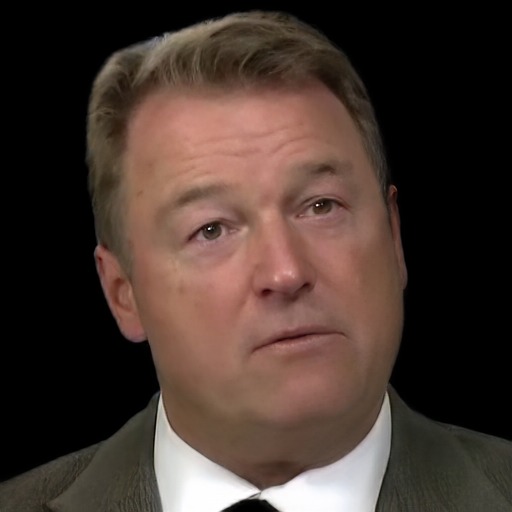} &
\includegraphics[width=0.125\textwidth]{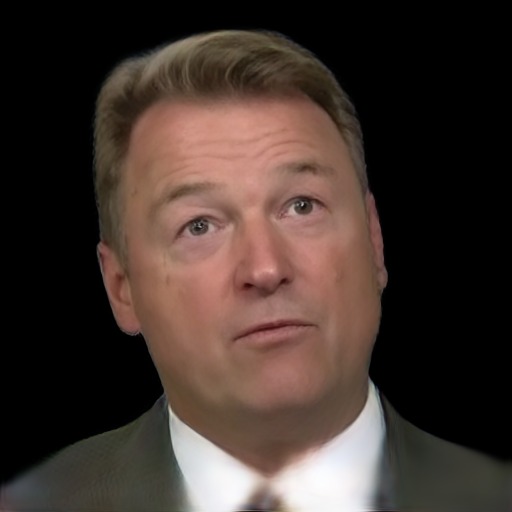} &
\includegraphics[width=0.125\textwidth]{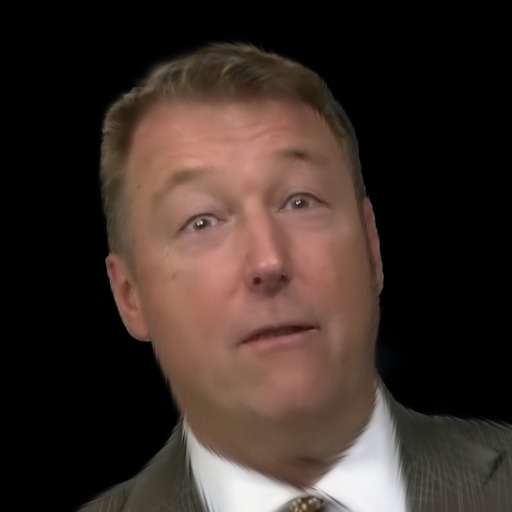} &
\includegraphics[width=0.125\textwidth]{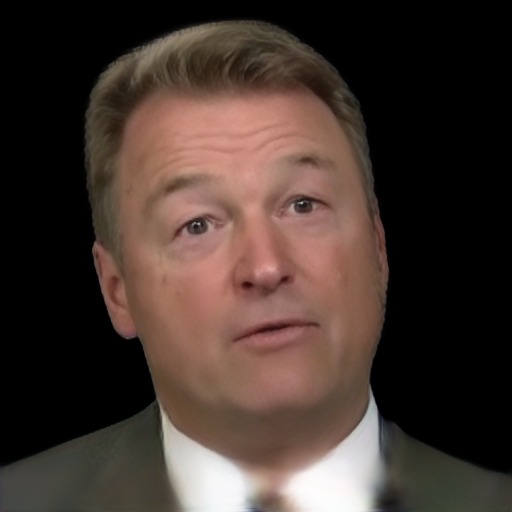}  \\
[5pt]
Source & Driving & GPAvatar & P4D & P4D-v2 & GAG & LAM & Ours \\
\end{tabular}
}
\caption{Additional cross reenactment results on VFHQ and HDTF datasets.}
\label{fig:supp-cross}
\vspace{-5mm}
\end{figure}

\begin{figure}[!p]
\centering
\setlength{\tabcolsep}{0pt}
\renewcommand{\arraystretch}{0}
\resizebox{\textwidth}{!}{
\begin{tabular}{@{}cccccccc@{}}
\includegraphics[width=0.125\textwidth]{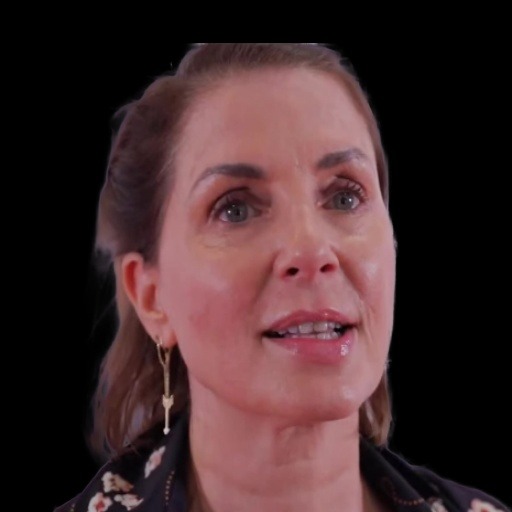} &
\includegraphics[width=0.125\textwidth]{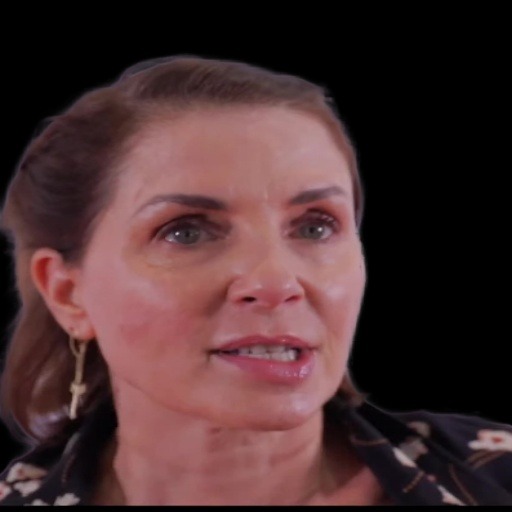} &
\includegraphics[width=0.125\textwidth]{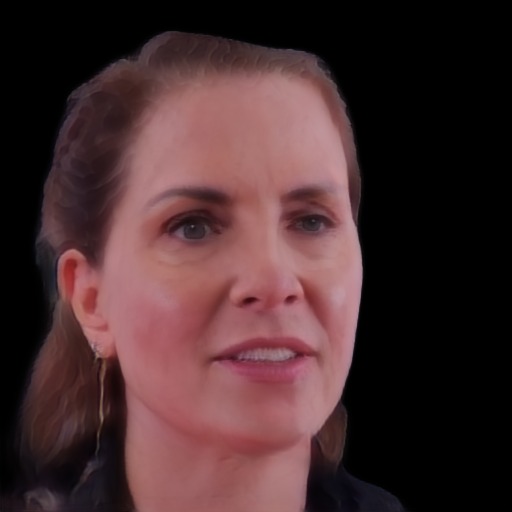} &
\includegraphics[width=0.125\textwidth]{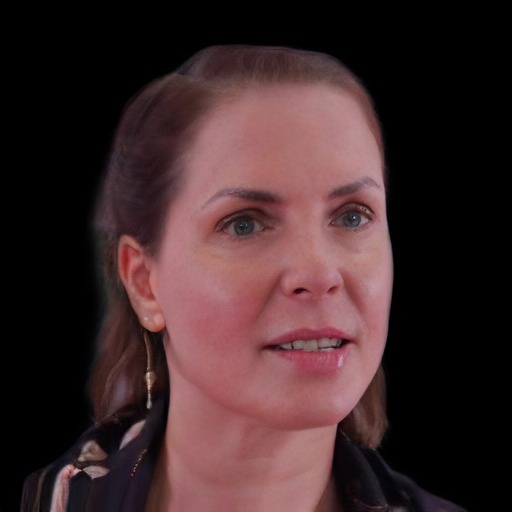} &
\includegraphics[width=0.125\textwidth]{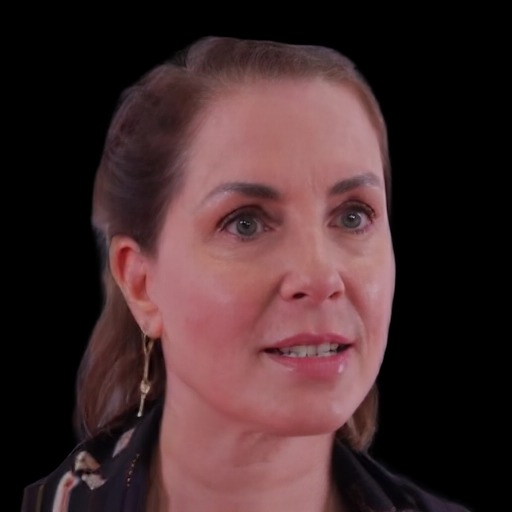} &
\includegraphics[width=0.125\textwidth]{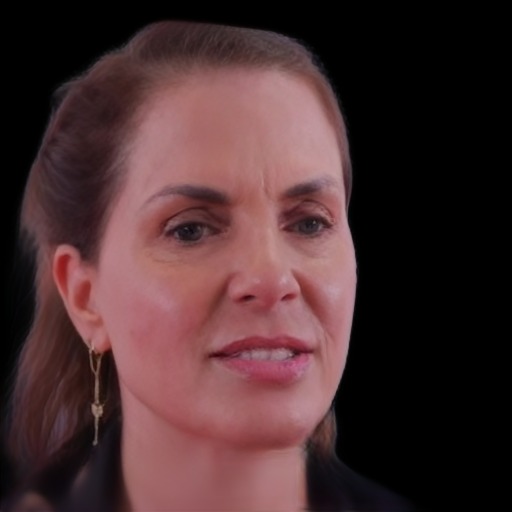} &
\includegraphics[width=0.125\textwidth]{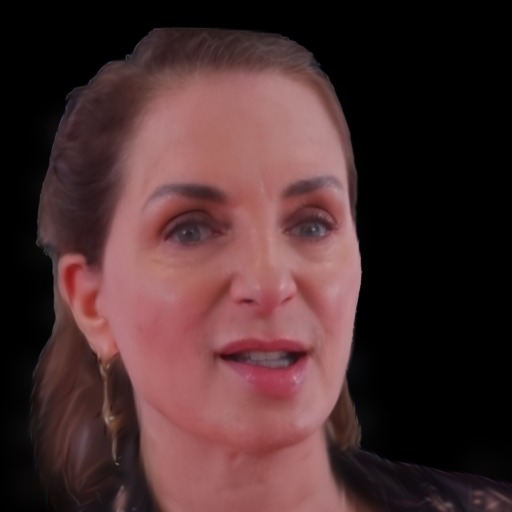} &
\includegraphics[width=0.125\textwidth]{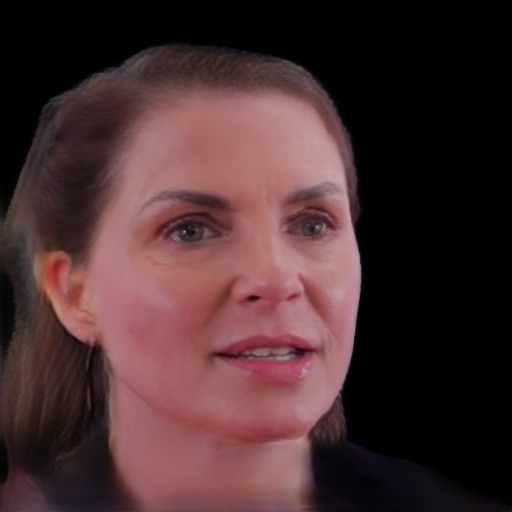}  \\
\includegraphics[width=0.125\textwidth]{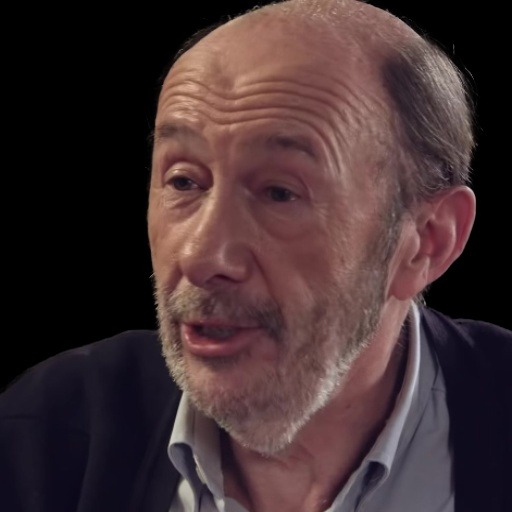} &
\includegraphics[width=0.125\textwidth]{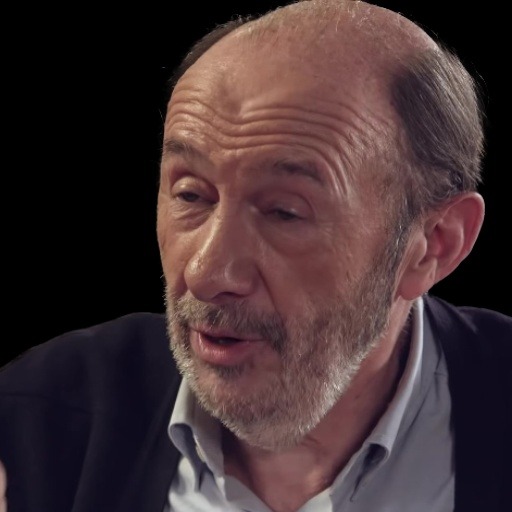} &
\includegraphics[width=0.125\textwidth]{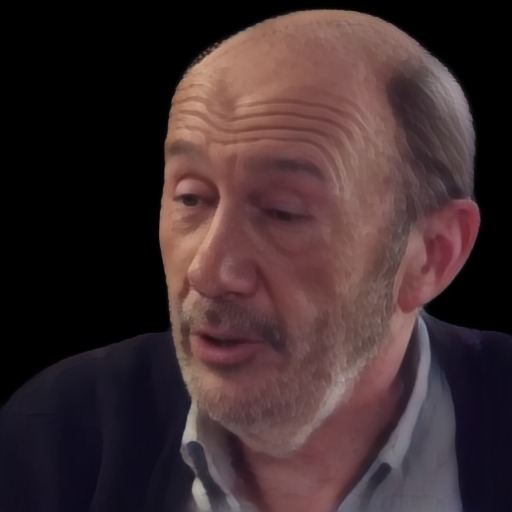} &
\includegraphics[width=0.125\textwidth]{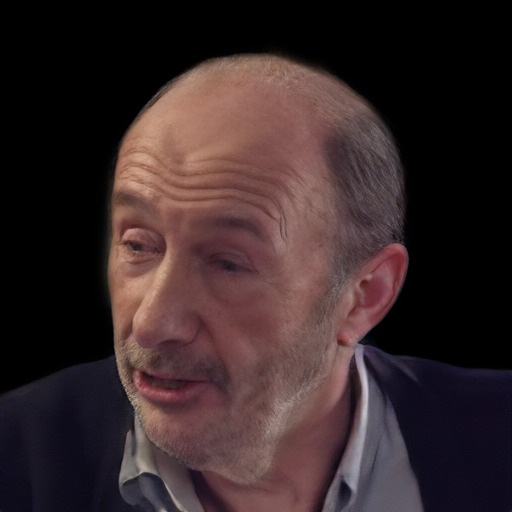} &
\includegraphics[width=0.125\textwidth]{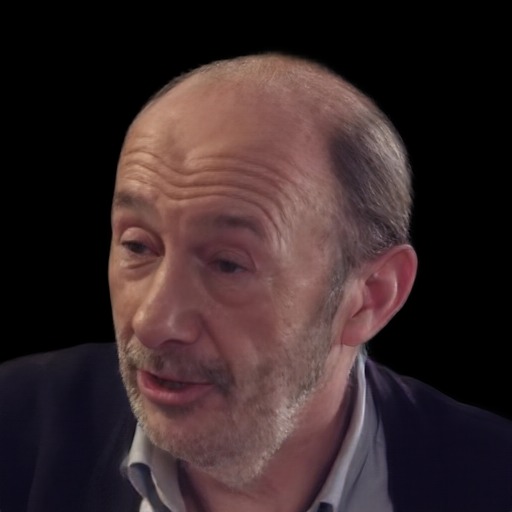} &
\includegraphics[width=0.125\textwidth]{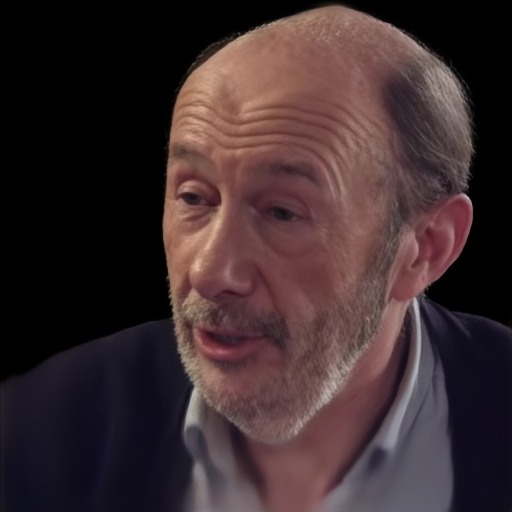} &
\includegraphics[width=0.125\textwidth]{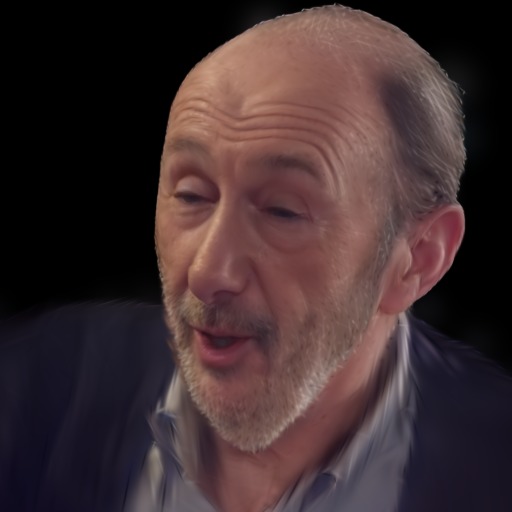} &
\includegraphics[width=0.125\textwidth]{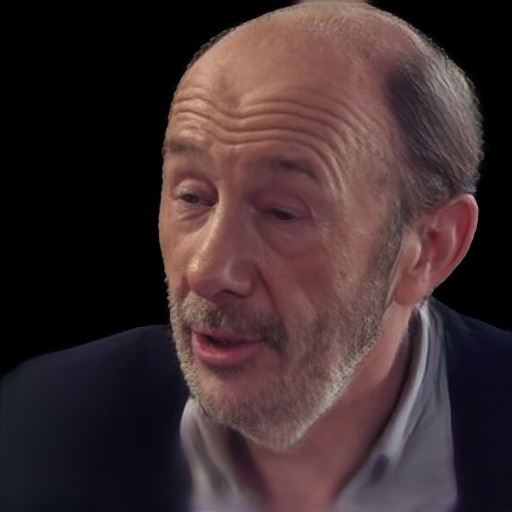}  \\
\includegraphics[width=0.125\textwidth]{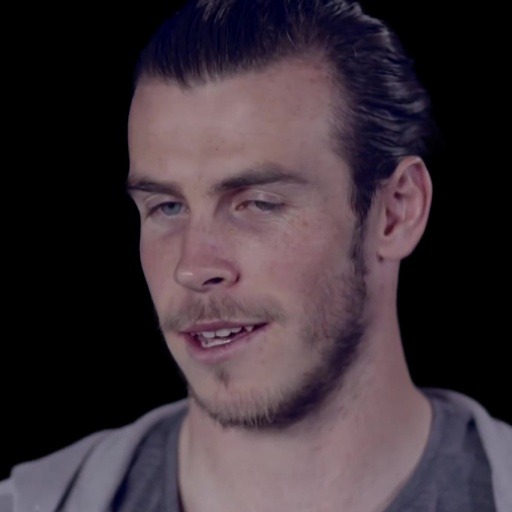} &
\includegraphics[width=0.125\textwidth]{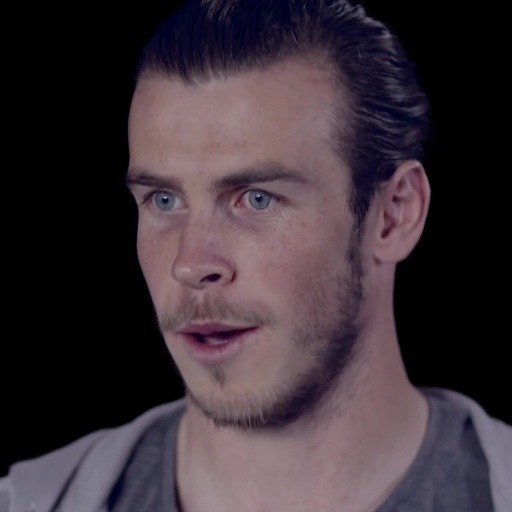} &
\includegraphics[width=0.125\textwidth]{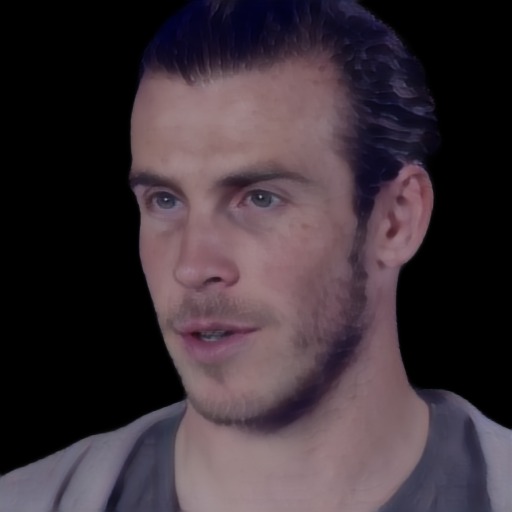} &
\includegraphics[width=0.125\textwidth]{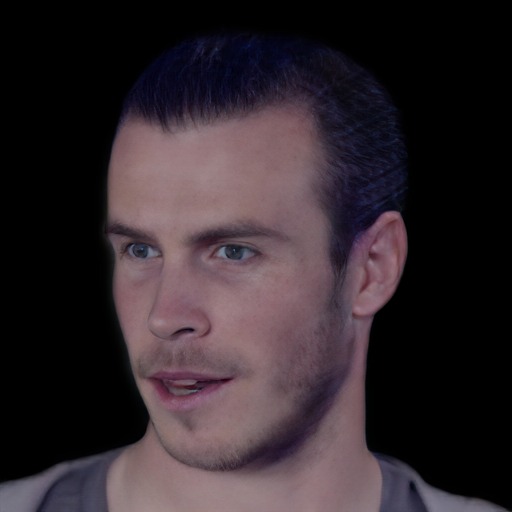} &
\includegraphics[width=0.125\textwidth]{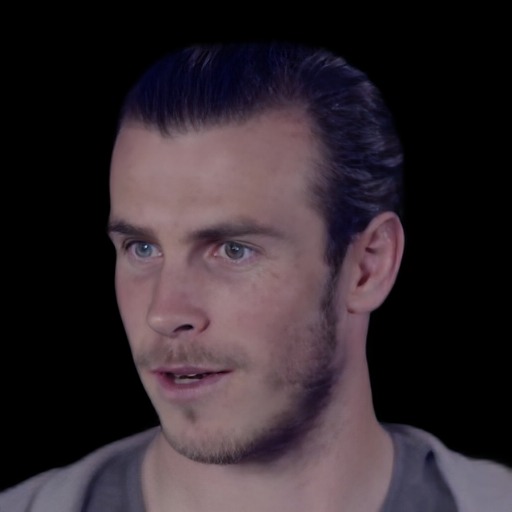} &
\includegraphics[width=0.125\textwidth]{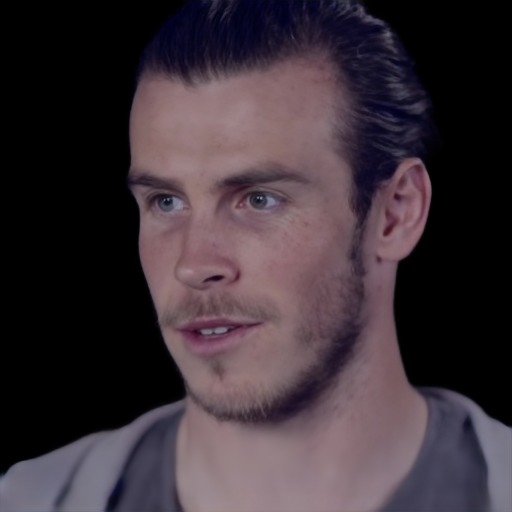} &
\includegraphics[width=0.125\textwidth]{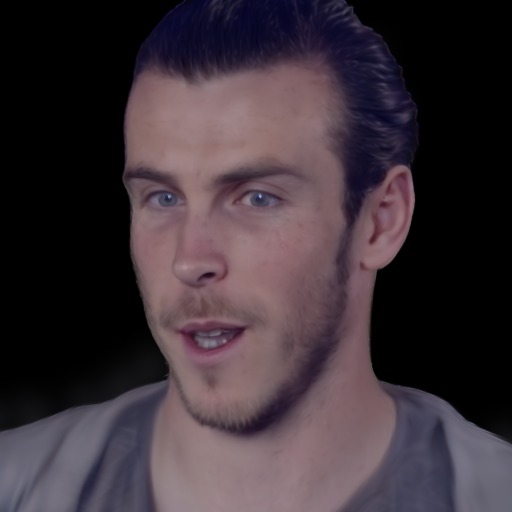} &
\includegraphics[width=0.125\textwidth]{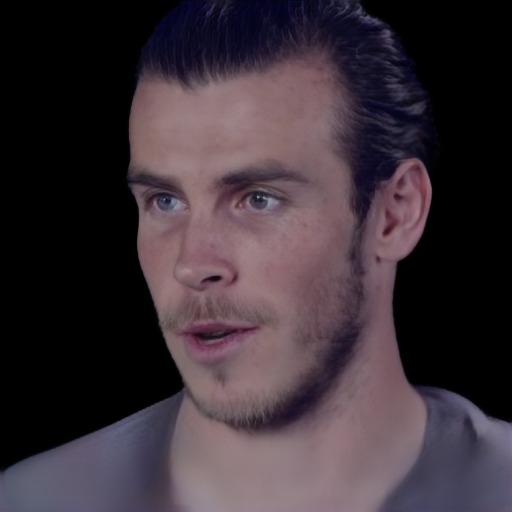}  \\
\includegraphics[width=0.125\textwidth]{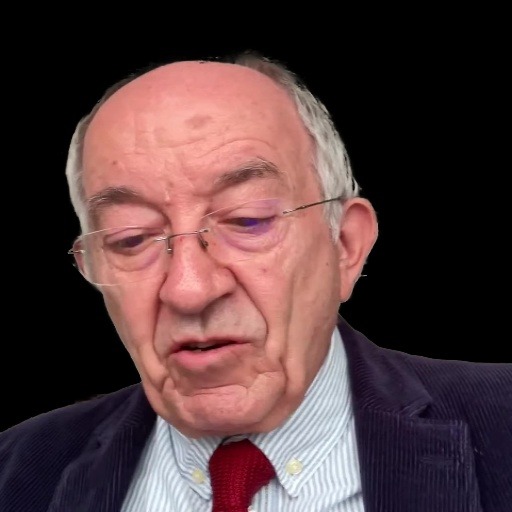} &
\includegraphics[width=0.125\textwidth]{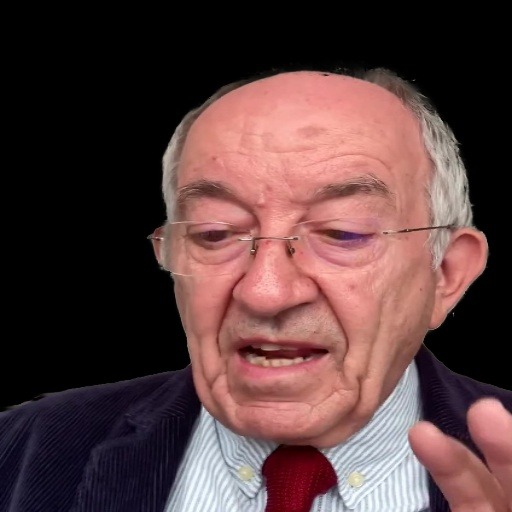} &
\includegraphics[width=0.125\textwidth]{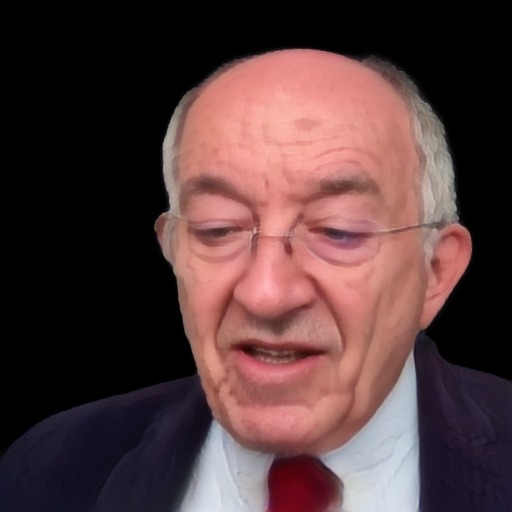} &
\includegraphics[width=0.125\textwidth]{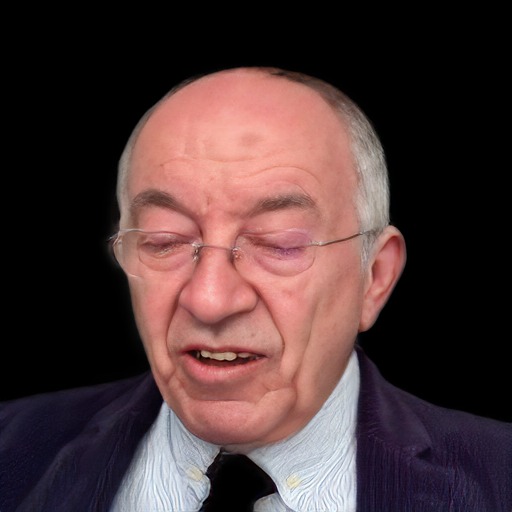} &
\includegraphics[width=0.125\textwidth]{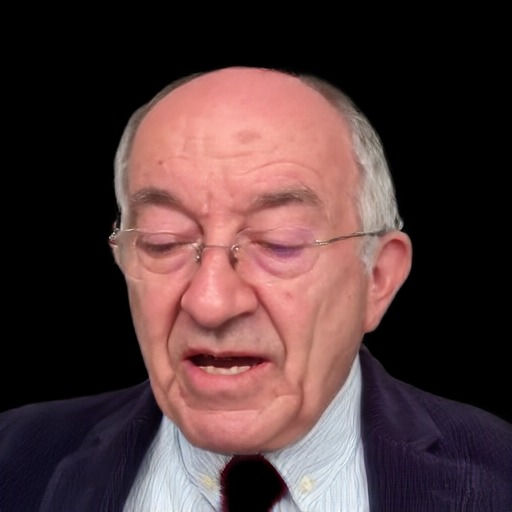} &
\includegraphics[width=0.125\textwidth]{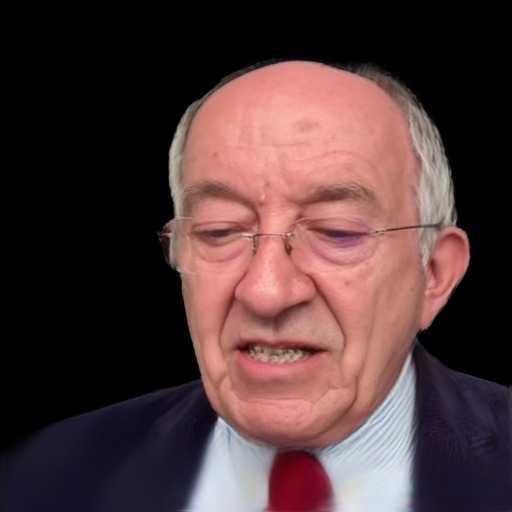} &
\includegraphics[width=0.125\textwidth]{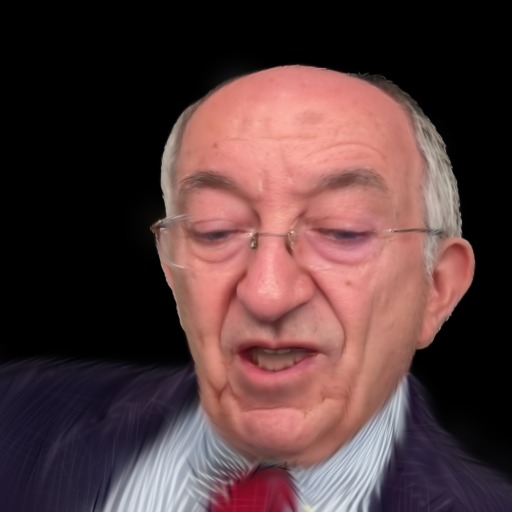} &
\includegraphics[width=0.125\textwidth]{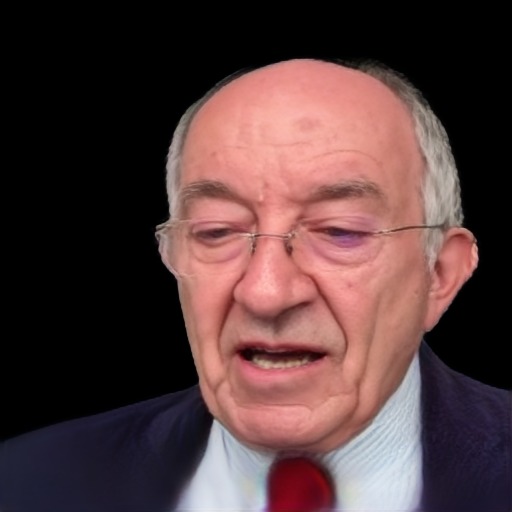}  \\
\includegraphics[width=0.125\textwidth]{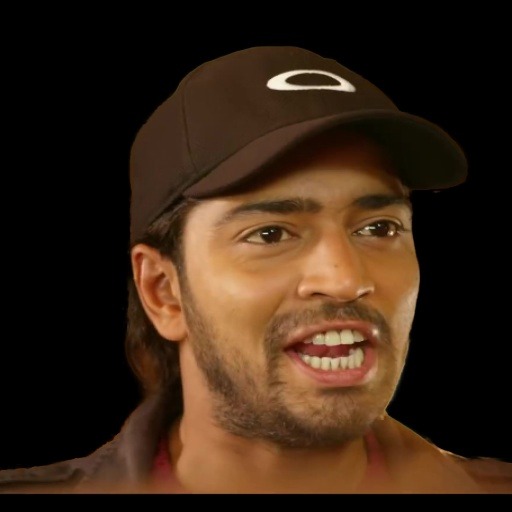} &
\includegraphics[width=0.125\textwidth]{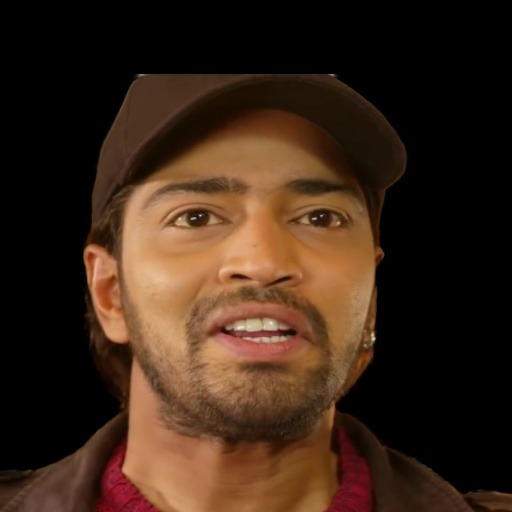} &
\includegraphics[width=0.125\textwidth]{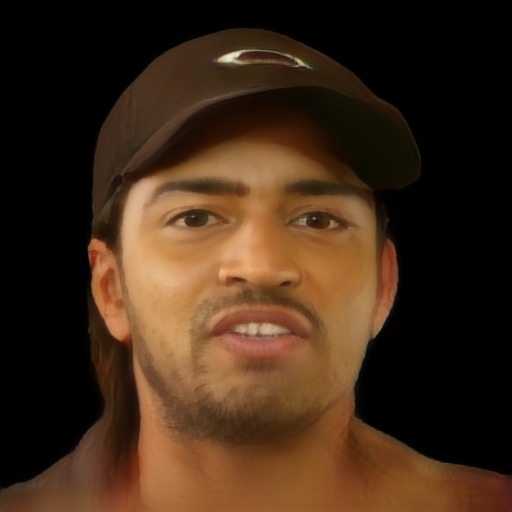} &
\includegraphics[width=0.125\textwidth]{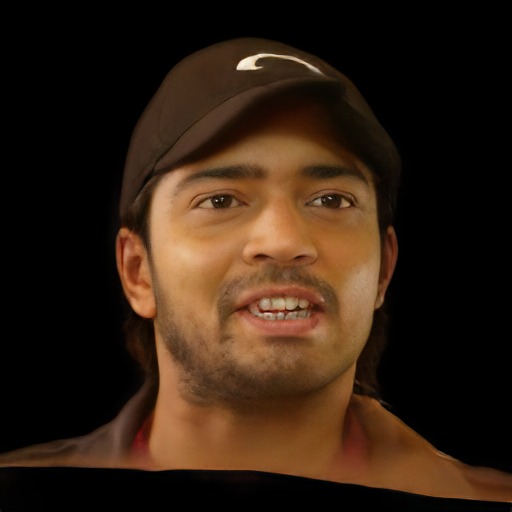} &
\includegraphics[width=0.125\textwidth]{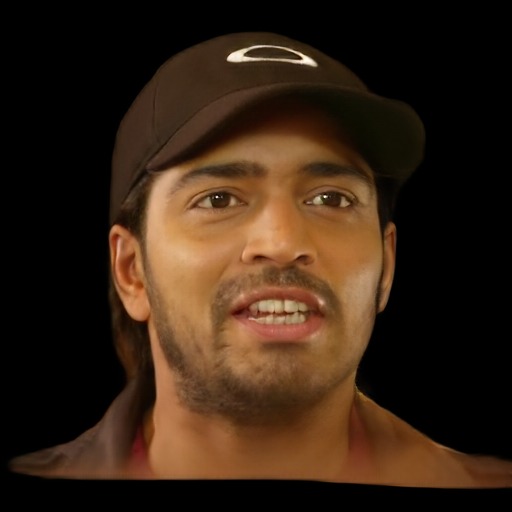} &
\includegraphics[width=0.125\textwidth]{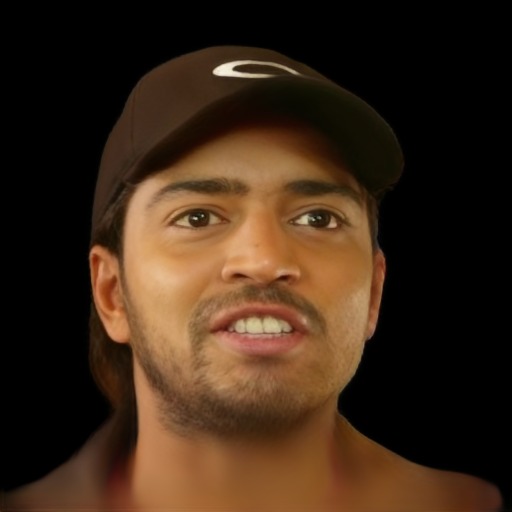} &
\includegraphics[width=0.125\textwidth]{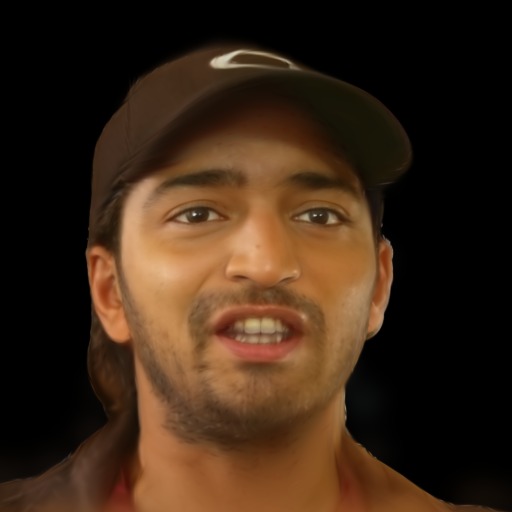} &
\includegraphics[width=0.125\textwidth]{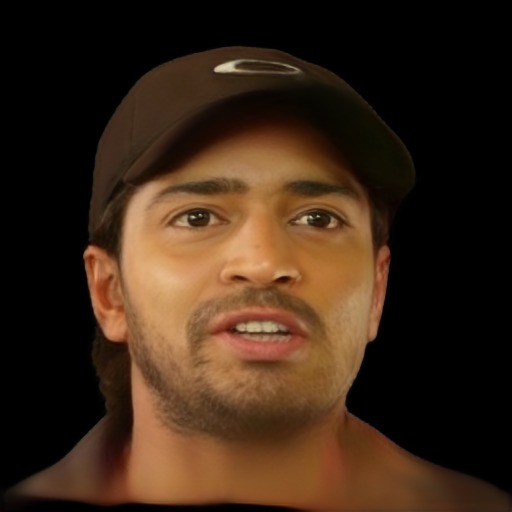}  \\
\includegraphics[width=0.125\textwidth]{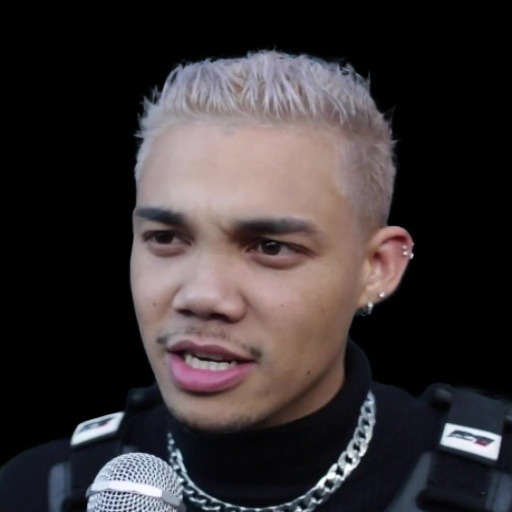} &
\includegraphics[width=0.125\textwidth]{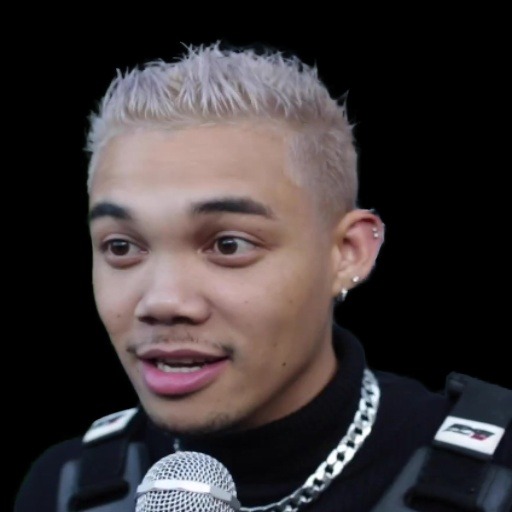} &
\includegraphics[width=0.125\textwidth]{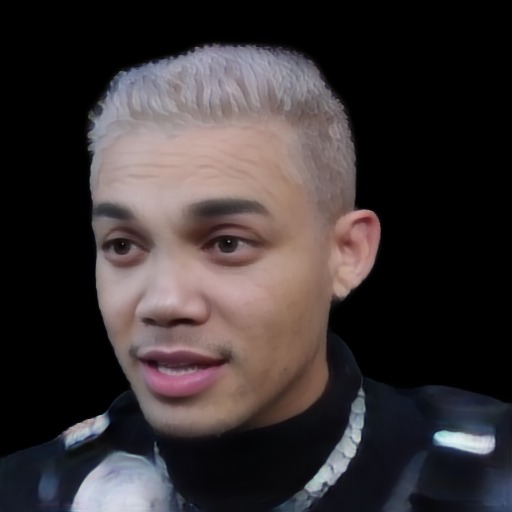} &
\includegraphics[width=0.125\textwidth]{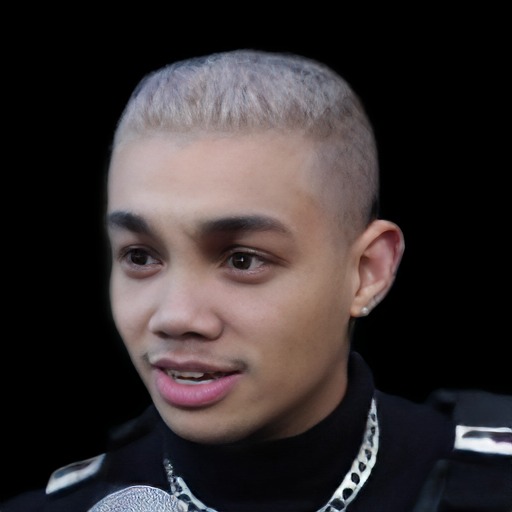} &
\includegraphics[width=0.125\textwidth]{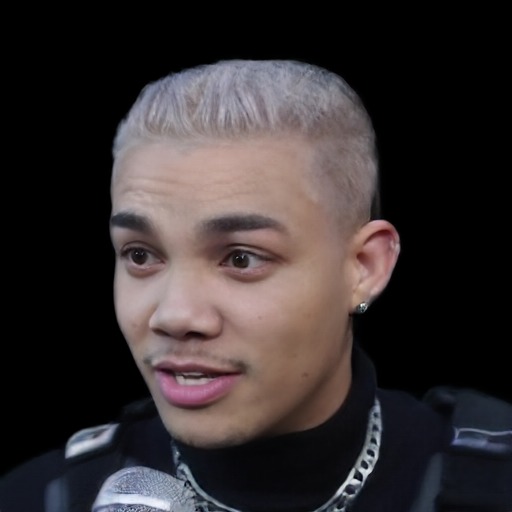} &
\includegraphics[width=0.125\textwidth]{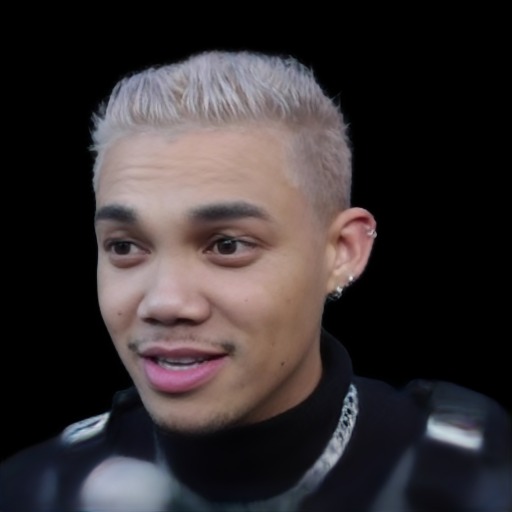} &
\includegraphics[width=0.125\textwidth]{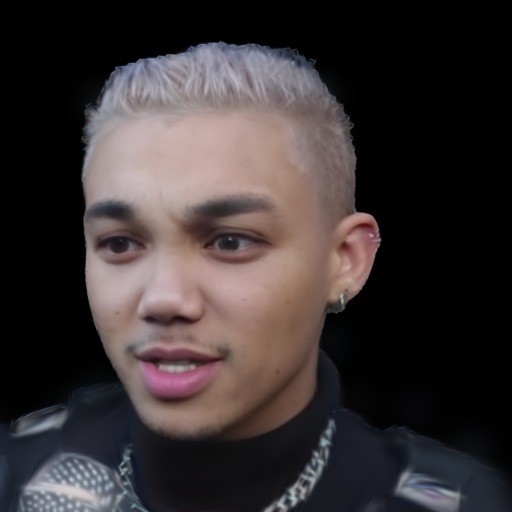} &
\includegraphics[width=0.125\textwidth]{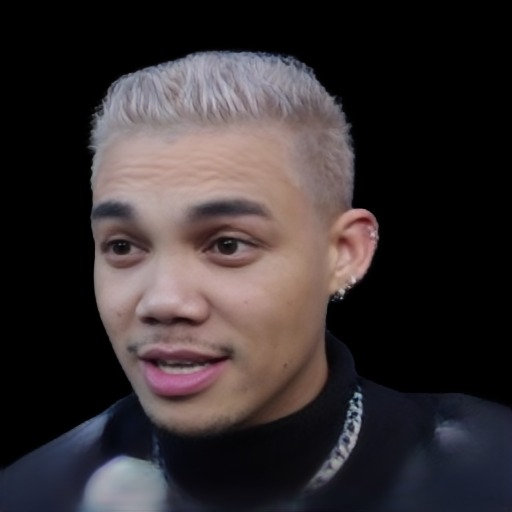}  \\
\includegraphics[width=0.125\textwidth]{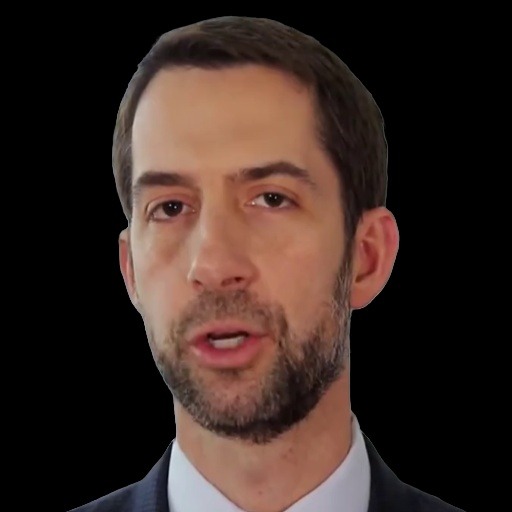} &
\includegraphics[width=0.125\textwidth]{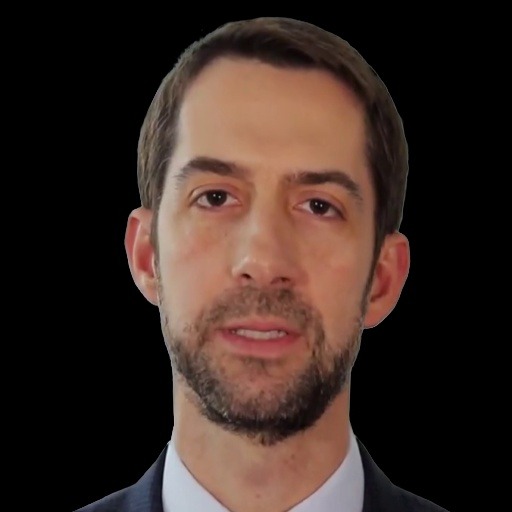} &
\includegraphics[width=0.125\textwidth]{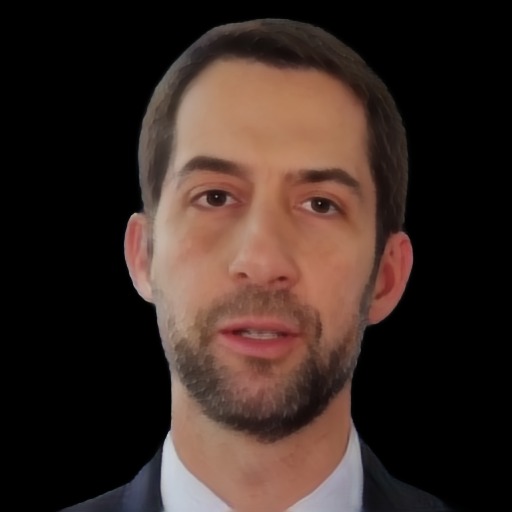} &
\includegraphics[width=0.125\textwidth]{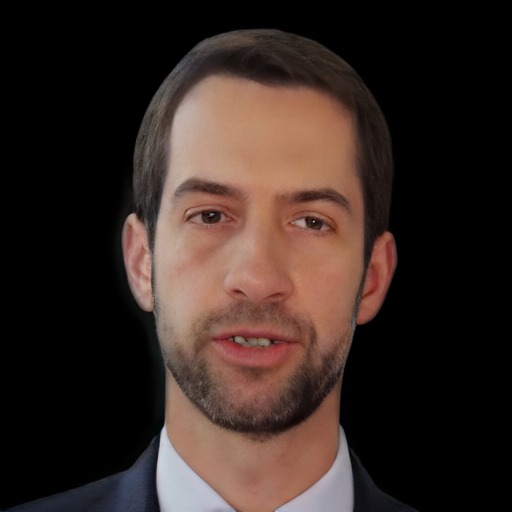} &
\includegraphics[width=0.125\textwidth]{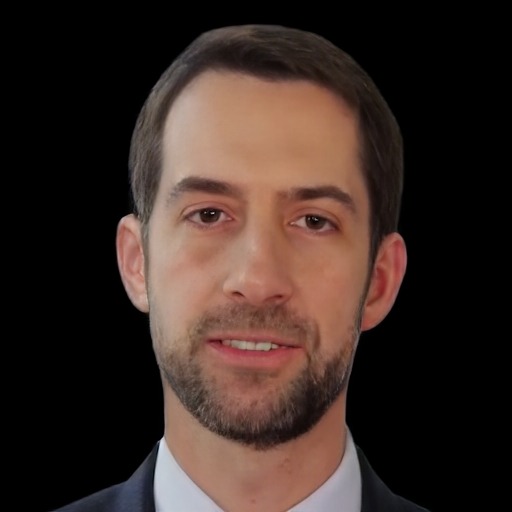} &
\includegraphics[width=0.125\textwidth]{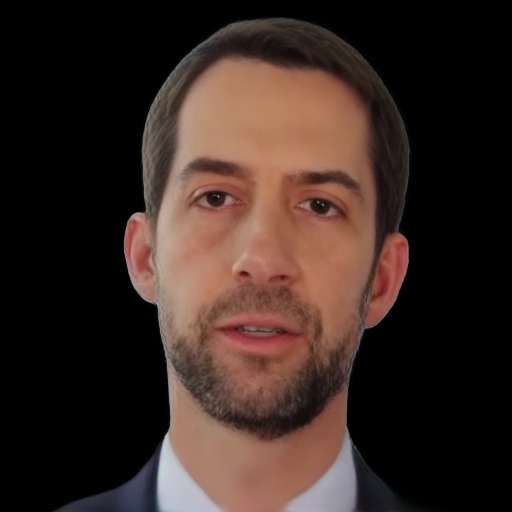} &
\includegraphics[width=0.125\textwidth]{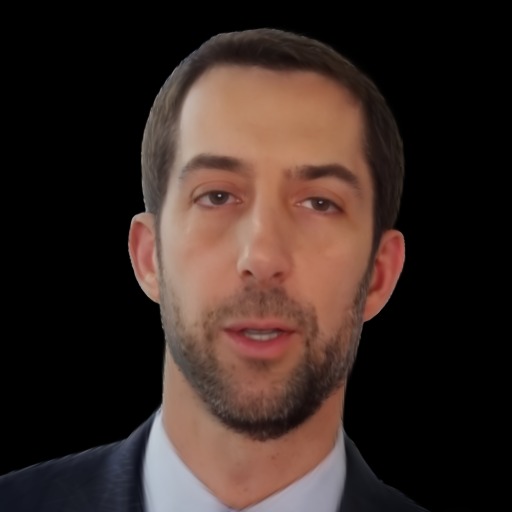} &
\includegraphics[width=0.125\textwidth]{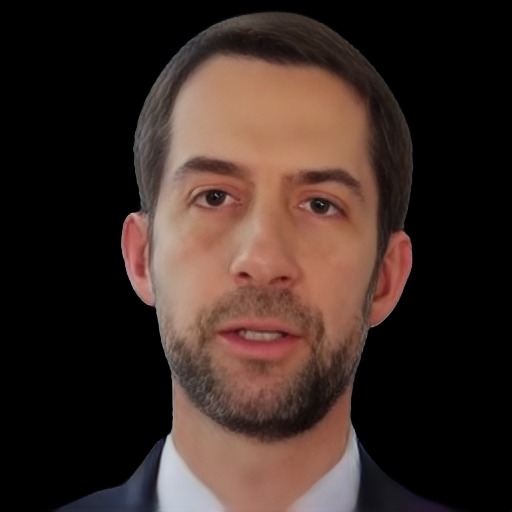}  \\
\includegraphics[width=0.125\textwidth]{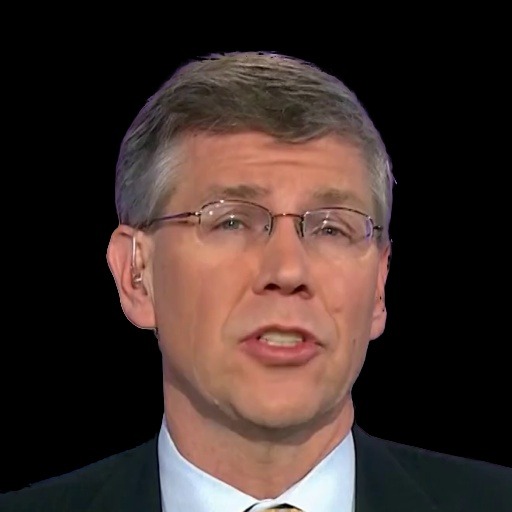} &
\includegraphics[width=0.125\textwidth]{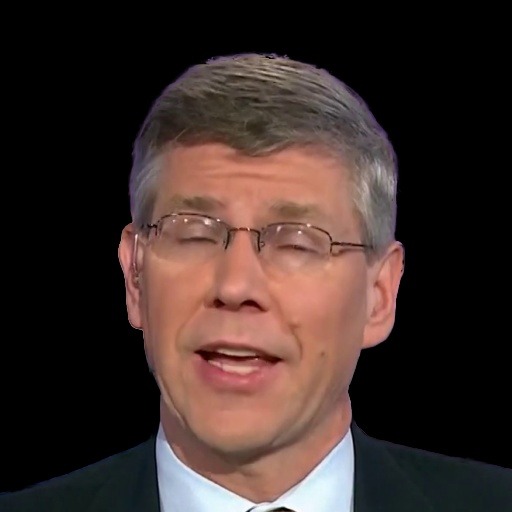} &
\includegraphics[width=0.125\textwidth]{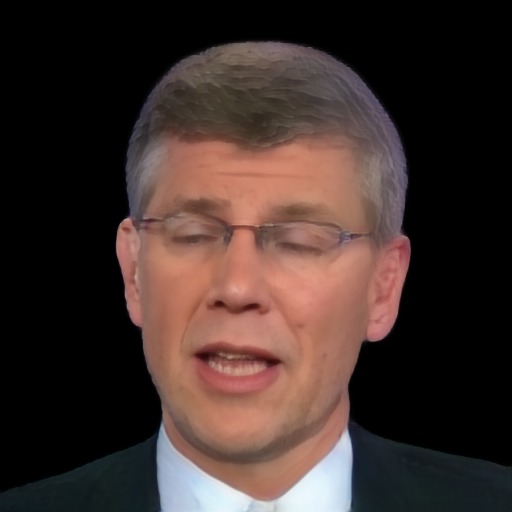} &
\includegraphics[width=0.125\textwidth]{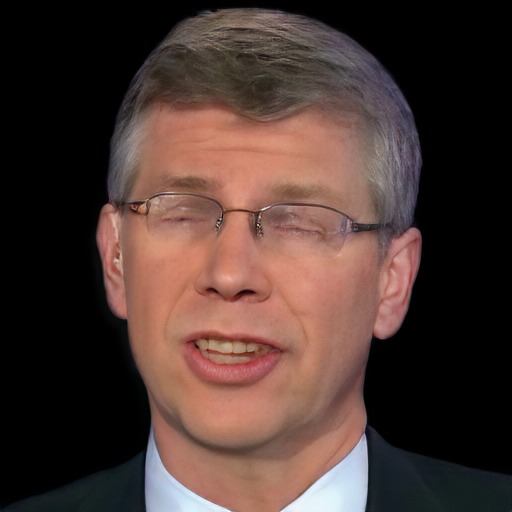} &
\includegraphics[width=0.125\textwidth]{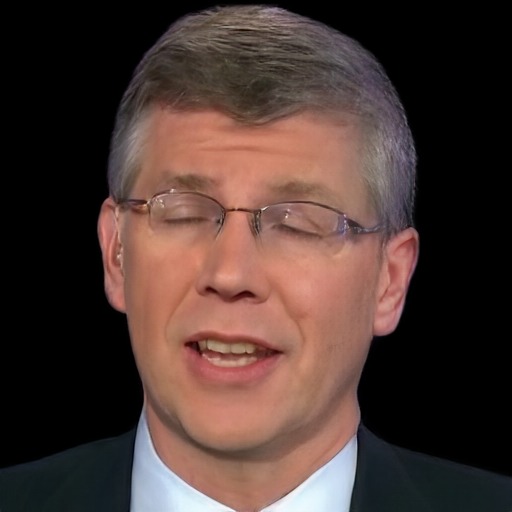} &
\includegraphics[width=0.125\textwidth]{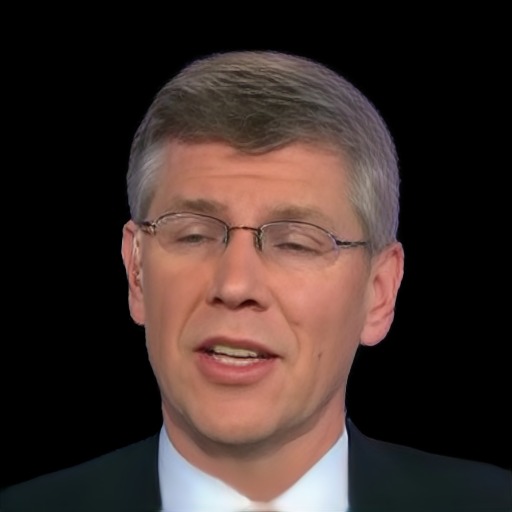} &
\includegraphics[width=0.125\textwidth]{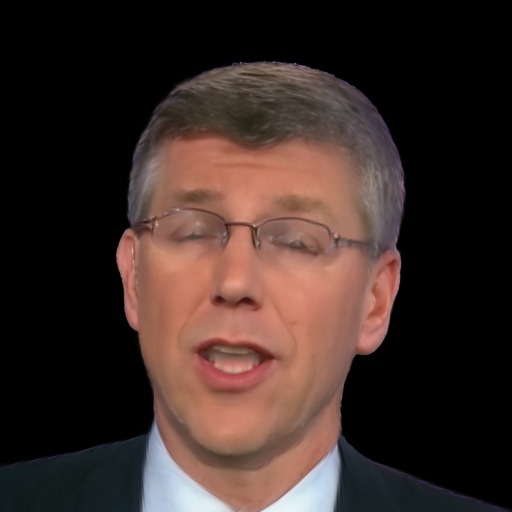} &
\includegraphics[width=0.125\textwidth]{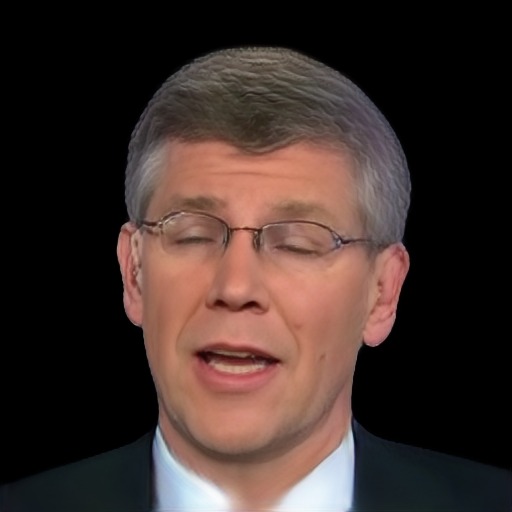}  \\
\includegraphics[width=0.125\textwidth]{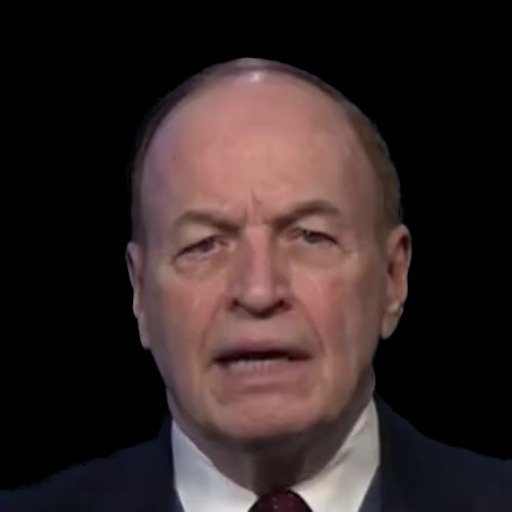} &
\includegraphics[width=0.125\textwidth]{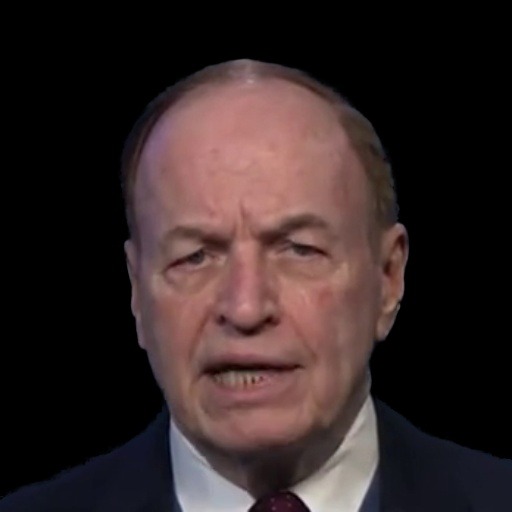} &
\includegraphics[width=0.125\textwidth]{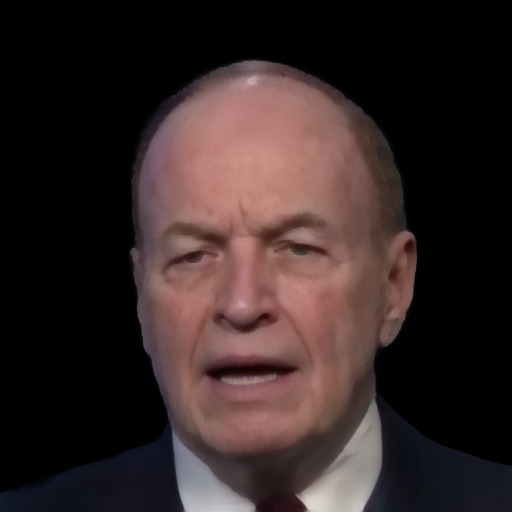} &
\includegraphics[width=0.125\textwidth]{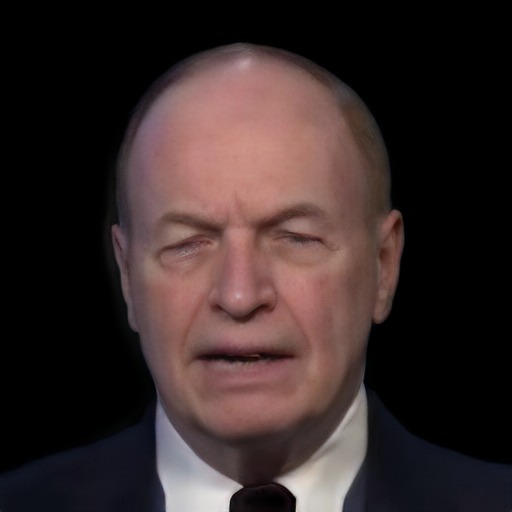} &
\includegraphics[width=0.125\textwidth]{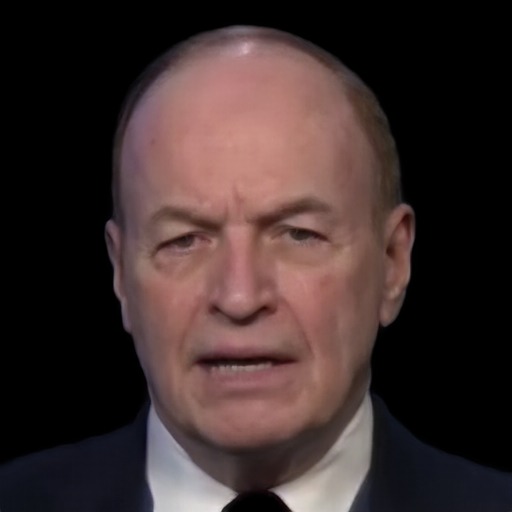} &
\includegraphics[width=0.125\textwidth]{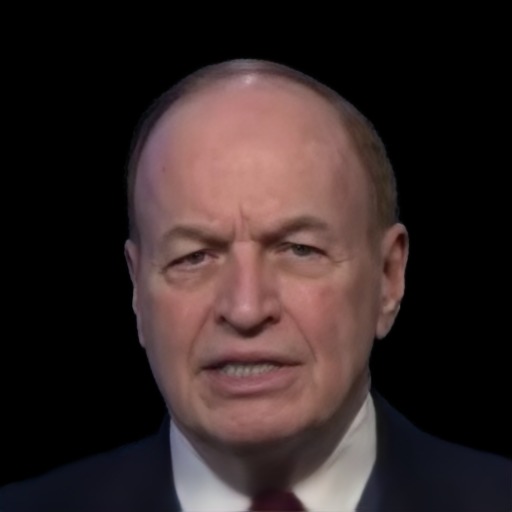} &
\includegraphics[width=0.125\textwidth]{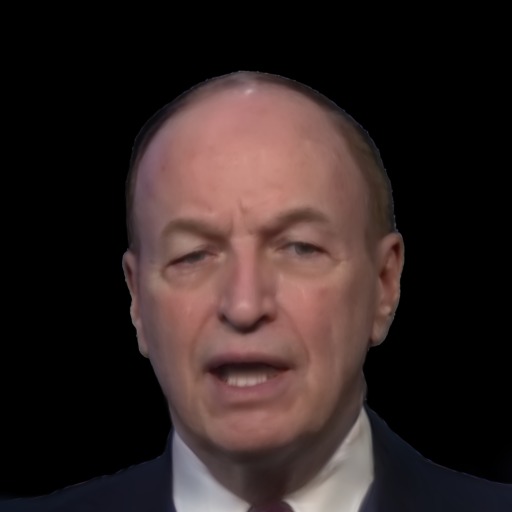} &
\includegraphics[width=0.125\textwidth]{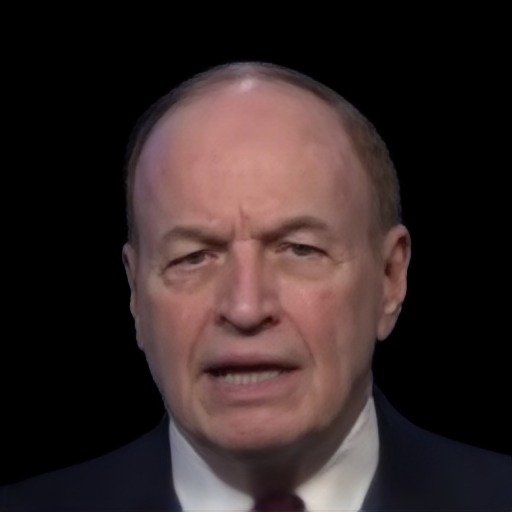}  \\
\includegraphics[width=0.125\textwidth]{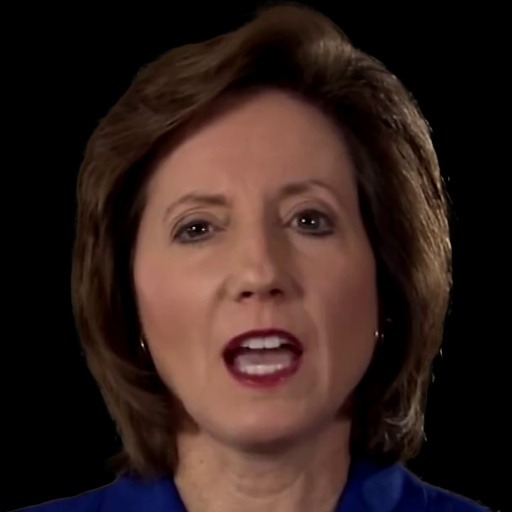} &
\includegraphics[width=0.125\textwidth]{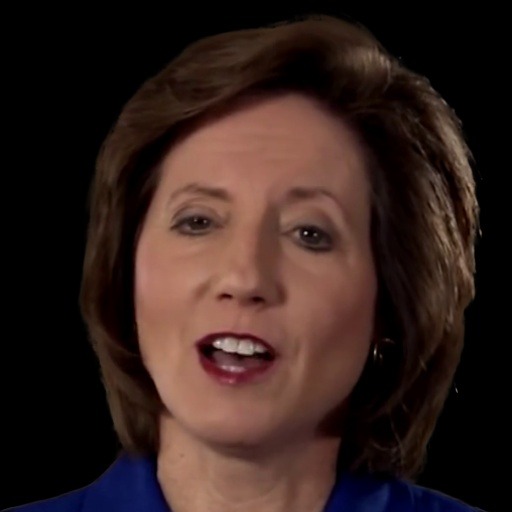} &
\includegraphics[width=0.125\textwidth]{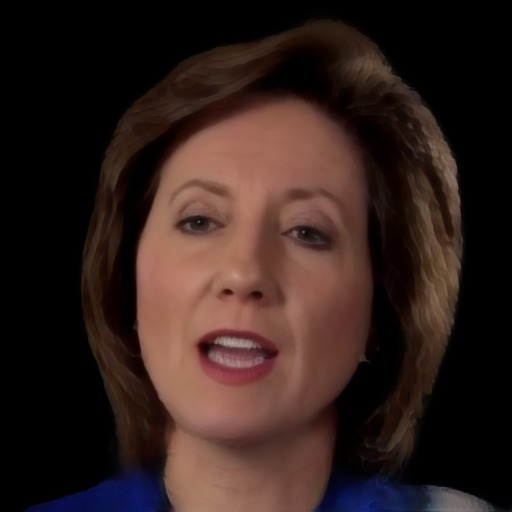} &
\includegraphics[width=0.125\textwidth]{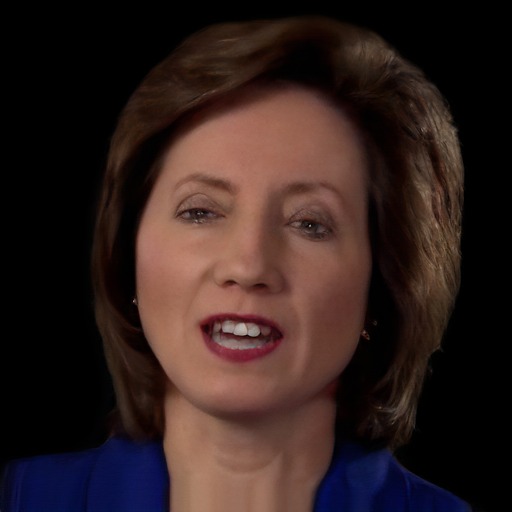} &
\includegraphics[width=0.125\textwidth]{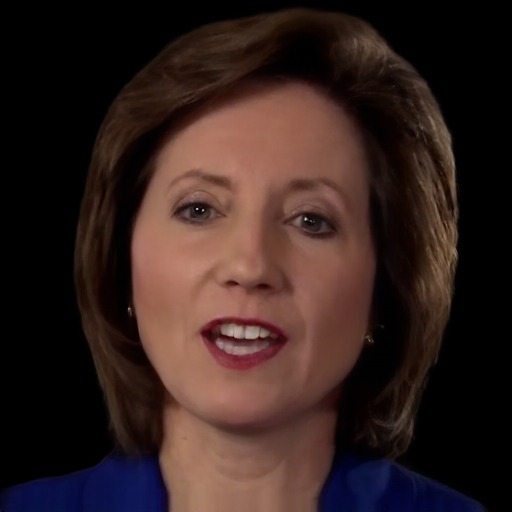} &
\includegraphics[width=0.125\textwidth]{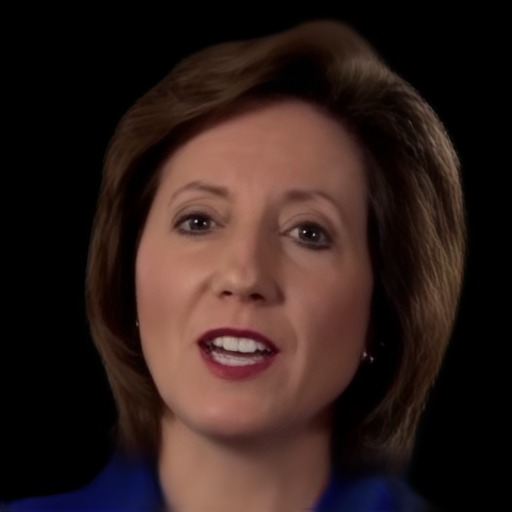} &
\includegraphics[width=0.125\textwidth]{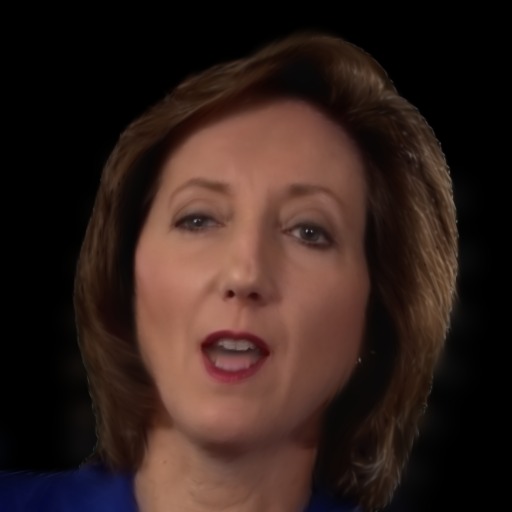} &
\includegraphics[width=0.125\textwidth]{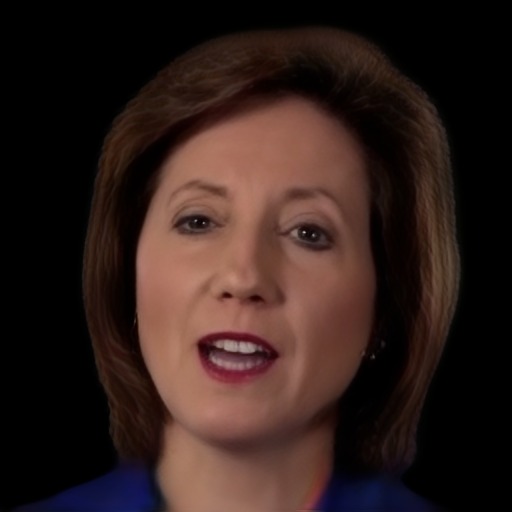}  \\
[5pt]
Source & Driving & GPAvatar & P4D & P4D-v2 & GAG & LAM & Ours \\
\end{tabular}
}
\caption{Self reenactment results on VFHQ and HDTF datasets.}
\label{fig:supp-self}
\vspace{-5mm}
\end{figure}

\begin{figure}[!p]
\centering
\setlength{\tabcolsep}{0pt}
\renewcommand{\arraystretch}{0}
\resizebox{\textwidth}{!}{
\begin{tabular}{@{}ccccccc@{}}
&
\includegraphics[width=0.142857\textwidth]{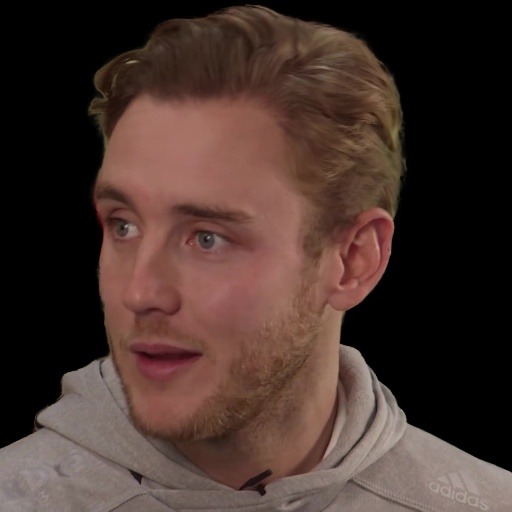} &
\includegraphics[width=0.142857\textwidth]{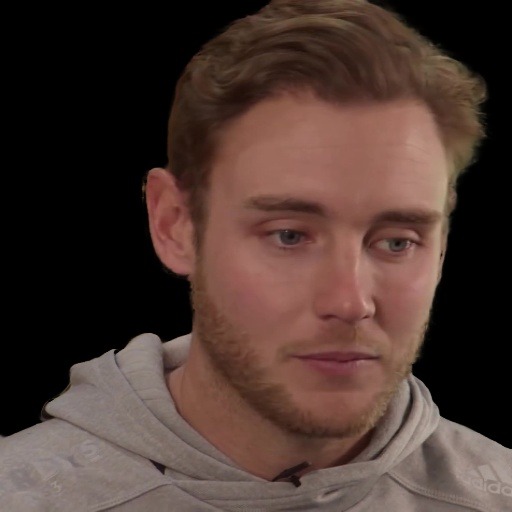} &
\includegraphics[width=0.142857\textwidth]{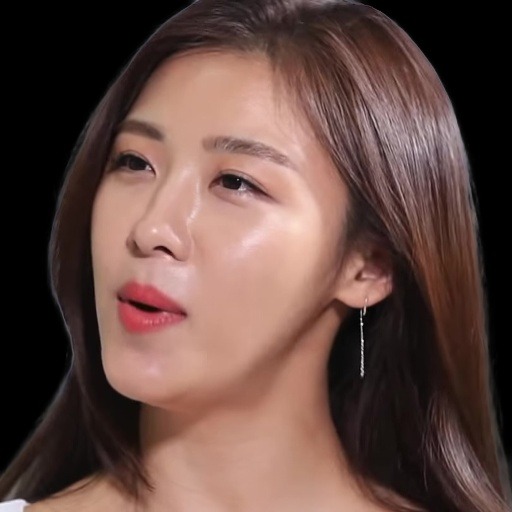} &
\includegraphics[width=0.142857\textwidth]{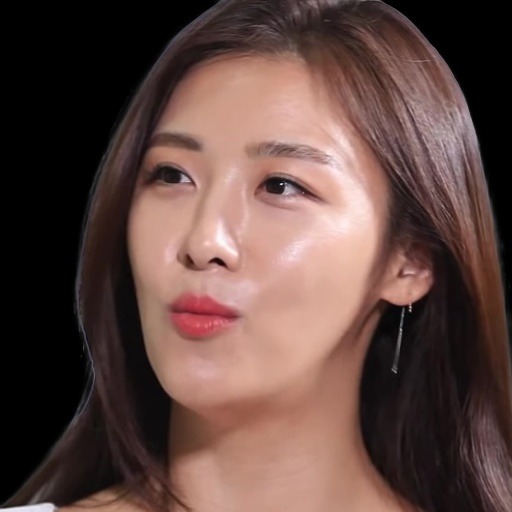} &
\includegraphics[width=0.142857\textwidth]{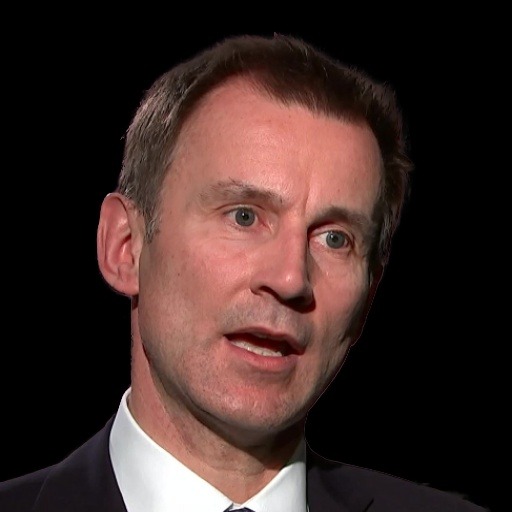} &
\includegraphics[width=0.142857\textwidth]{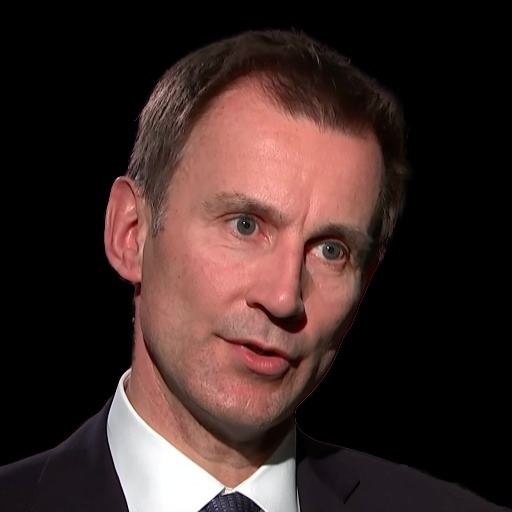}  \\
\includegraphics[width=0.142857\textwidth]{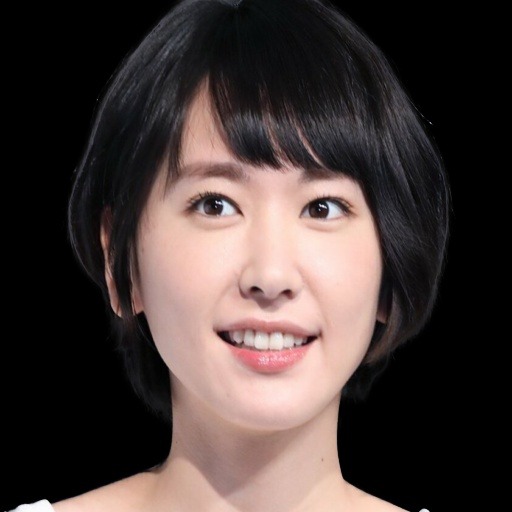} &
\includegraphics[width=0.142857\textwidth]{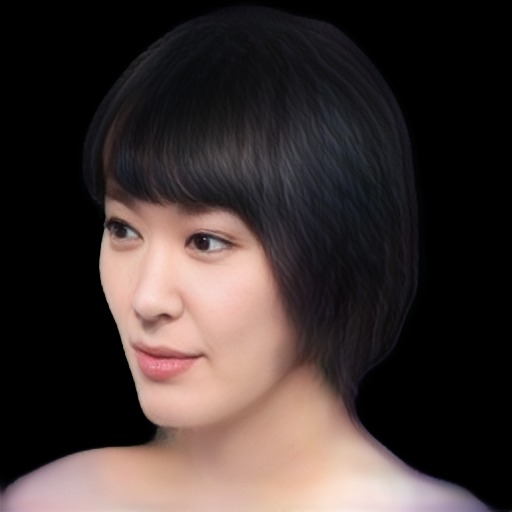} &
\includegraphics[width=0.142857\textwidth]{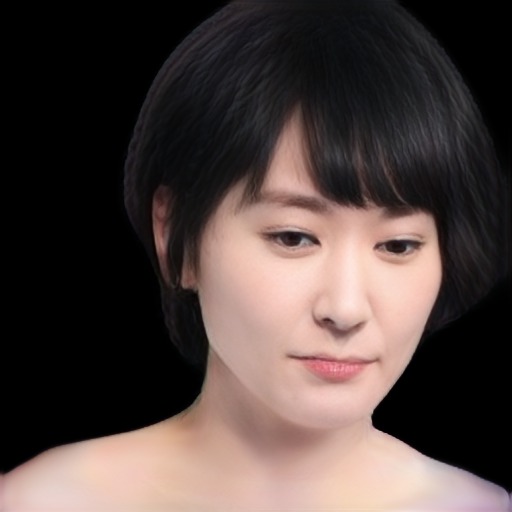} &
\includegraphics[width=0.142857\textwidth]{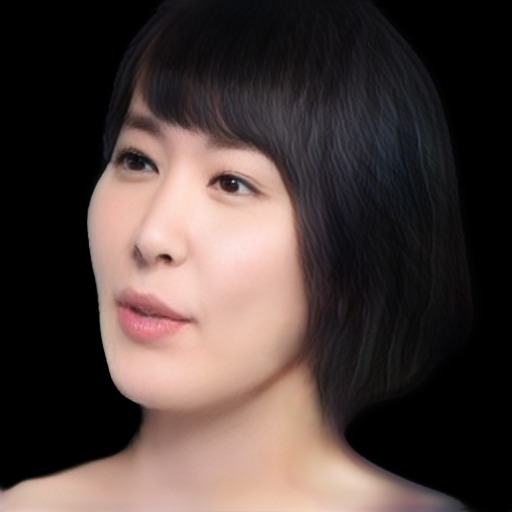} &
\includegraphics[width=0.142857\textwidth]{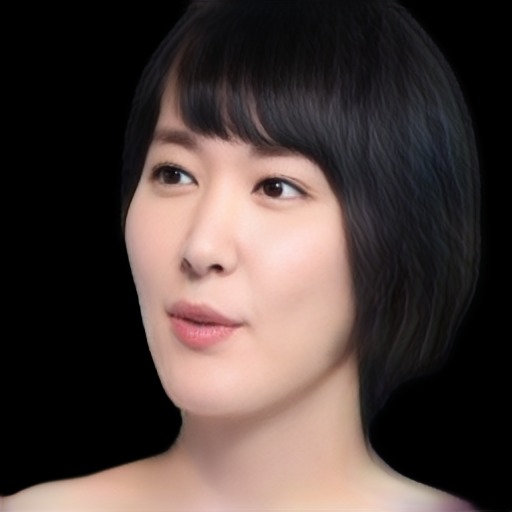} &
\includegraphics[width=0.142857\textwidth]{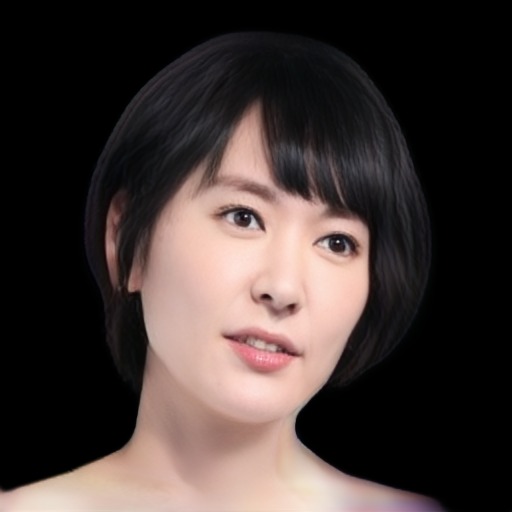} &
\includegraphics[width=0.142857\textwidth]{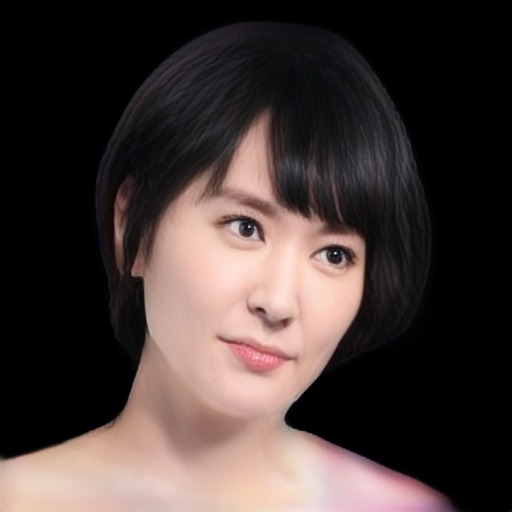}  \\
\includegraphics[width=0.142857\textwidth]{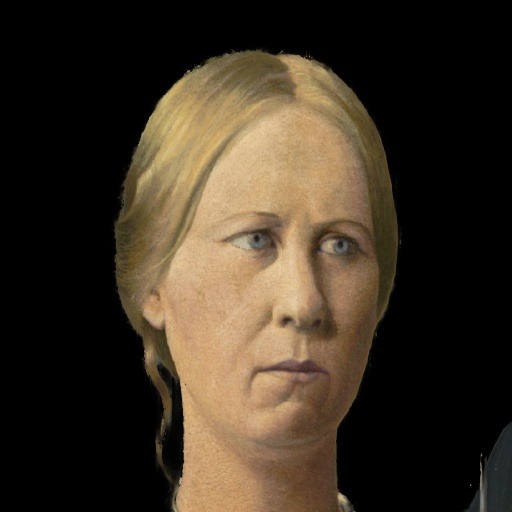} &
\includegraphics[width=0.142857\textwidth]{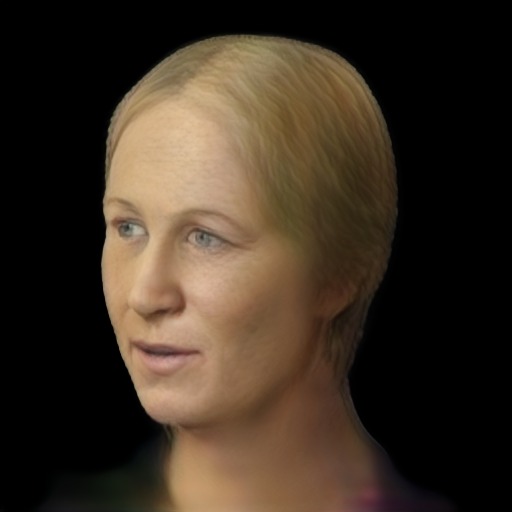} &
\includegraphics[width=0.142857\textwidth]{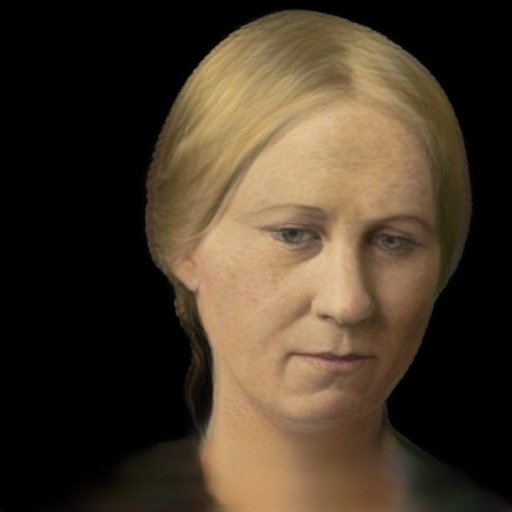} &
\includegraphics[width=0.142857\textwidth]{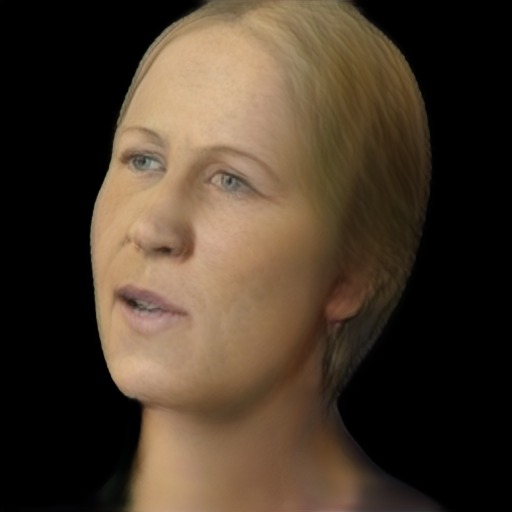} &
\includegraphics[width=0.142857\textwidth]{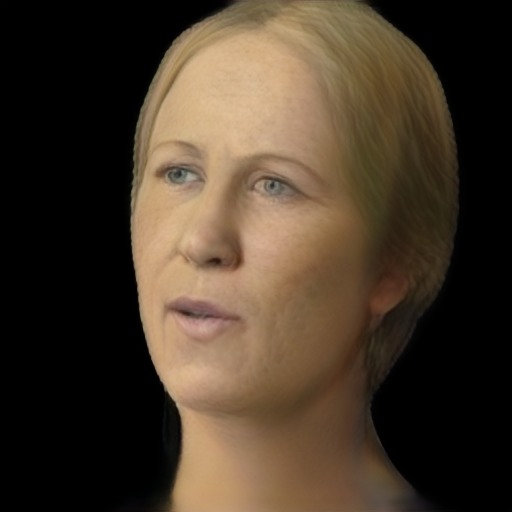} &
\includegraphics[width=0.142857\textwidth]{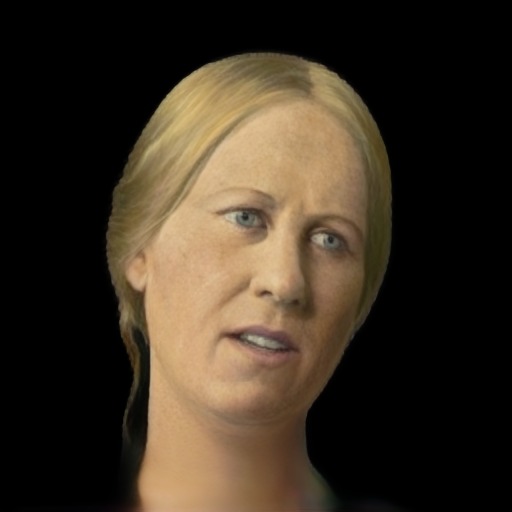} &
\includegraphics[width=0.142857\textwidth]{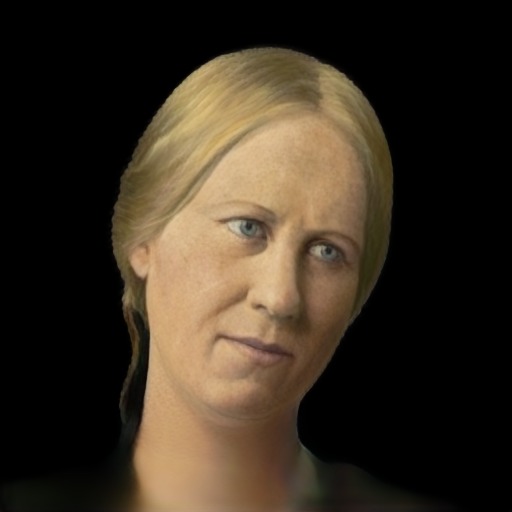}  \\
\includegraphics[width=0.142857\textwidth]{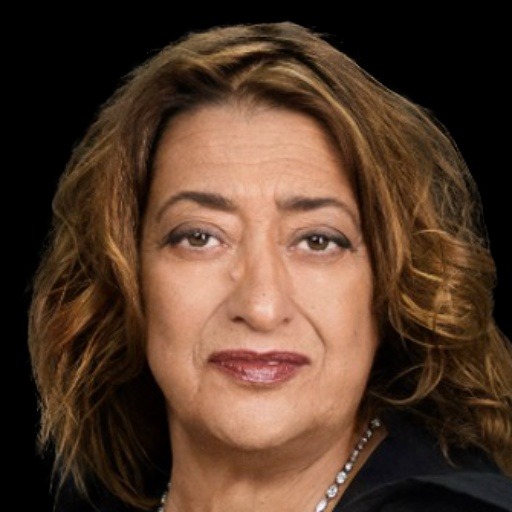} &
\includegraphics[width=0.142857\textwidth]{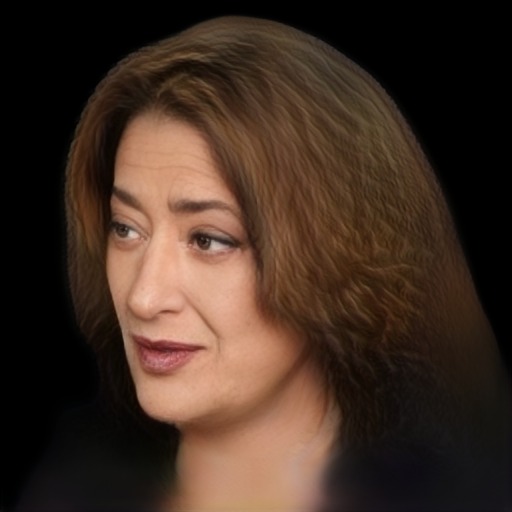} &
\includegraphics[width=0.142857\textwidth]{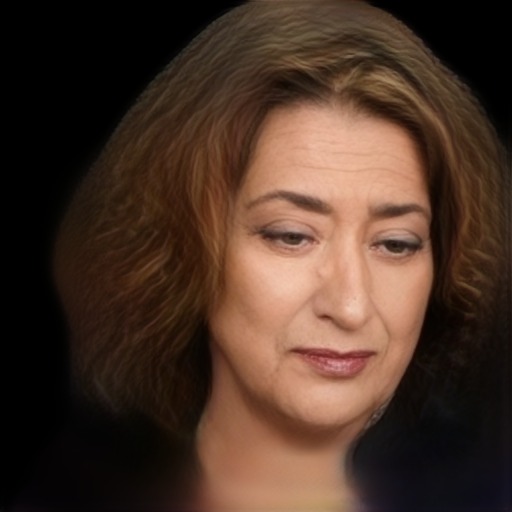} &
\includegraphics[width=0.142857\textwidth]{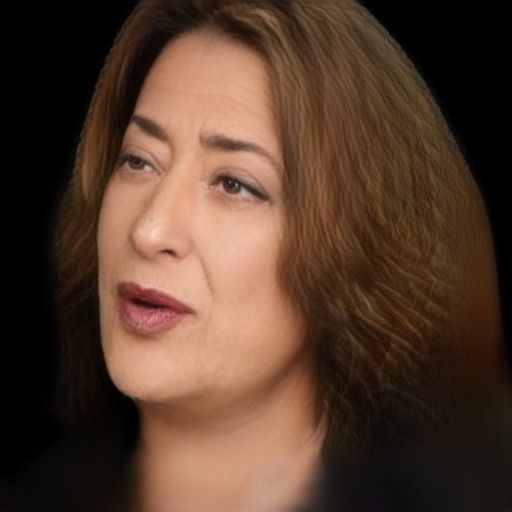} &
\includegraphics[width=0.142857\textwidth]{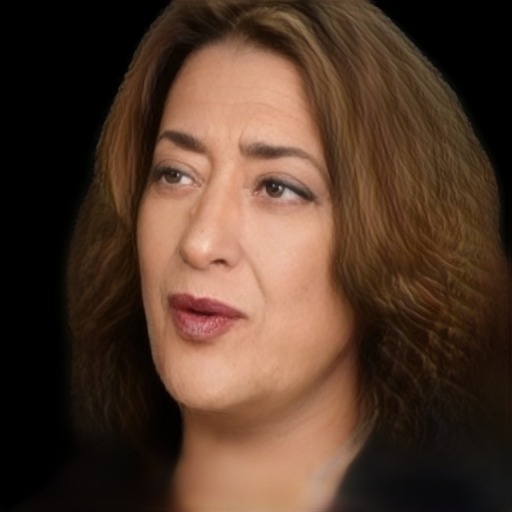} &
\includegraphics[width=0.142857\textwidth]{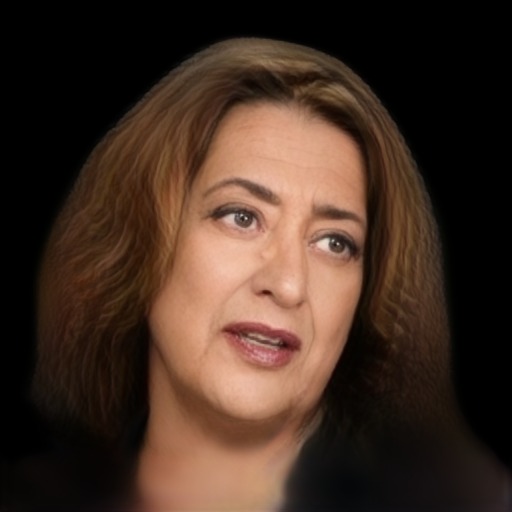} &
\includegraphics[width=0.142857\textwidth]{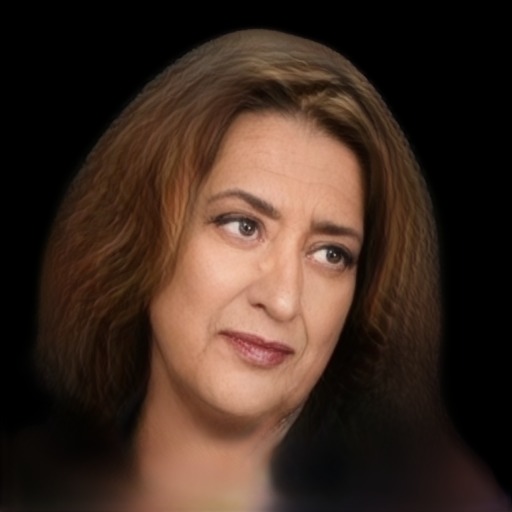}  \\
\includegraphics[width=0.142857\textwidth]{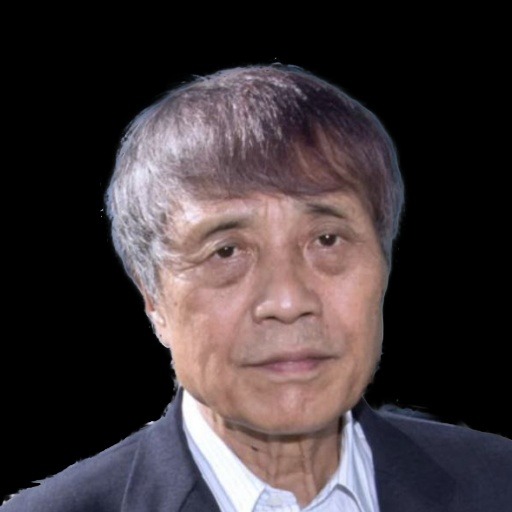} &
\includegraphics[width=0.142857\textwidth]{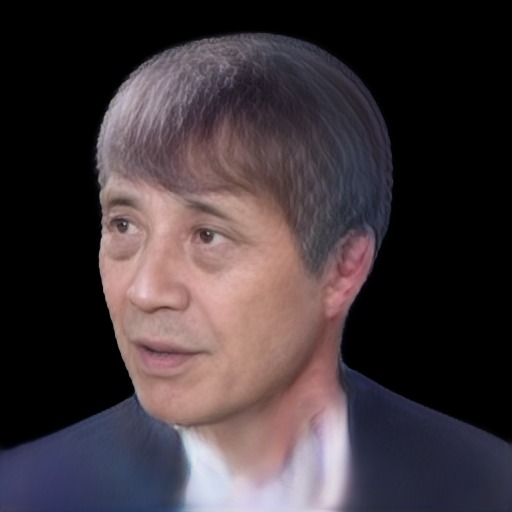} &
\includegraphics[width=0.142857\textwidth]{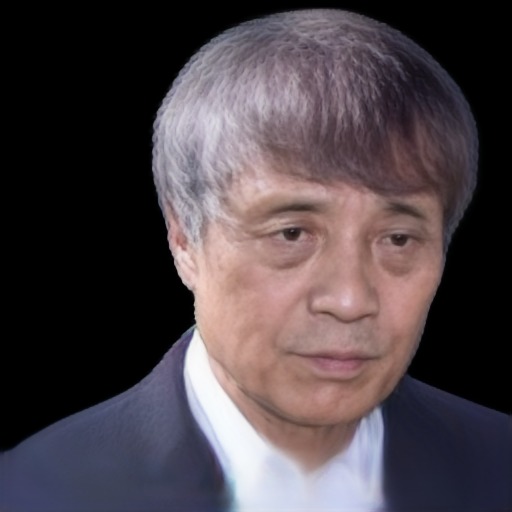} &
\includegraphics[width=0.142857\textwidth]{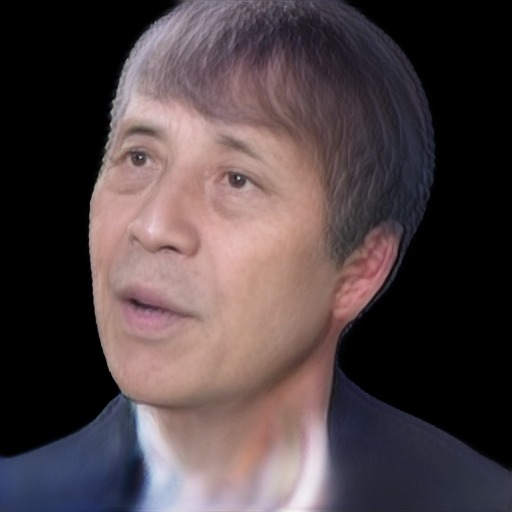} &
\includegraphics[width=0.142857\textwidth]{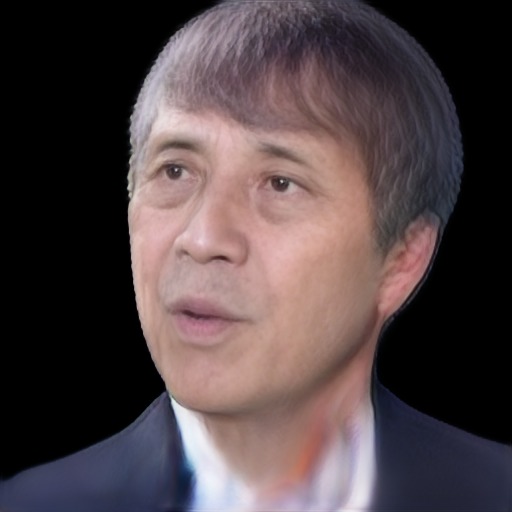} &
\includegraphics[width=0.142857\textwidth]{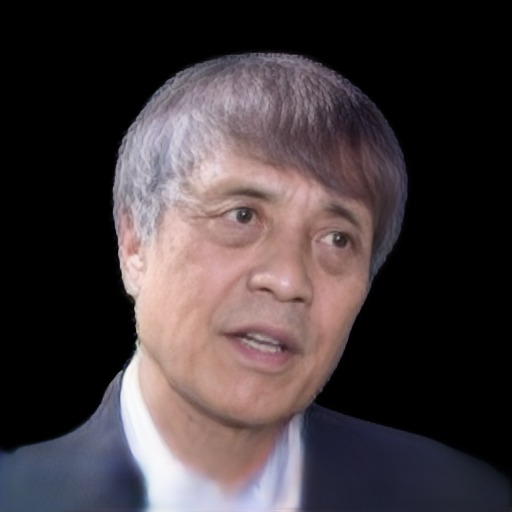} &
\includegraphics[width=0.142857\textwidth]{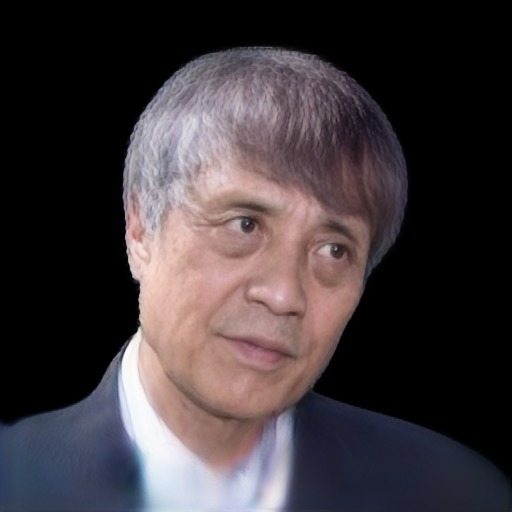}  \\
\includegraphics[width=0.142857\textwidth]{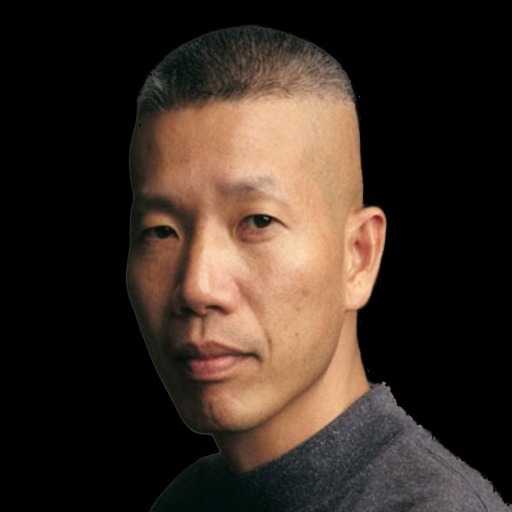} &
\includegraphics[width=0.142857\textwidth]{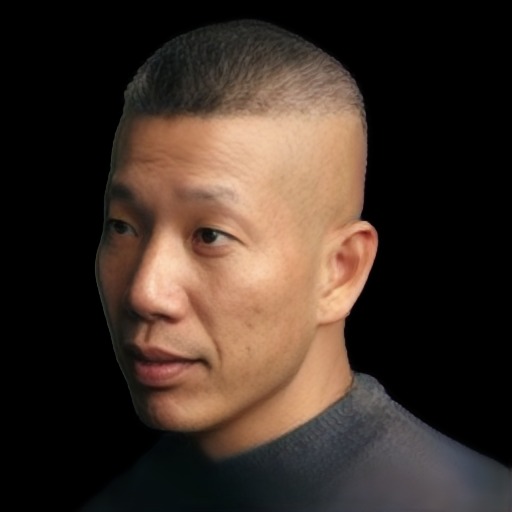} &
\includegraphics[width=0.142857\textwidth]{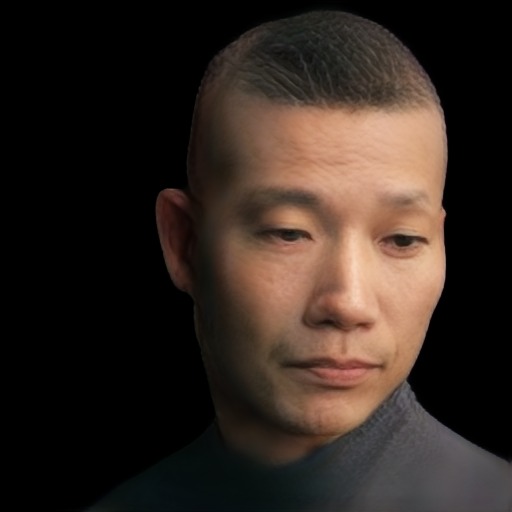} &
\includegraphics[width=0.142857\textwidth]{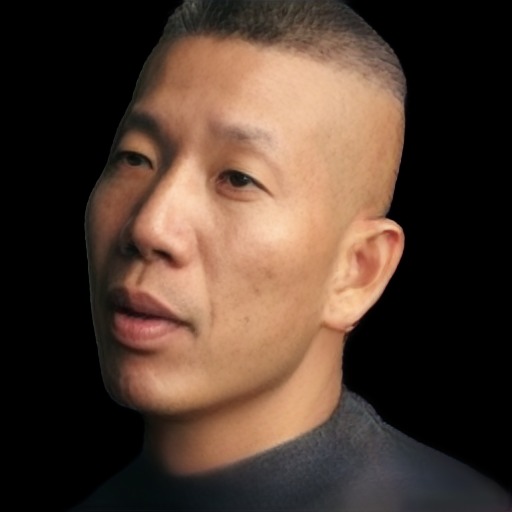} &
\includegraphics[width=0.142857\textwidth]{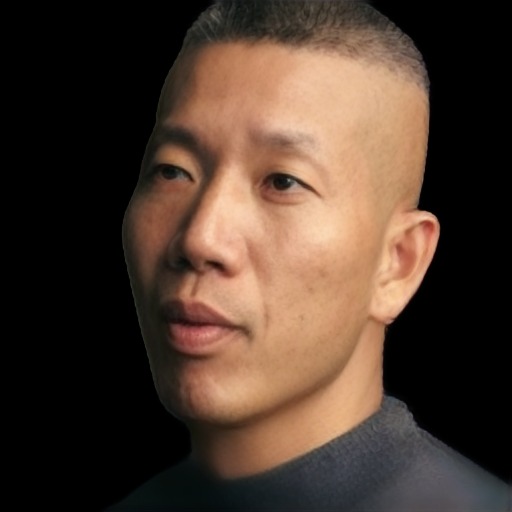} &
\includegraphics[width=0.142857\textwidth]{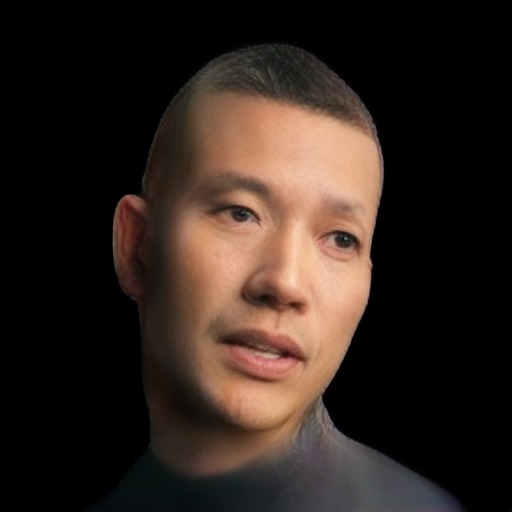} &
\includegraphics[width=0.142857\textwidth]{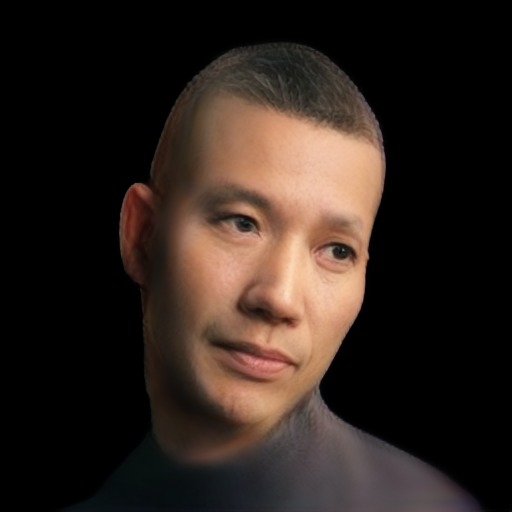}  \\
\includegraphics[width=0.142857\textwidth]{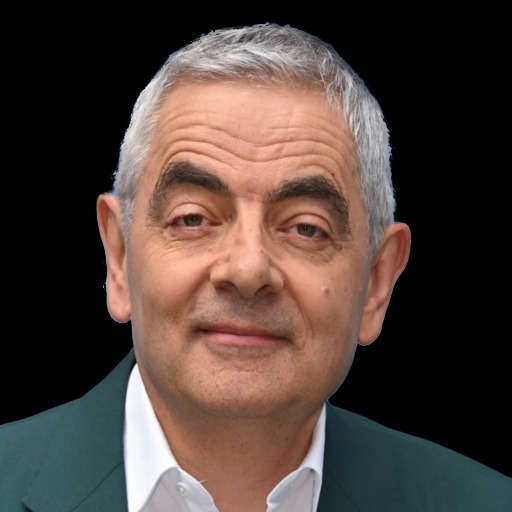} &
\includegraphics[width=0.142857\textwidth]{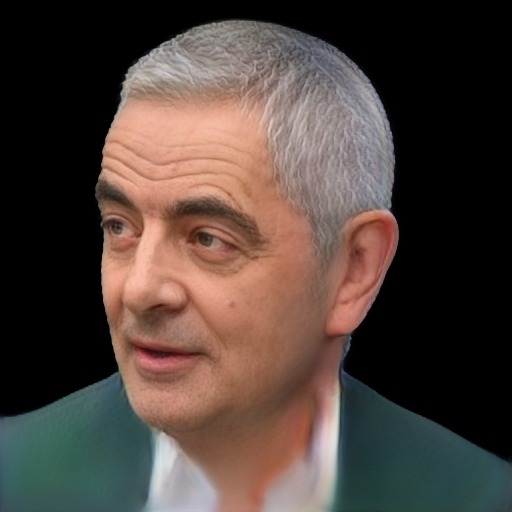} &
\includegraphics[width=0.142857\textwidth]{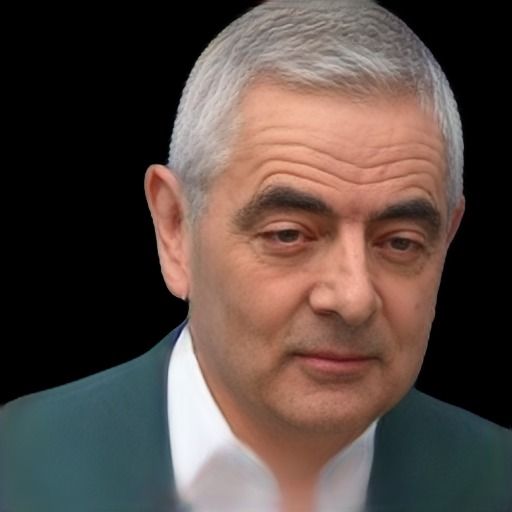} &
\includegraphics[width=0.142857\textwidth]{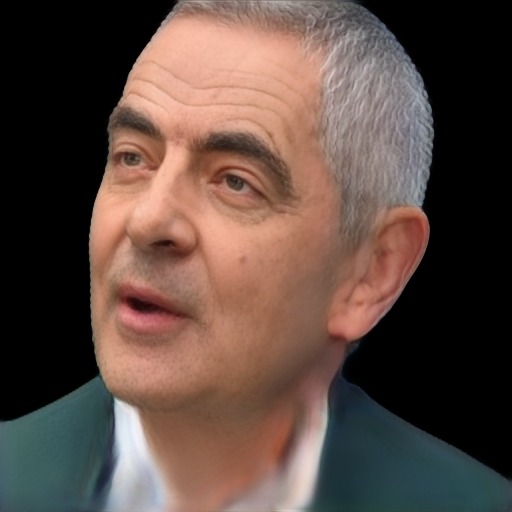} &
\includegraphics[width=0.142857\textwidth]{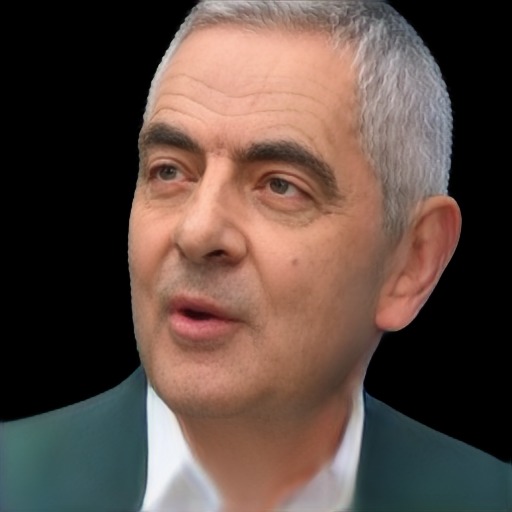} &
\includegraphics[width=0.142857\textwidth]{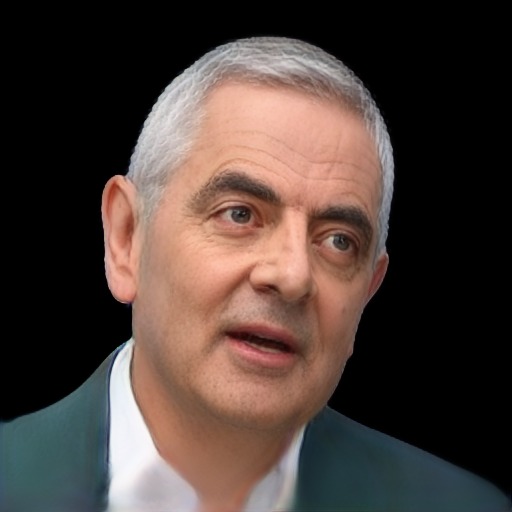} &
\includegraphics[width=0.142857\textwidth]{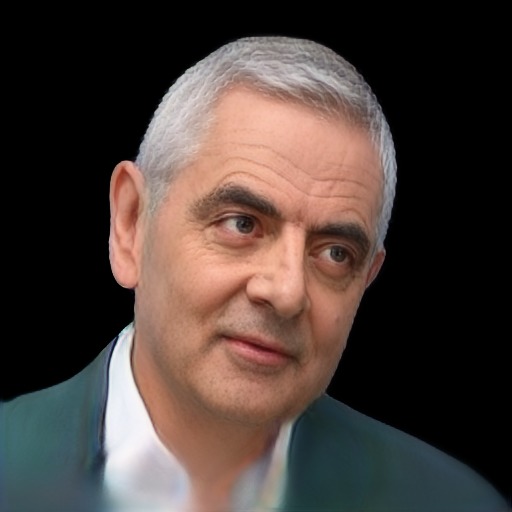}  \\
\includegraphics[width=0.142857\textwidth]{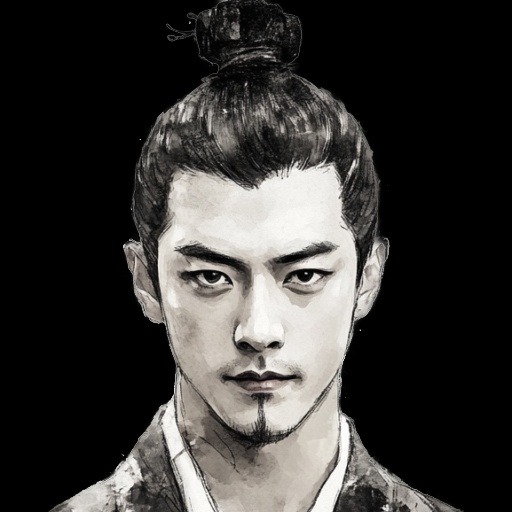} &
\includegraphics[width=0.142857\textwidth]{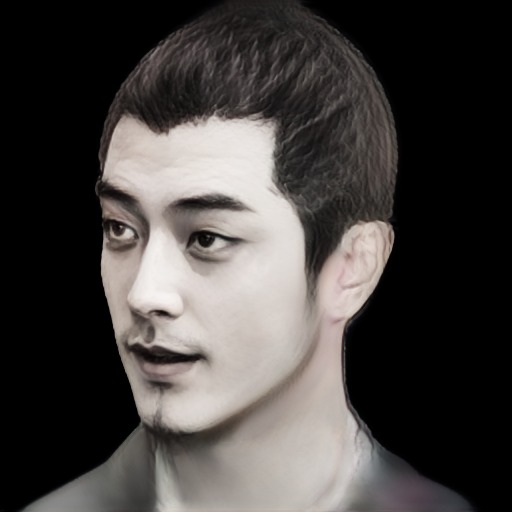} &
\includegraphics[width=0.142857\textwidth]{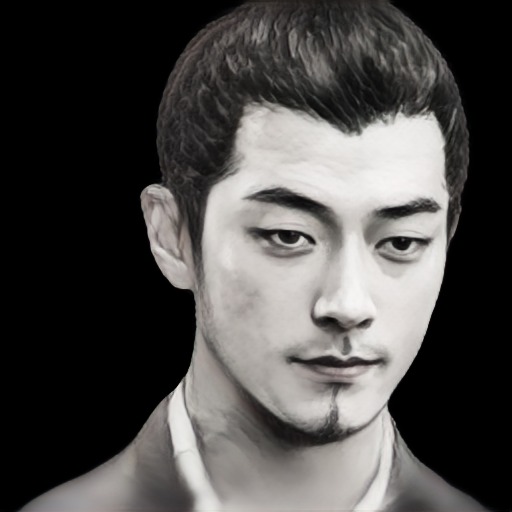} &
\includegraphics[width=0.142857\textwidth]{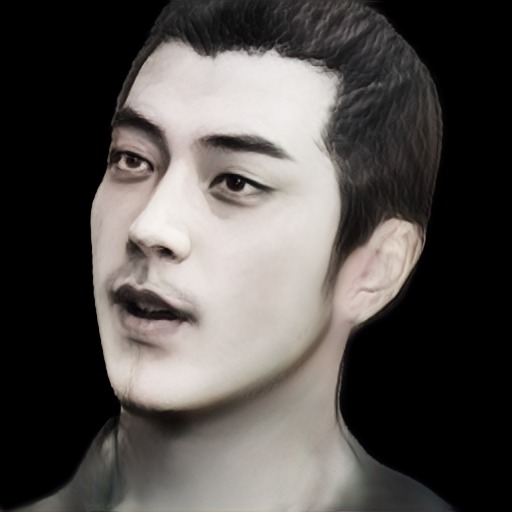} &
\includegraphics[width=0.142857\textwidth]{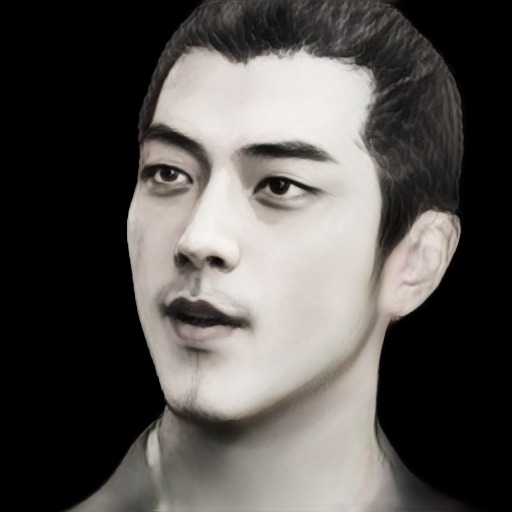} &
\includegraphics[width=0.142857\textwidth]{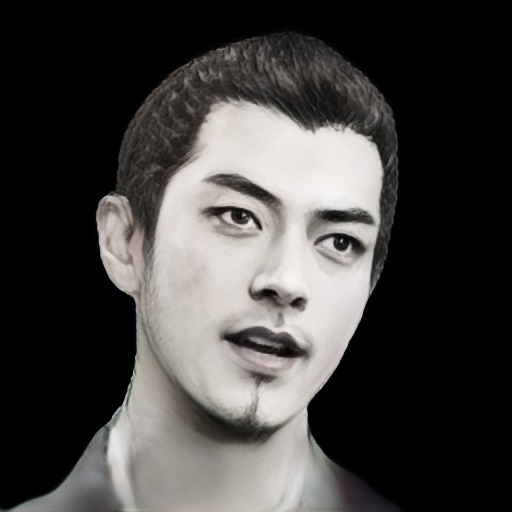} &
\includegraphics[width=0.142857\textwidth]{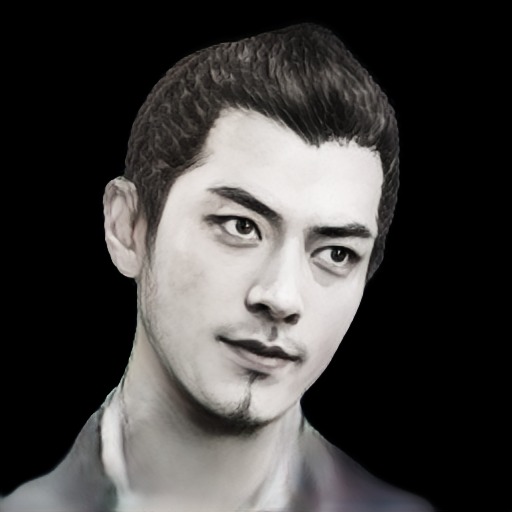}  \\
\includegraphics[width=0.142857\textwidth]{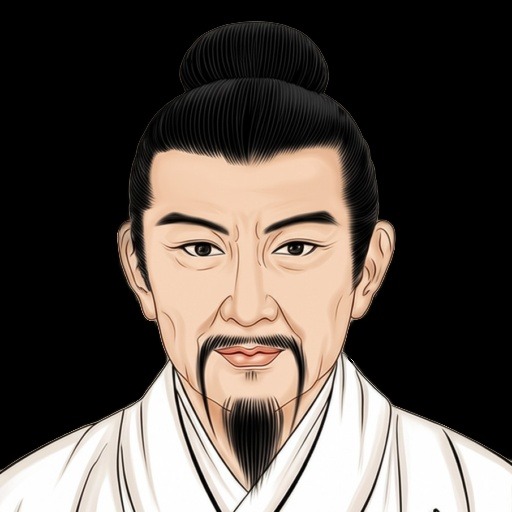} &
\includegraphics[width=0.142857\textwidth]{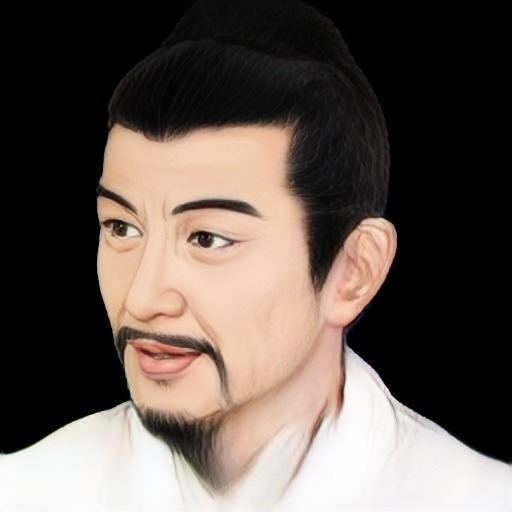} &
\includegraphics[width=0.142857\textwidth]{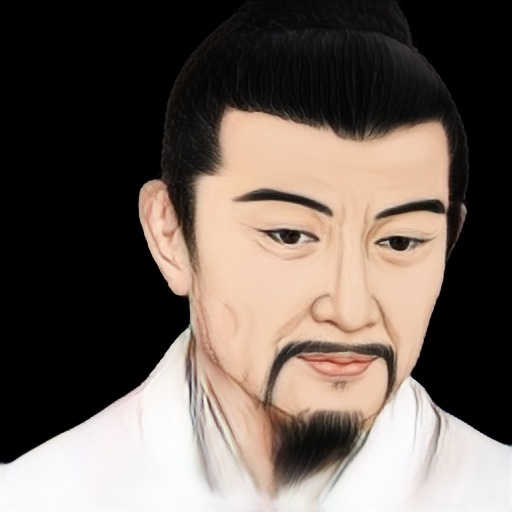} &
\includegraphics[width=0.142857\textwidth]{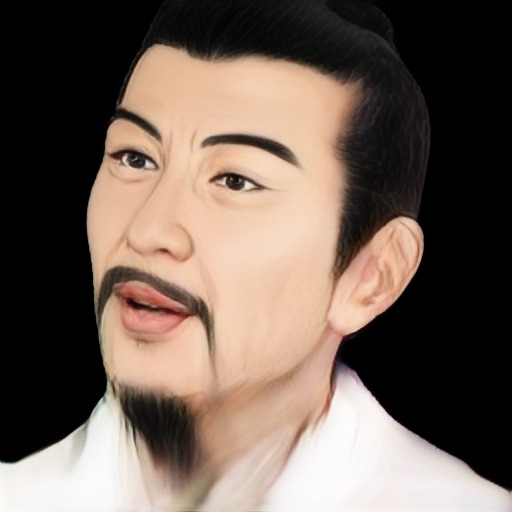} &
\includegraphics[width=0.142857\textwidth]{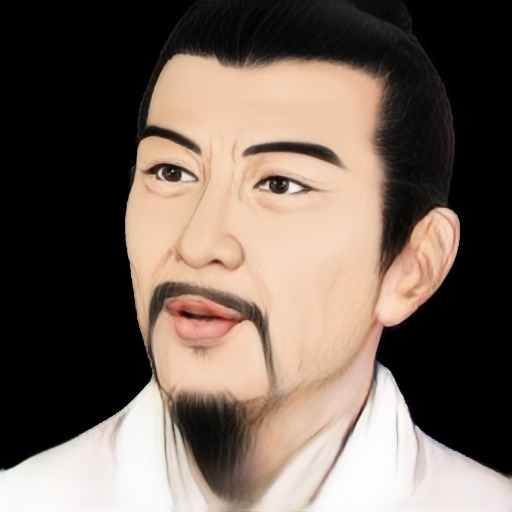} &
\includegraphics[width=0.142857\textwidth]{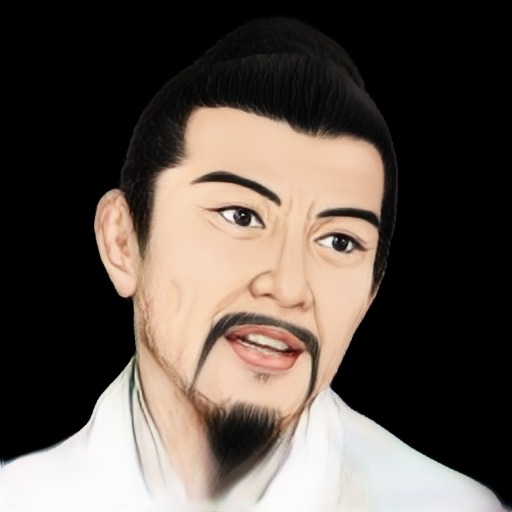} &
\includegraphics[width=0.142857\textwidth]{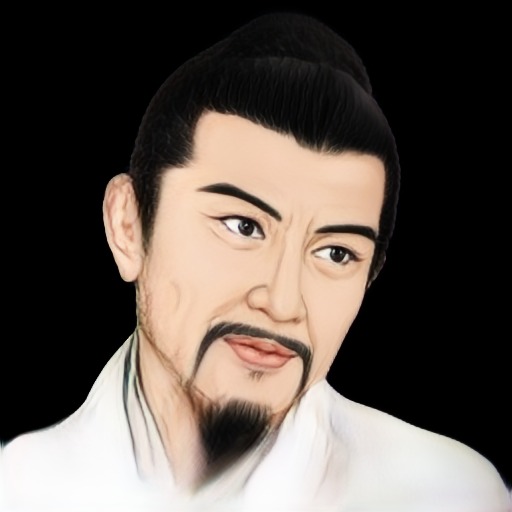}  \\
\includegraphics[width=0.142857\textwidth]{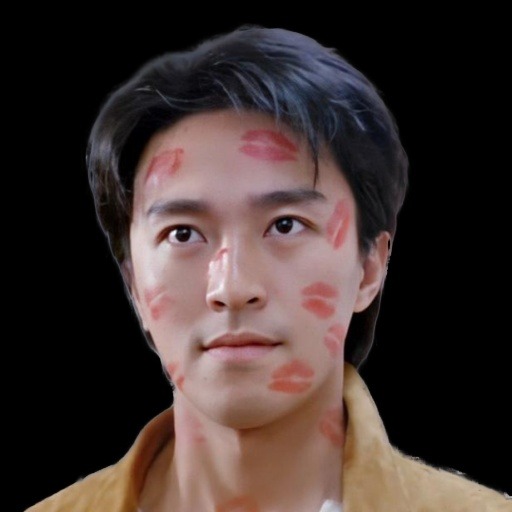} &
\includegraphics[width=0.142857\textwidth]{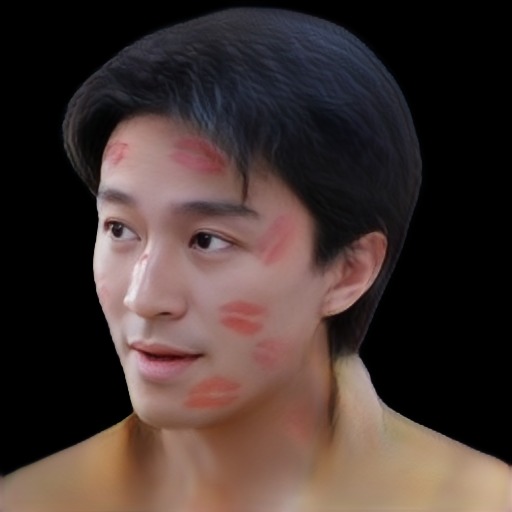} &
\includegraphics[width=0.142857\textwidth]{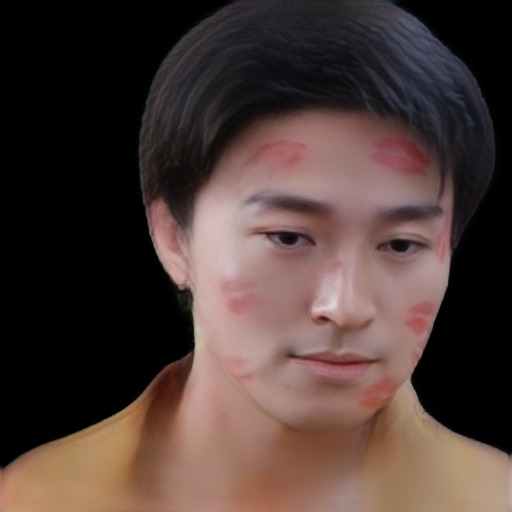} &
\includegraphics[width=0.142857\textwidth]{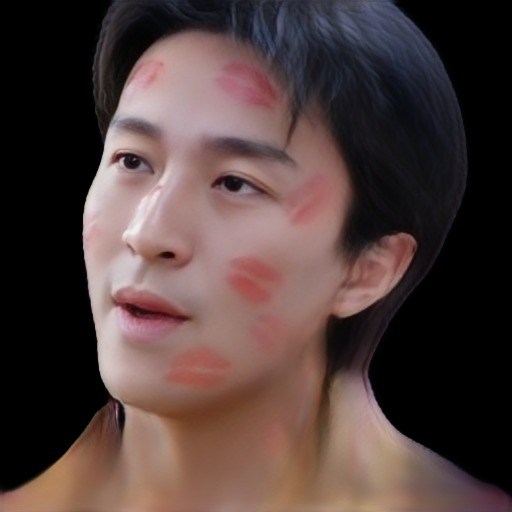} &
\includegraphics[width=0.142857\textwidth]{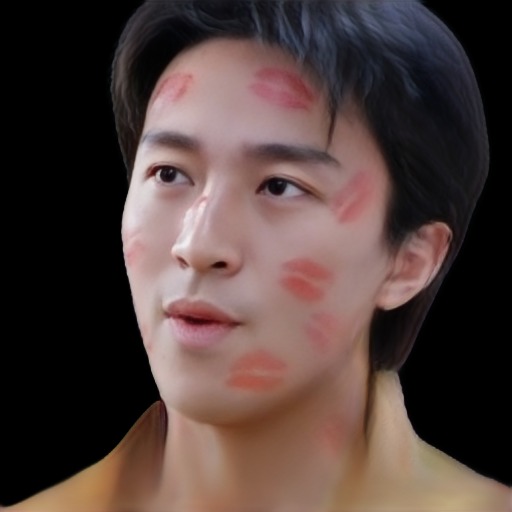} &
\includegraphics[width=0.142857\textwidth]{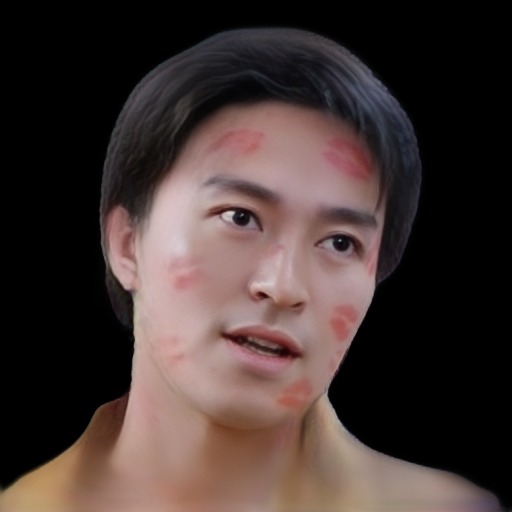} &
\includegraphics[width=0.142857\textwidth]{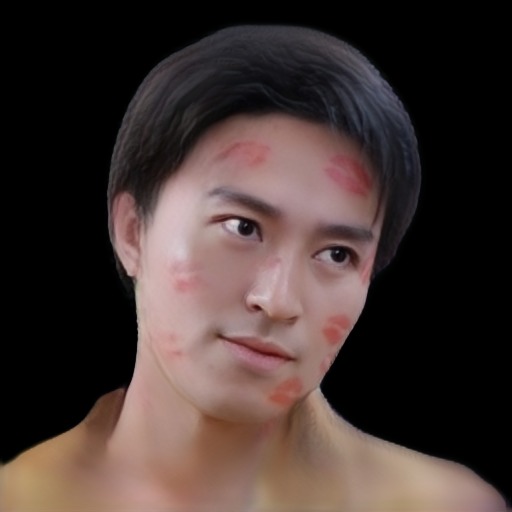}  \\
\end{tabular}
}
\caption{Qualitative results of our method on in-the-wild images. Source images are shown in the first column and driving images in the top row, and the remaining images show the generated results.}
\label{fig:supp-wild}
\vspace{-5mm}
\end{figure}

We show more qualitative results on the VFHQ and HDTF datasets, including cross reenactment comparisons in Fig.~\ref{fig:supp-cross} and self reenactment comparisons in Fig.~\ref{fig:supp-self}. Even in self reenactment cases with small pose differences between the source and driving frames, our method better transfers the driving expression while preserving fine identity details. We further show qualitative results of our method on in-the-wild images in Fig.~\ref{fig:supp-wild}. These results demonstrate that our method generalizes well while preserving identity and 3D consistency across varying viewpoints.

\section{Dataset Licenses and Ethical Considerations}
\label{license}
Our experiments are conducted using two publicly available datasets: VFHQ~\cite{xie2022vfhq} and HDTF~\cite{zhang2021flow}. These datasets are released for non-commercial research use. We further include in-the-wild images for qualitative evaluation to assess generalization under unconstrained conditions. Beyond the facial images required for the task, we do not use any additional personally identifiable information.


\section{Potential Societal Impacts}
\label{impact}
Photorealistic human head avatars can raise concerns regarding potential misuse, such as the creation of deceptive manipulated media involving real individuals. We do not support or encourage such applications. One possible countermeasure is to train deepfake detection models~\cite{malp,reverse,finegrined,yang2022confidence,kang2025legion,xu2025fakeshield} using avatars generated by our method, enabling them to better separate synthesized avatar renderings from real images.



%
%
\bibliographystyle{splncs04}
\bibliography{main}